\documentclass[10pt,twocolumn,letterpaper]{article}

\usepackage[pagenumbers]{cvpr}
\usepackage[pagebackref,breaklinks,colorlinks,allcolors=cvprblue]{hyperref}
\usepackage{url}
\usepackage{enumitem}
\usepackage{amssymb}            
\usepackage{mathtools}          
\usepackage{mathrsfs} 
\usepackage{wrapfig}
\usepackage{graphicx}           
\usepackage{subcaption}         
\usepackage[space]{grffile}     
\usepackage{url}
\usepackage{lipsum}             
\usepackage{booktabs}
\usepackage{tabularx}
\usepackage{amsmath}    
\usepackage{float} 

\usepackage{amssymb}    
\usepackage{algorithmic} 
\usepackage{mathtools}  
\setcitestyle{numbers}

\usepackage{amsmath, amssymb, mathtools} 
\usepackage{tcolorbox}
\usepackage[ruled,vlined]{algorithm2e}   
\usepackage{graphicx}
\usepackage{subcaption}
\usepackage{multicol}
\usepackage{caption}
\usepackage{adjustbox} 
\usepackage{amsthm}
\usepackage{placeins}

\theoremstyle{definition}

\usepackage[dvipsnames,table]{xcolor}
\newcolumntype{Y}{>{\raggedright\arraybackslash}X} 

\usepackage{tikz}

\definecolor{cvprblue}{rgb}{0.21,0.49,0.74}

\def\paperID{21275} 
\def\confName{CVPR}
\def\confYear{2026}

\title{World Model Robustness via Surprise Recognition}

\author{Geigh Zollicoffer$^{*}$, Tanush Chopra$^{*}$, Mingkuan Yan, Xiaoxu Ma, Kenneth Eaton, Mark Riedl\\
Georgia Institute of Technology\\
{\tt\small \{gzollicoffer3, tchopra32, myan71, xma394, keaton30, riedl\}@gatech.edu}
\thanks{$^{*}$Equal contribution}
}

\begin{document}
\maketitle
\begin{abstract}
AI systems deployed in the real world must contend with distractions and out-of-distribution (OOD) noise that can destabilize their policies and lead to unsafe behavior. While robust training can reduce sensitivity to some forms of noise, it is infeasible to anticipate all possible OOD conditions. To mitigate this issue, we develop an algorithm that leverages a world model’s inherent measure of surprise to reduce the impact of noise in world model–based reinforcement learning agents. We introduce both multi-representation and single-representation rejection sampling, enabling robustness to settings with multiple faulty sensors or a single faulty sensor. While the introduction of noise typically degrades agent performance, we show that our techniques preserve performance relative to baselines under varying types and levels of noise across multiple environments within self-driving simulation domains (CARLA and Safety Gymnasium). Furthermore, we demonstrate that our methods enhance the stability of two state-of-the-art world models with markedly different underlying architectures: Cosmos and DreamerV3. Together, these results highlight the robustness of our approach across world modeling domains. 
We release our code at \url{https://github.com/Bluefin-Tuna/WISER}.
\end{abstract}    
\section{Introduction}
\label{sec:intro}
AI systems that operate in the real world can often encounter noise, distractions, environmental interference, transmission errors, or new semantic classes of objects.
When these occur, AI policies---the part of the AI system that translates from observation to action---can become unpredictable, unreliable, or unsafe. 
When observations contain out-of-distribution noise, the policy may respond inappropriately to the noise, resulting in execution that is unsafe to the agent or to any human operators in the vicinity.

At the heart of the challenge, an AI system cannot know what is noise and what is actionable information in the observation.
While AI system developers may do their best to train systems that are robust to as many types of noise as possible, in real-world settings, it is impossible to anticipate all possible ways in which sensor information may be corrupted.
That is: AI systems can always encounter noise and distractors that are interpreted incorrectly.
Yet, we may be able to design AI systems that degrade gracefully by being able to (a)~identify when parts of their observations are likely to be noise versus actionable information and (b)~learn how to make smarter decisions despite the lack of actionable information. 

In the context of world model-based deep reinforcement learning agents, we present a technique for improving the resilience of an agent when {\em parts} of an observation are likely to be noise and/or completely corrupted. 
We use the agent's world model to evaluate the amount of {\em surprise} in the observational data. 
We apply this to multi-sensor and single-sensor agent settings.
Multi-sensor settings are those in which an agent has multiple sensors to which a policy model responds, or has multiple representations of the same raw observational data.
For example, in automated driving, a policy may integrate inputs from cameras, LiDAR, and radar, or fuse both pixel-level images and semantic maps derived from the same visual stream.
It is likely that the underlying cause of the noise does not affect all sensors or representations equally.
In this setting, the surprise measure 
is used to help identify which sensors are likely to distract the agent and to disable those sensors.

While the above is useful for agents that have multiple sensors, such as self-driving cars, some agents may only have a single sensor.
In the single-sensor setting,
we use the world model's
abnormal reconstruction of the observation as a signal to help avoid making decisions on observations that are likely to be distractions or misinterpreted by the agent.
In both settings, we show that our technique is able to reduce the amount of noise introduced to the world model's internal state.

Although the introduction or removal of noise typically degrades agent performance, we show that our techniques preserve performance relative to baselines under varying types and levels of noise across multiple environments within self-driving simulation domains (CARLA~\cite{Dosovitskiy17} and Safety Gymnasium~\cite{ji2023safety}). Furthermore, we demonstrate that our methods enhance the stability of two state-of-the-art world models with markedly different underlying architectures: Cosmos~\cite{nvidia2025worldsimulationvideofoundation} and DreamerV3~\cite{hafner2025mastering}. Together, these results demonstrate the robustness of our approach across world modeling domains.
\begin{figure}[H]
    \centering
    \includegraphics[width=1\linewidth]{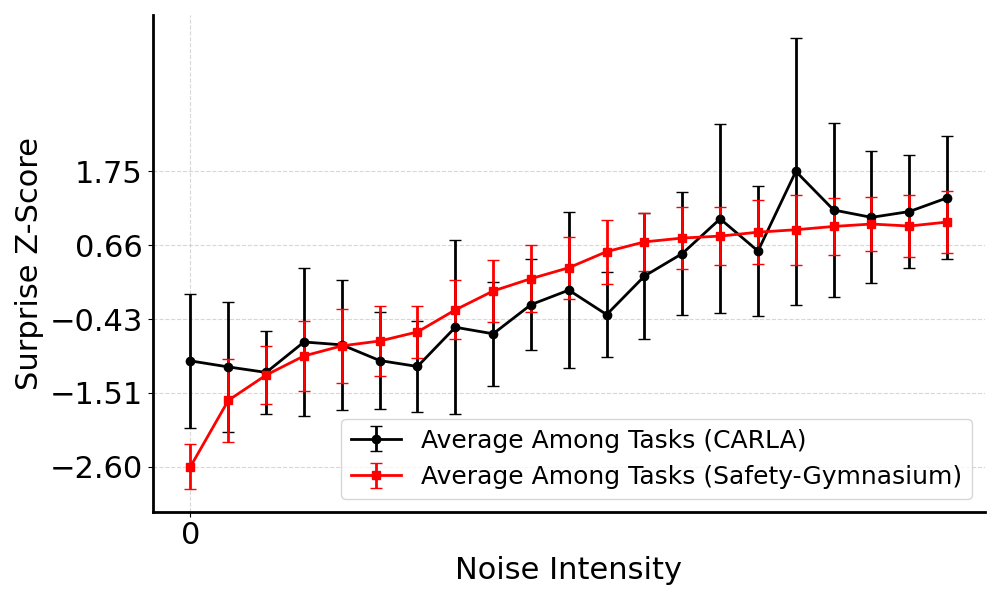}
    \caption{The change in the world model's measure of surprise as noise increases in all tested CARLA~\citep{Dosovitskiy17} and Safety-Gymnasium~\citep{Dosovitskiy17} environments over 15000 sampled steps per intensity. We leverage this insight to understand the degree to which surprise signal can help identify noise given the absence of ground truth.}
    \label{fig:correlation_plot}
\end{figure}

\section{Related Work}
\label{sec:related_work}
Improving the robustness of reinforcement learning (RL) agents against noise capabilities largely remains a difficult challenge~\citep{tuble2022the}. Procedural generation has been employed to improve agent robustness to novel scenarios~\citep{DBLP:journals/corr/abs-1912-01588}, often relying on data augmentation techniques~\citep{wang2024comprehensivesurveydataaugmentation}. However, these approaches can lead to high sample complexity~\citep{10.5555/3535850.3536113}. A further complication is the burden on the policy to learn a wide range of representations for rapid adaptation~\citep{erden2025continualreinforcementlearningautoencoderdriven}.

Recent work has also focused on identifying irrelevant components of the observation space to enhance generalization~\citep{NEURIPS2024_17af4352,zhang2021learning,chang2024offline,DBLP:conf/icml/SunZ0I24}. Such methods often rely on foundation models or Siamese-like architectures during the training process, which can significantly increase training time~\citep{grooten2024madi}. Moreover, these frameworks may require an augmented dataset for initial training on similar OOD situations to the target domain, or are tested in cases where the task-relevant information is always present. Simple Gaussian noise, despite the OOD situation, has been proposed to help models learn OOD scenarios, but might not be effective in certain situations~\citep{sanyal2024accuracy}.

A promising approach for improving model robustness is the use of dropouts across various sensors or multimodal representations~\citep{skand2024simple,pmlr-v78-liu17a,yu2022maskbased,seo2023maskedworldmodelsvisual}. However, these methods typically ignore the issue of determining which sensors to mask out or assume that the agent can already identify which sensors have failed or have become corrupted—a challenging problem in itself~\cite{LI2020111990}. Another limitation is that such dropout techniques often provide limited insight into how accurately an agent can predict actions using only the remaining modalities, and often leave the dropout strategy unchanged in the evaluation domain~\cite{skand2024simple}.

World models~\cite{worldModelIntro} have recently been shown to have some success in out-of-distribution scenarios~\cite{zollicoffer2025novelty} that cannot be taken into account during training. Specifically, results have shown that utilizing the world model's measure of surprise does appear to be a useful heuristic in detecting if an agent's observation is out of distribution. We investigate whether world models can mitigate the issues seen in previous dropout-related frameworks in RL from the perspective of state representation for novel situations. 

Fault Detection, Isolation, Recovery~(FDIR)~\citep{MathWorksFDIR2025,Blanke2016} is a common technique for removing noise, which, at a high level, aims to shut off or mitigate the impact of known faulty sensors. 
For physical systems, such as automotive applications, this typically manifests through the deployment of multiple redundant sensors that measure the same physical state. This redundancy is typically used to enable potential cross-validation among sensors, allowing faulty readings to be detected and suppressed while preserving overall system integrity~\cite{ChenPatton1999}. 

In our work, we do not assume access to redundant sensors, as they would experience the same OOD noisy failures. 
Instead, we assume the potential of having different data per sensor or a single sensor with different representations of the data.
We design an FDIR algorithm centered on world models so the agent can focus on consistent and informative inputs when noise or sensor failures are observed.
\section{Preliminaries}
\label{sec:background}
\paragraph{Partially observable Markov Decision Processes}
We study sequential Partially Observable Markov decision processes (POMDPs) denoted by the tuple $M = (\mathcal{S},\mathcal{A},\mathcal{T}, r,\Omega, O)$, where $\mathcal{S}$ is the state space, $\mathcal{A}$ is the action space, $\mathcal{T}$ is the transition distribution $\mathcal{T}(s_t|s_{t-1},a_{t-1})$, r is the reward function,
$\Omega$ is an emissions model from ground truth states to sensor information, and $O(s',a)$ is the sensory information distribution which emits an observation $x_*\sim O(s',a)$ at each step~\citep{strm1965OptimalCO}.
\paragraph{DreamerV3 World Model}
We conduct experiments using our methods primarily within the DreamerV3 framework~\citep{hafner2025mastering}, which is based on a Recurrent State-Space Model (RSSM). A typical RSSM incorporates both a Variational Autoencoder (VAE) and a Recurrent Neural Network (RNN)~\cite{ha2018worldmodels,worldModelIntro}. DreamerV3 first learns a world model and then uses it to simulate rollouts for training a policy model.
Note that:
\begin{itemize}[noitemsep,topsep=0pt]
    \item $x_*$ is the set of sensory information emitted from the state.
    \item $x_t$ is the \emph{encoded} set of sensory information emitted from the state.
    \item $h_t$ is the encoded history/hidden state of the agent.
    \item $z_t$ is an encoding of the current sensory information $x_t$ and $h_t$ that incorporates the learned dynamics of the world.
    \item $s_t = (z_t,h_t)$ is the agent's compact model state.
    \item At each step, the agent takes an action based on the compact model state (i.e $\pi(s_t) = \pi((z_t,h_t))$)
\end{itemize}
For each representation of state \( s_t \), DreamerV3 defines six additional learned transition distributions—each conditionally independent—based on the trained world model:
\begin{equation}
\label{eq:dv2}
\footnotesize
\text{DreamerV3: } 
\begin{cases}
\text{Sequence model: } 
h_t = f_{\phi}(h_{t-1},z_{t-1},a_{t-1})\\
\text{Representation model: }
q_{\phi}(z_t | h_t, x_t)\\
\text{Dynamics predictor: }
p_{\phi}(\hat{z_t}|h_t)\\
\text{Image prediction model: }
p_{\phi}(\hat{x_t}|h_t,z_t)\\
\text{Reward prediction model: }
p_{\phi}(\hat{r_t}|h_t,z_t)\\
\text{Continue prediction model: }
p_{\phi}(\hat{c_t}|h_t,z_t)
\end{cases}
\end{equation}
where $\phi$ describes the parameter vector for all distributions optimized. The loss function during training~\citep{hafner2025mastering} is:
\begin{align*}
\footnotesize
\mathcal{L}(\phi) \doteq
E_{q_\phi}\Big[\sum_{t=1}^T(&\beta_{\mathrm{pred}}\mathcal{L}_{\mathrm{pred}}(\phi) + \\
    &\beta_{\mathrm{dyn}}\mathcal{L}_{\mathrm{dyn}}(\phi) 
    + \beta_{\mathrm{rep}}\mathcal{L}_{\mathrm{rep}}(\phi)
)\Big].
\label{eq:wm_loss}
\end{align*}
where:
{\small
\begin{align*}
\footnotesize
\mathcal{L}_{\mathrm{pred}}(\phi)
&\doteq
-\ln p_{\phi}(x_t \mid z_t, h_t) \\
&\quad
-\ln p_{\phi}(r_t \mid z_t, h_t)
-\ln p_{\phi}(c_t \mid z_t, h_t) \\[0.4em]
\mathcal{L}_{\mathrm{dyn}}(\phi)
&\doteq
\max\Bigl(
  1,\,
  KL\bigl[
    sg(q_{\phi}(z_t \mid h_t, x_t))
    \,\big\|\, p_{\phi}(z_t \mid h_t)
  \bigr]
\Bigr) \\[0.4em]
\mathcal{L}_{\mathrm{rep}}(\phi)
&\doteq
\max\Bigl(
  1,\,
  KL\bigl[
    q_{\phi}(z_t \mid h_t, x_t)
    \,\big\|\, sg(p_{\phi}(z_t \mid h_t))
  \bigr]
\Bigr)
\end{align*}
}
and $sg(\cdot)$ is a stop gradient operator.

\section{Multiple Sensor Representation Selection}
\label{sec:Multi-Rep-Select}
In this section, we address the multiple-sensor representation setting and describe how a surprise measure on the world model identifies and eliminates sensor representations that are likely to distract the agent with noisy data.
The multiple sensor setting could mean that an agent has multiple sensors that collect different data, such as in the case of an autonomous vehicle with cameras with different views.
It could also mean an agent with a single sensor but representing that sensor data in different ways. 

We assume a {\em world model}, such as that used in DreamerV3~\cite{hafner2025mastering}. 
The world model learns the world transition dynamics and is used to more efficiently train the policy model by predicting state-action-state transitions, effectively separating the agent from the true environment.
Although many of the world model’s learned predictors are rarely used at inference time, we find that the model’s surprise grows proportionally with the intensity of injected noise. This relationship is shown in Figure \ref{fig:correlation_plot} for two domains
The trained world model’s measure of Bayesian surprise~\cite{NIPS2005_0172d289} is the divergence between the world model's learned posterior and prior of the next state: 
\begin{equation}
    KL\bigl[
    q_{\phi}(z_t \mid h_t, x_t)
    \,\big\|\, p_{\phi}(z_t \mid h_t)\bigr]
    \label{eq:bayesian_sup_norm}
\end{equation}
This measure naturally provides a correlated signal for detecting unexpected noise. 
When the model consistently registers unusually high surprise without corresponding structure in the environment, it may indicate that the observations are dominated by noise rather than meaningful dynamics. 
Building on this hypothesis, we introduce a surprise-guided rejection sampling method that mitigates corruption in the world model’s predicted latent state while maintaining an $\mathcal{O}(n \log n)$ computational cost.
during inference time.

\usetikzlibrary{shapes.geometric, arrows.meta, positioning}

\tikzstyle{startstop} = [ellipse, minimum width=3cm, minimum height=1cm, text centered, draw=black, fill=blue!20]
\tikzstyle{process} = [rectangle, minimum width=3cm, minimum height=1cm, text centered, draw=black, fill=purple!20]
\tikzstyle{decision} = [diamond, minimum width=3cm, minimum height=1cm, text centered, draw=black, fill=yellow!20, aspect=2]
\tikzstyle{accept} = [rectangle, rounded corners, minimum width=3cm, minimum height=1cm, text centered, draw=black, fill=green!20]
\tikzstyle{reject} = [rectangle, rounded corners, minimum width=3cm, minimum height=1cm, text centered, draw=black, fill=red!20]
\tikzstyle{arrow} = [thick,->,>=Stealth]

\paragraph{Surprise-Guided Multi-Representation Rejection Sampling for Confident Decision
Making}
We investigate the world model’s ability to \textit{adaptively switch representations} in an online failure setting, where sensor failures occur without prior knowledge. 
In the multi-sensor setting, the agent samples a set of $M$ high-dimensional sensory observations at each time step,
\begin{align*}
\mathbf{x}_* = \{ y_t^{(1)}, y_t^{(2)}, \dots, y_t^{(M)} \},
\end{align*}
from which it forms a fused observation
\begin{align*}
x_t = f_{\text{enc}}(\mathbf{x}_*),
\end{align*}
where $f_{\text{enc}}$ denotes the encoder or sensor-fusion module.
$x_t$ is then used to sample the latent state $z_t$ via the representation model (Eq.~\ref{eq:dv2}).

During evaluation, if any subset of sensory inputs $y_t\in \mathbf{x}_*$ exhibits abnormally high surprise, indicating a potential corruption or failure, the model avoids relying on the full fused observation $x_t$. 
Instead, it samples a latent state prediction $z_t^{(i)}$ conditioned on a fused observation $x_t^{(i)}$ formed by a subset of uncorrupted inputs $\mathbf{x}_*^{(i)} \subseteq \mathbf{x}_*$. 
This selection is formulated as an optimization that chooses the input subset minimizing the Bayesian surprise between the representation model and the dynamics predictor (Eq~\ref{eq:dv2}):
\begin{align}
\label{eq:multi-rep-objective}
\arg\min_{\mathbf{x}_*^{(i)} \subseteq \mathbf{x}_*} 
\mathrm{KL}\!\left[ 
    q_{\phi}(z_t^{(i)} \mid h_t, x_t^{(i)}) 
    \,\|\, 
    p_{\phi}(z_t \mid h_t) 
\right].
\end{align}
The resulting sampled latent $z_t^{(i)}$ serves as an \textit{alternative latent state representation}, which is then used to select an action $a_t \sim \pi((z_t^{(i)}, h_t))$.\footnotemark
\footnotetext{Optionally, the optimization can be relaxed to respect a set of required sensors (see Algorithm~\ref{alg:nonadaptive_masking}).}

\paragraph{Multi-Representation Dropout Training}
To simulate the removal of sensory inputs, we apply a mask-based sensor dropout. Naturally, the absence of expected sensory inputs induces increased surprise within the world model.
Therefore, to prepare the agent for dropout variation in its sensors, we expose it to projections onto random subspaces of the full representation space via random dropout of representations—a form of state representation learning [7]. As shown in Algorithm 1, this encourages the world model to learn a latent distribution that remains robust even when some components of the sensory inputs are missing. By experiencing these partial views during training, we observe (Figure~\ref{fig:dropout_vs_normal_training} Appendix~\ref{sec:ablations}) that the world model is still capable of inferring consistent and informative latent states despite variability in the available representations.

\begin{figure}[]
    \centering
    \includegraphics[width=0.95\linewidth]{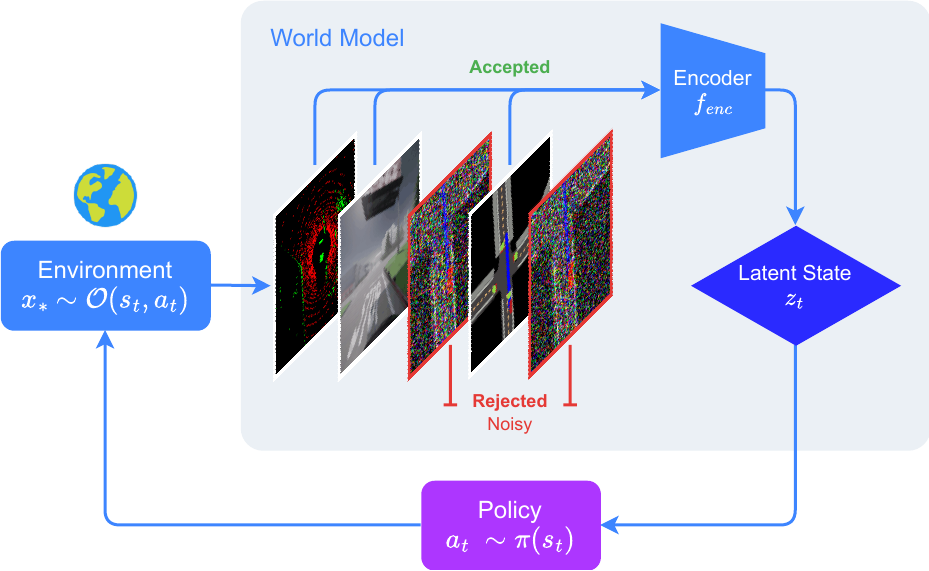}
    
    \caption{Multi-Sensor Rejection Sampling. In practice, we utilize Algorithm~\ref{alg:nonadaptive_masking} to simulate this behavior in the agent.}
    \label{fig:multi-representation_dropout}
\end{figure}

\section{Multiple Sensor Experiments}
\begin{figure*}[]

    \centering

        \begin{subfigure}[t]{0.3\linewidth}
            \centering
            \includegraphics[width=\linewidth]{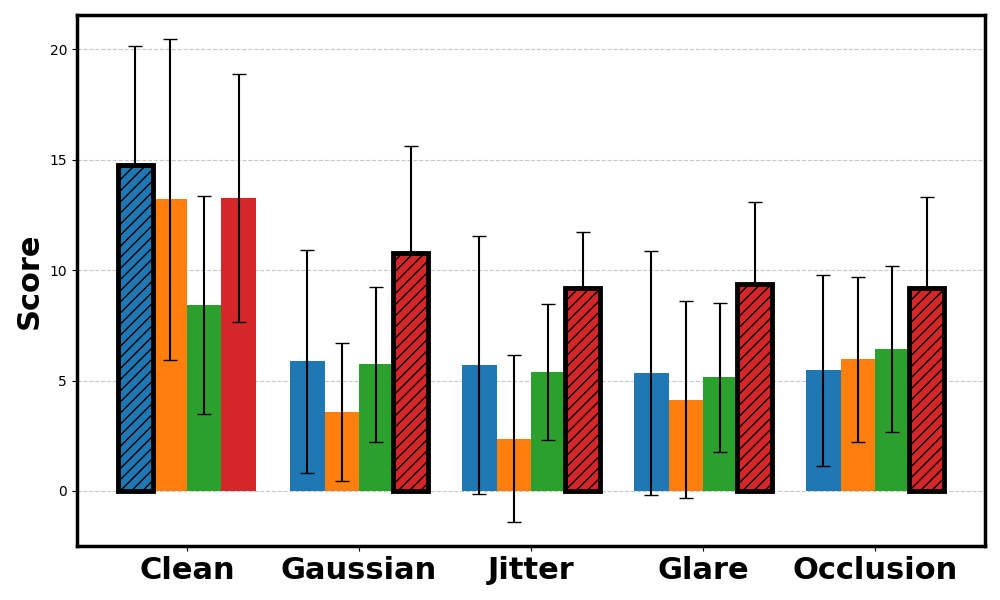}
            \caption{PointGoal Average Score}
        \end{subfigure}
        \hfill
        \begin{subfigure}[t]{0.3\linewidth}
            \centering
            \includegraphics[width=\linewidth]{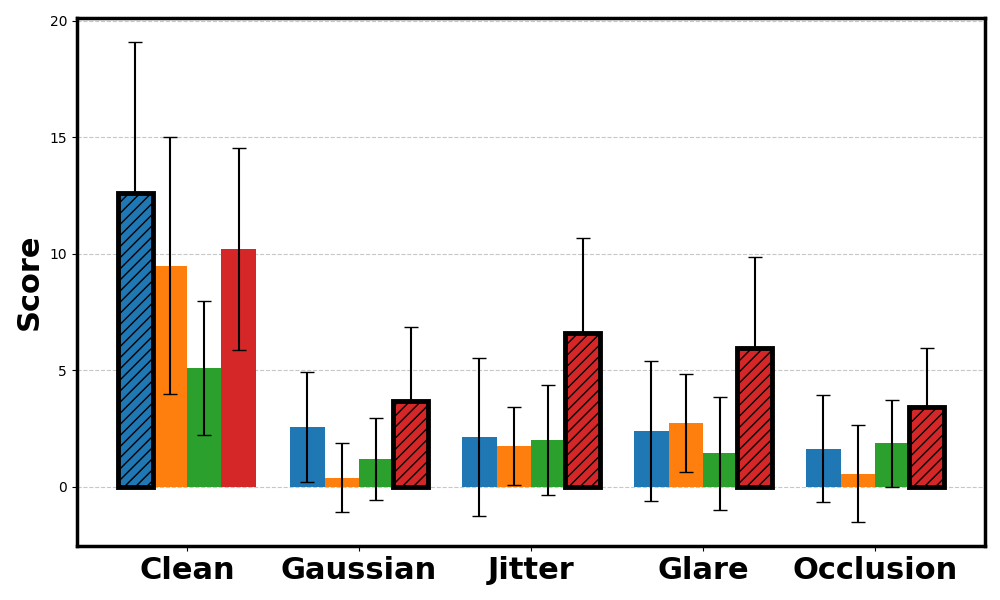}
            \caption{PointButton Average Score}
        \end{subfigure}
        \hfill
        \begin{subfigure}[t]{0.3\linewidth}
            \centering
            \includegraphics[width=\linewidth]{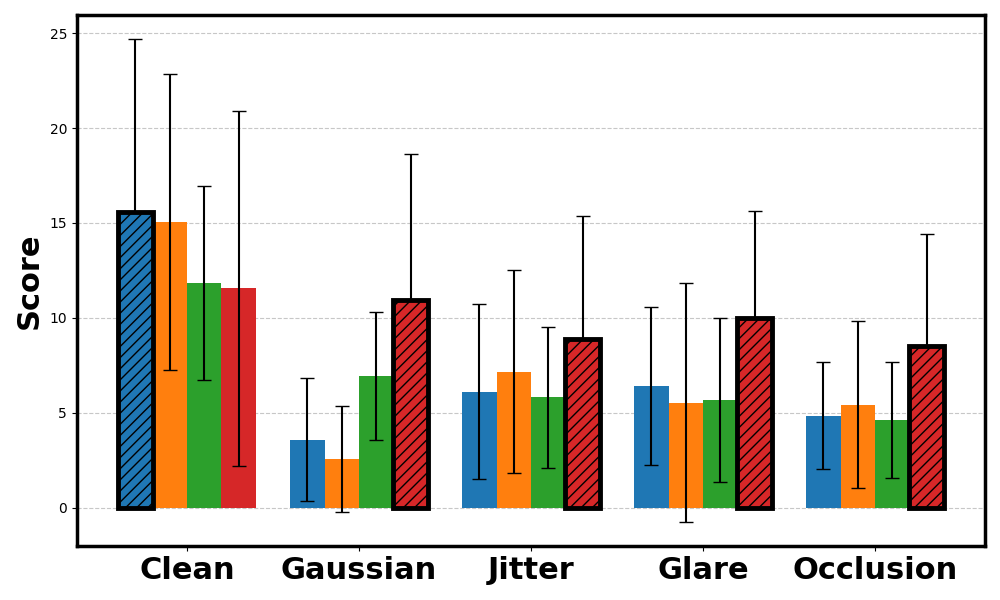}
            \caption{CarGoal Average Score}
        \end{subfigure}

        \vspace{0.5em}

        \begin{subfigure}[t]{0.3\linewidth}
            \centering
            \includegraphics[width=\linewidth]{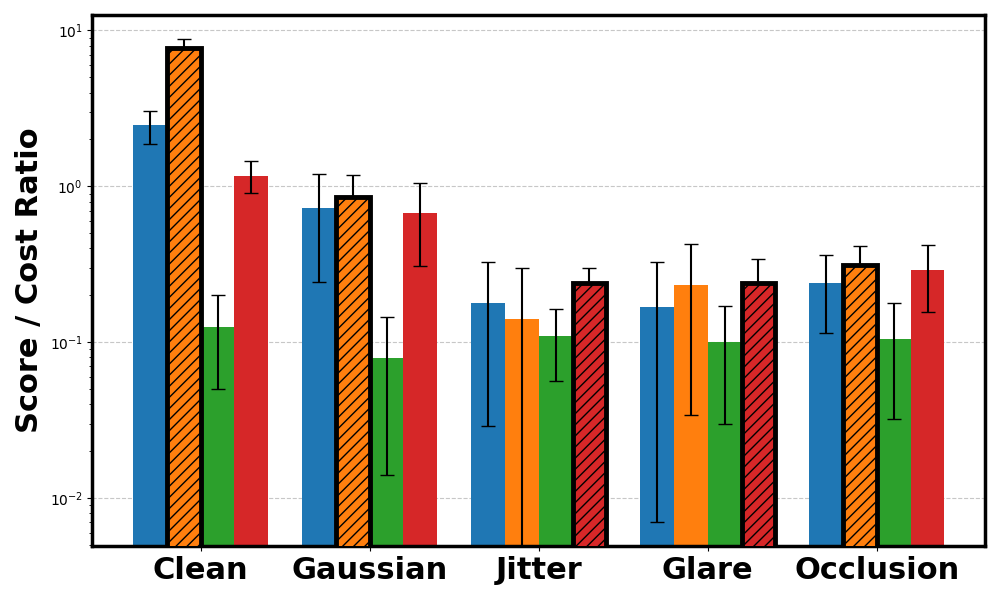}
            \caption{PointGoal Score-Cost Ratio}
        \end{subfigure}
        \hfill
        \begin{subfigure}[t]{0.3\linewidth}
            \centering
            \includegraphics[width=\linewidth]{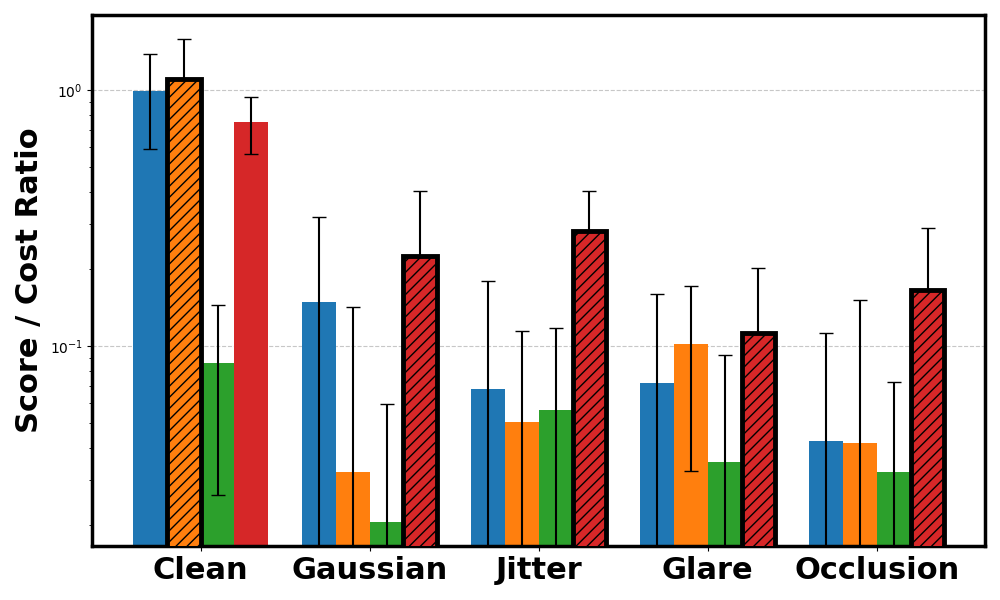}
            \caption{PointButton Score-Cost Ratio}
        \end{subfigure}
        \hfill
        \begin{subfigure}[t]{0.3\linewidth}
            \centering
            \includegraphics[width=\linewidth]{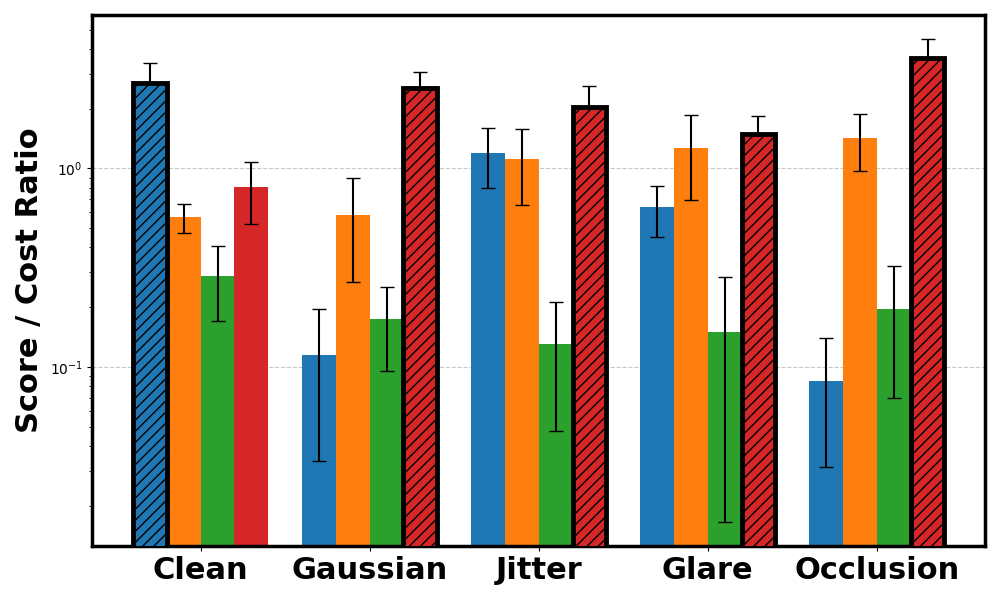}
            \caption{CarGoal Score-Cost Ratio}
        \end{subfigure}
        \begin{subfigure}[t]{\linewidth}
            \centering\includegraphics[width=.7\linewidth]{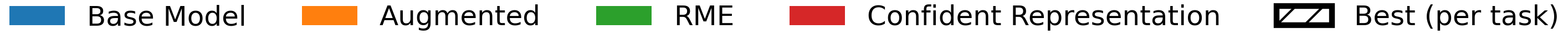}
        \end{subfigure}


    \caption{Performance (top) and Score-Cost ratio (bottom) across three Safety Gymnasium tasks. We display the Score-Cost ratio to measure how safely the task can be accomplished with respect to the score. Each column corresponds to a noise type, with clean being the nominal setting.}
    \label{fig:safetygymnasium}
\end{figure*}
For our main results, we primarily focus on settings that naturally involve multiple sensors, such as automated driving tasks. Unless stated otherwise, we use the default sensors and reward functions provided by each environment for each respective task.
 
\paragraph{Environments}
We show the generalization of our proposed method through experiments conducted primarily in 
the following domains:
\begin{itemize}
\item \textbf{Safety Gymnasium}~\citep{ji2023safety} is a benchmark suite designed to evaluate the trade-off between performance and safety in reinforcement learning, extending the traditional RL paradigm by introducing environments where agents must complete goal-oriented tasks while simultaneously minimizing safety violations such as collisions or entering hazardous regions. 
The environments include a variety of robotic control and navigation tasks that simulate real-world constraints, providing explicit cost signals associated with unsafe behaviors.
Our experiments use the default camera sensors given by the environment (see Appendix~\ref{app:representations} for details).
Safety Gymnasium and the SafeDreamer~\citep{safedreamer} baseline assume that the training and evaluation environments are identical. In practice, this assumption may not hold, and our experiments test this by injecting out-of-distribution data..
\item \textbf{CARLA}~~\cite{Dosovitskiy17} is an open‑source urban driving simulator. It was developed as a free platform for autonomous driving research. Alongside the core simulator code and APIs, CARLA includes a library of custom-designed urban environments and digital assets (e.g., buildings, vehicles, pedestrians) that are freely available for use. The simulator supports flexible configuration of sensor suites, such as RGB cameras, Bird Eye view, LiDAR, GPS, and collision detectors.
\end{itemize}
We list each utilized sensor in detail in Appendix~\ref{app:representations}. 
In each environment, we simulate sensor failures by injecting various types of structured and unstructured out-of-distribution noise into the sensor data. Examples include gaussian, occlusion, glare, jitter, chromatic aberration (chrome), and latency. We provide example visualizations of noises applied to the CARLA Bird Eye View (BEV) sensor in~\ref{fig:carla_noise_vertical}. Additionally, all noises are described in Appendix~\ref{app:noises}.
The agent is never notified of any properties of the failure scenario. 
In all experiments, there is always at least one sensor that is unaffected.
\paragraph{Agent Models}
In this setting, we focus our experiments on the State-of-the-Art DreamerV3 world model~\cite{hafner2025mastering}.
For each experiment, the {\em Base} agent is a model trained normally. An {\em Augmented} agent is the base model trained on a dataset augmented with Gaussian noise. The {\em Random Masked Encoder} (RME)~\cite{skand2024simple}, is a technique that also employs masked based sensor dropout, however does not provide a means to decide which sensors should be masked out.
For the multi-sensor domain, our rejection sampling technique is labeled as {\em Confident Representation} (as to not be confused with single-sensor results), which consists of agents equipped with the surprise mechanism, dropout training, and multi-sensor selection technique described in Section~\ref{sec:Multi-Rep-Select}. For all experiments during inference time, we deploy Algorithm~\ref{alg:nonadaptive_masking}, as our proposed multi-sensor selection algorithm, designed to run in $O(nlogn)$ without parallel optimizations.
Unless otherwise specified, each experiment data point represents an accumulation of 15,000 steps across multiple independent and identically distributed environments for a particular method within the given task and noise scenario. We evaluate each method in a fully closed-loop setting, where the agent's actions directly influence subsequent observations. For both Safety Gymnasium and CARLA, we focus on the agent's reward in all tasks. In addition, for the Safety Gymnasium domain, we also focus on cost, an empirical safety score that the Safety Gymnasium domain provides.

\begin{figure*}
    \centering
    \begin{adjustbox}{minipage={0.95\linewidth},fbox}
        \centering
        \textbf{Carla: Stop Sign Task} \\[4pt]
        \includegraphics[width=1\linewidth]{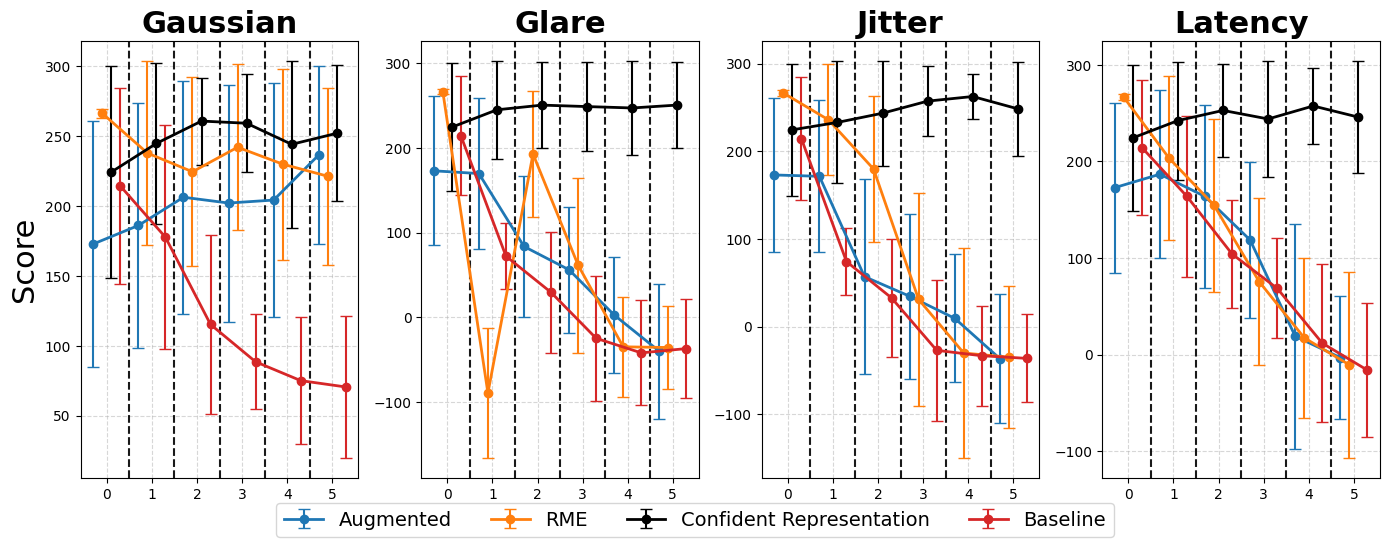}
    \end{adjustbox}

    \caption{Agent performance as the number of sensor failures (from 0-5 sensors) during the stop sign task. We observe that as the sensors begin to become affected, we are able to reduce the effect that noise has on the main policy. For all 72 tested settings over the Stop Sign, Right Turn, and Four Lane Driving Tasks, see Appendix~\ref{fig:All_Coarse_Carla}.}
    \label{fig:Carla}
\end{figure*}

\paragraph{Results}
Figure~\ref{fig:safetygymnasium} illustrates our results on the Safety Gymnasium environment. Our Confident Representation method is capable of consistently achieving the highest level of robustness on a variety of simulated sensor failures.
It is also able to capable of accomplishing said tasks with a competitive score to cost ratio compared to other methods tested in failure scenarios. 

In CARLA, we can have up to 5 sensor failures. 
Results are shown in Figure~\ref{fig:Carla}. 
We observe that our Confident Representation method is capable of identifying failed sensors to shut off during inference time. In comparison to other methods, as sensors begin to fail, the agent is able to identify and use representations that provide meaningful signals and achieve high rewards.

Also, despite not being trained on the exact Gaussian noise that the Gaussian failure was trained on, Confident Representation selection is capable of mirroring the performance of the Augmented baseline in Figure~\ref{fig:Carla}. 
Finally, we compare our $O(n\log n)$ sensor selection technique (Algorithm~\ref{alg:nonadaptive_masking} in Appendix~\ref{sec:algorithms}) to an exhaustive search of sensors and show in Figure~\ref{fig:2nComparisonFull}  that our technique can expect similar performance despite being magnitudes lower in computational complexity. 

\section{Single Sensor Representation Selection}

We next consider a setting where the agent samples a single high-dimensional observation \(x_* \in \mathcal{X}\) at each decision step from the environment, which may vary from a clean, informative signal to a heavily corrupted input with limited recoverable information. 
As with the previous multi-sensor setting, the OOD noise resulting from potential sensor failure may cause the agent to take the wrong action, or may have no effect on action choice.

To design an algorithm capable of improving robustness, we first categorize the possible forms of \(x_*\) into three cases:
\begin{enumerate}
    \item \(x_* = x_t\): The observation is clean, and the agent can directly encode and rely on it for decision-making. 
    \item \(x_* = x_{light}\): The observation contains some noise, but the underlying informative signal can still be recovered or emphasized through selective processing or denoising.
    \item \(x_* = x_{heavy}\): The observation is dominated by noise to the extent that the true signal is no longer recoverable and should therefore be ignored.
\end{enumerate}
Given the possible classes of \(x_*\), we enable the agent to switch between two modalities during inference: a \textit{predictive mode}, wherein the agent rejects \(x_{heavy}\) and the agent instead relies on its internally predicted next state from the world model, and a \textit{ground-truth mode}, wherein the agent accepts the sampled observation \(x_t\) or its denoised version \(x_{light} = D(x_*)\), for some choice of denoiser $D(\cdot)$. 
In essence, when the agent determines that the input data is sufficiently corrupted such that it jeopardizes action selection, it prefers its internal predictions of the world dynamics over what it observes.

To decide if a observation $x_*$ should be rejected, a rejection score is assigned by a function, $M(x_*)$, and is rejected if it reaches a threshold $\tau$. 
For our experiments we employ $M(\cdot)$ as an expected reconstruction loss of the unconditioned ($h_t=h_0$) observations: $\frac{1}{|N|}
   \sum_{i \in N}
   \big|\, x_t^{(i)} - \hat{x}_t^{(i)} \,\big|$ 
where $\hat{x}_t = 
   \mathbb{E}_{p_{\phi}(x_t \mid z_t, h_0)}[x_t]$, and employ $D(\cdot)$ to be the predicted posterior representation generated by the world model variational autoencoder: $\mathbb{E}_{p_{\phi}(x_t \mid z_t, h_t)}[x_t]$, as proof of concept to identify abnormal sensory input. 
Intuitively, \(M(\cdot)\) and \(\tau\) are defined to indicate the presence of a failed sensor, signaling that directly processing \(x_*\) would likely lead to unpredictable behavior. The objective of our rejection sampling mechanism is to enable the agent to produce \emph{predictable and conservative responses} in the presence of unknown sensor failures. Since the purpose of rejection sampling is to preserve the predicted latent state from corruption, downstream policies can then implement conservative strategies—such as pulling over—minimizing the risk of unpredictable behavior occurring during their execution.
\paragraph{Aligning Context} 
During predictive mode, the agent relies on its own imagined or simulated dynamics, which can gradually drift from the true environment state. 
The agent transitions from predictive mode to ground-truth mode when an observation is accepted as being $x_t$ or $x_n$.
When the agent has been in predictive mode for a period, it becomes necessary to reconcile the incoming sensory input with the agent’s internal context or trajectory $h_t$. 
Upon re-entering ground-truth mode, the agent performs a \textit{context reset}---realigning its latent representation or belief state to ensure consistency between the accepted observation \(x_*\) and its previously predicted trajectory. This step prevents discontinuities in state estimation and allows the agent to resume grounded interaction with the environment.

\begin{figure}[t]
    \centering
    \includegraphics[width=.9\linewidth]{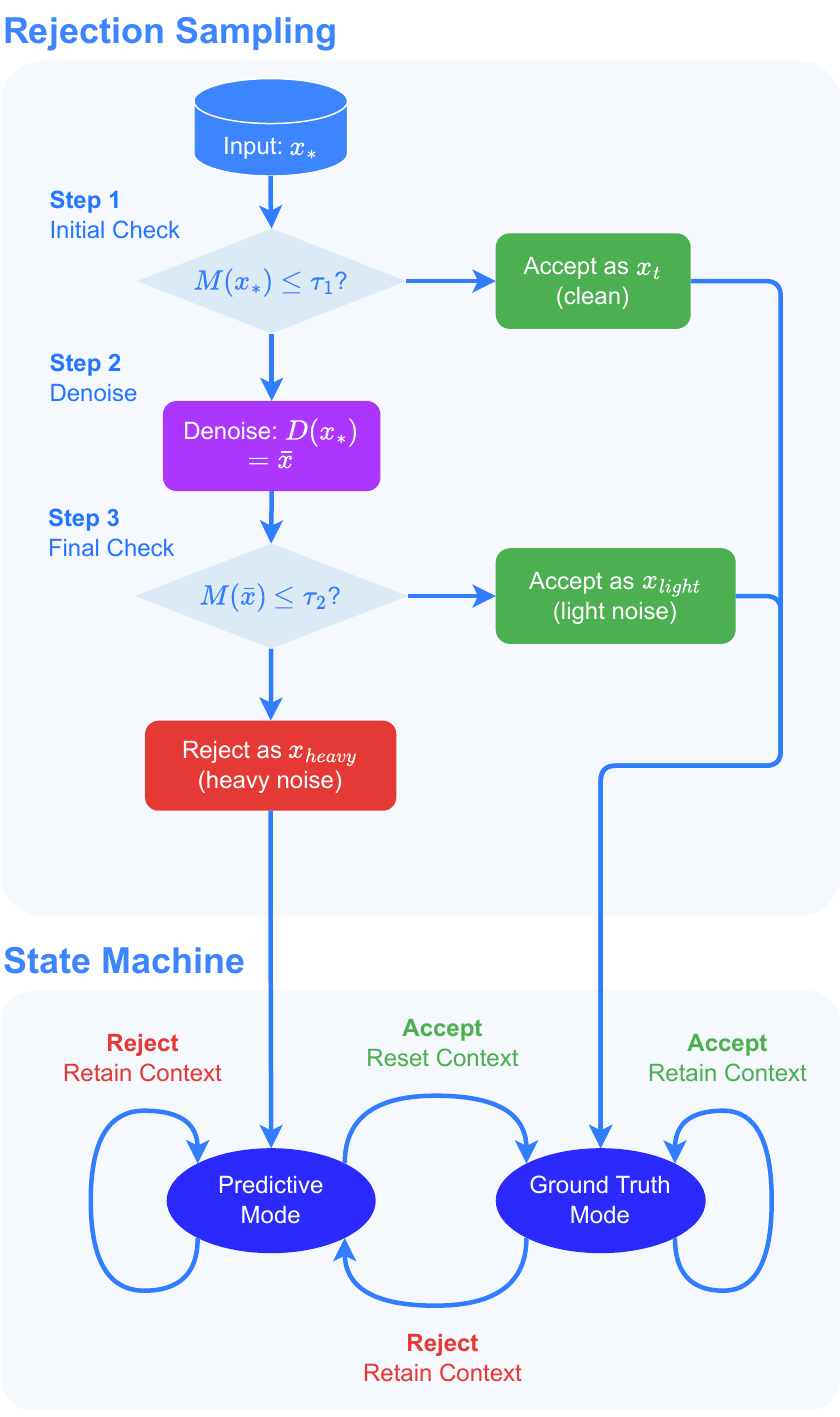}
    \caption{Rejection sampling process for noise classification and context state machine for the world model's latent state.}
    \label{fig:rejection_sampling}
\end{figure}

Conversely, when an observation is rejected after operating in ground-truth mode, the agent retains its most recent valid internal context as a persistent residual. This residual serves as a stabilizing prior, allowing the agent to maintain coherent belief evolution during predictive inference until a new, trustworthy observation is accepted. In effect, the agent alternates between correcting drift upon acceptance and preserving continuity upon rejection, ensuring robust operation across varying observation qualities. We illustrate this interleaving of switching between modes in Figure~\ref{fig:rejection_sampling}.


\section{Single Sensor Experiments}
 We focus our experiments on two types of world models: a VAE-based model (DreamerV3)~\citep{hafner2025mastering} and a diffusion-based model (Cosmos Predict-2.5)~\citep{nvidia2025worldsimulationvideofoundation}.
For unmodified components of the world models, we use the exact hyperparameters recommended in \cite{hafner2025mastering}, \cite{safedreamer}, \cite{nvidia2025worldsimulationvideofoundation}, and \cite{10714437}. For all other additional modifications, we list exact hyperparameters used and hardware details in Appendix~\ref{app:hardware}.


\begin{figure*}[t]
    \centering
    \begin{subfigure}{0.40\textwidth}
        \centering
        \includegraphics[width=\linewidth]{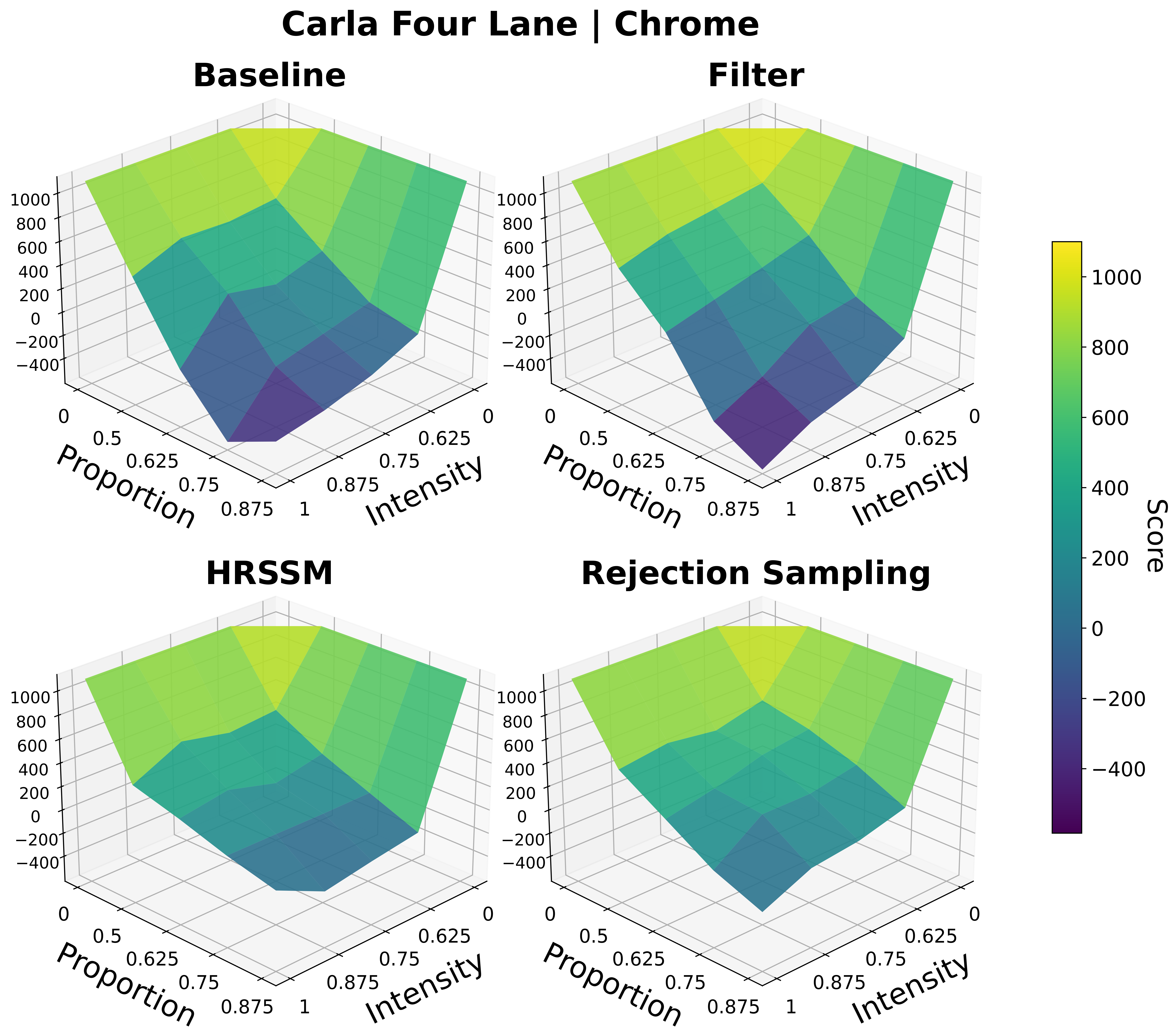}
        \caption{Chromatic Aberration corruption}
    \end{subfigure}
    \hfill
    \begin{subfigure}{0.40\textwidth}
        \centering
        \includegraphics[width=\linewidth]{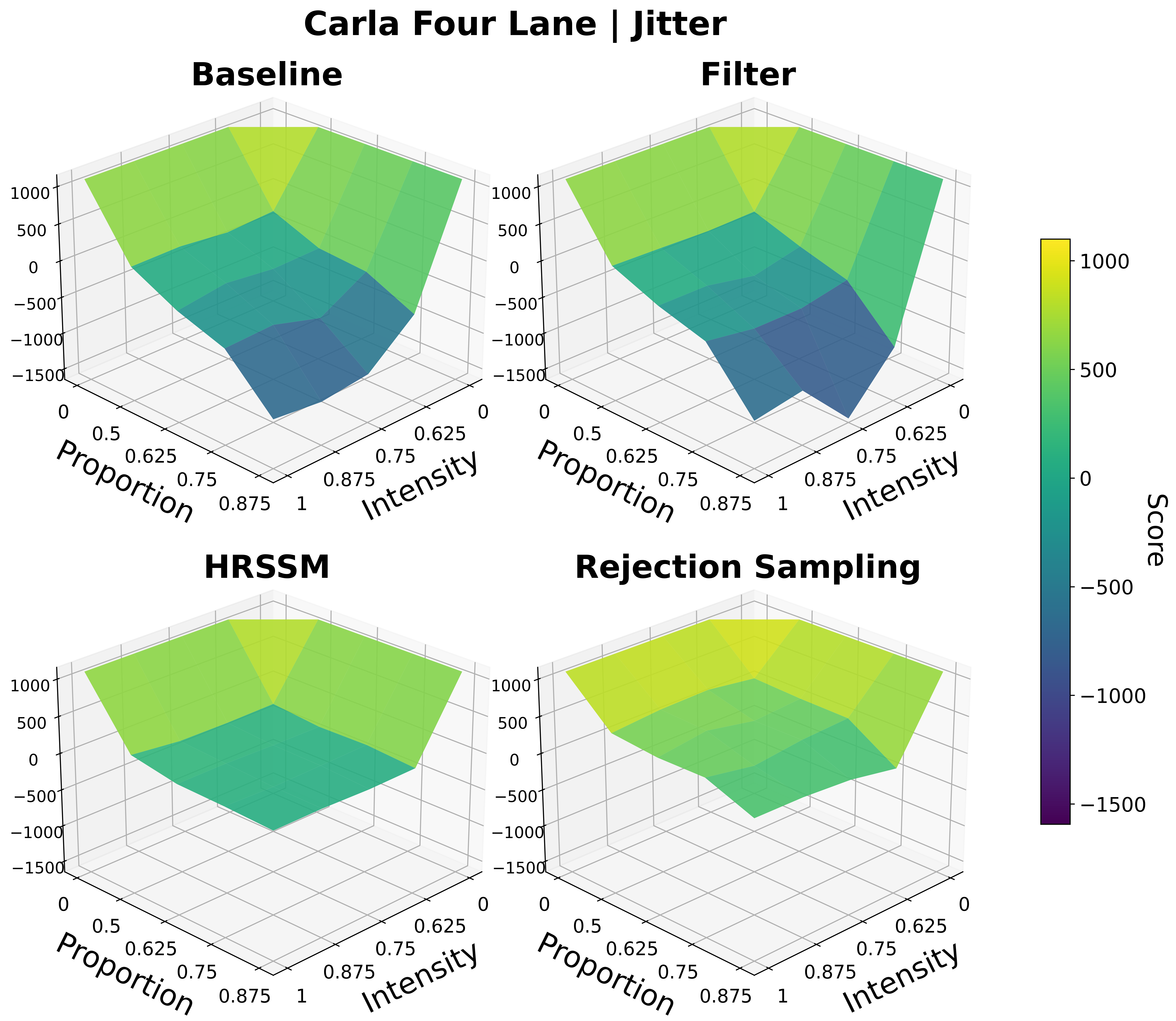}
        \caption{Jitter corruption}
    \end{subfigure}

    \vspace{0.5em}

    \begin{subfigure}{0.40\textwidth}
        \centering
        \includegraphics[width=\linewidth]{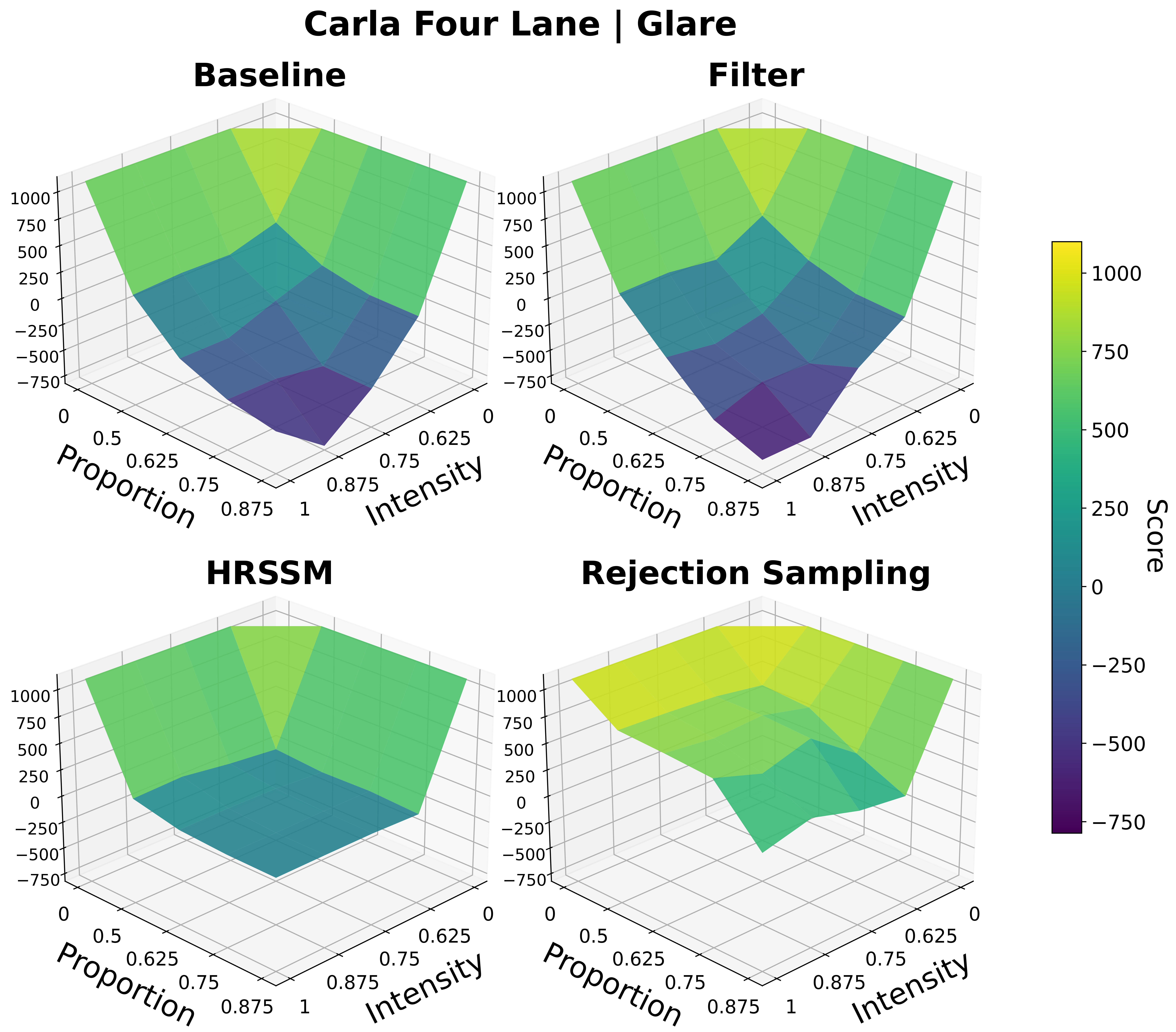}
        \caption{Glare corruption}
    \end{subfigure}
    \hfill
    \begin{subfigure}{0.40\textwidth}
        \centering
        \includegraphics[width=\linewidth]{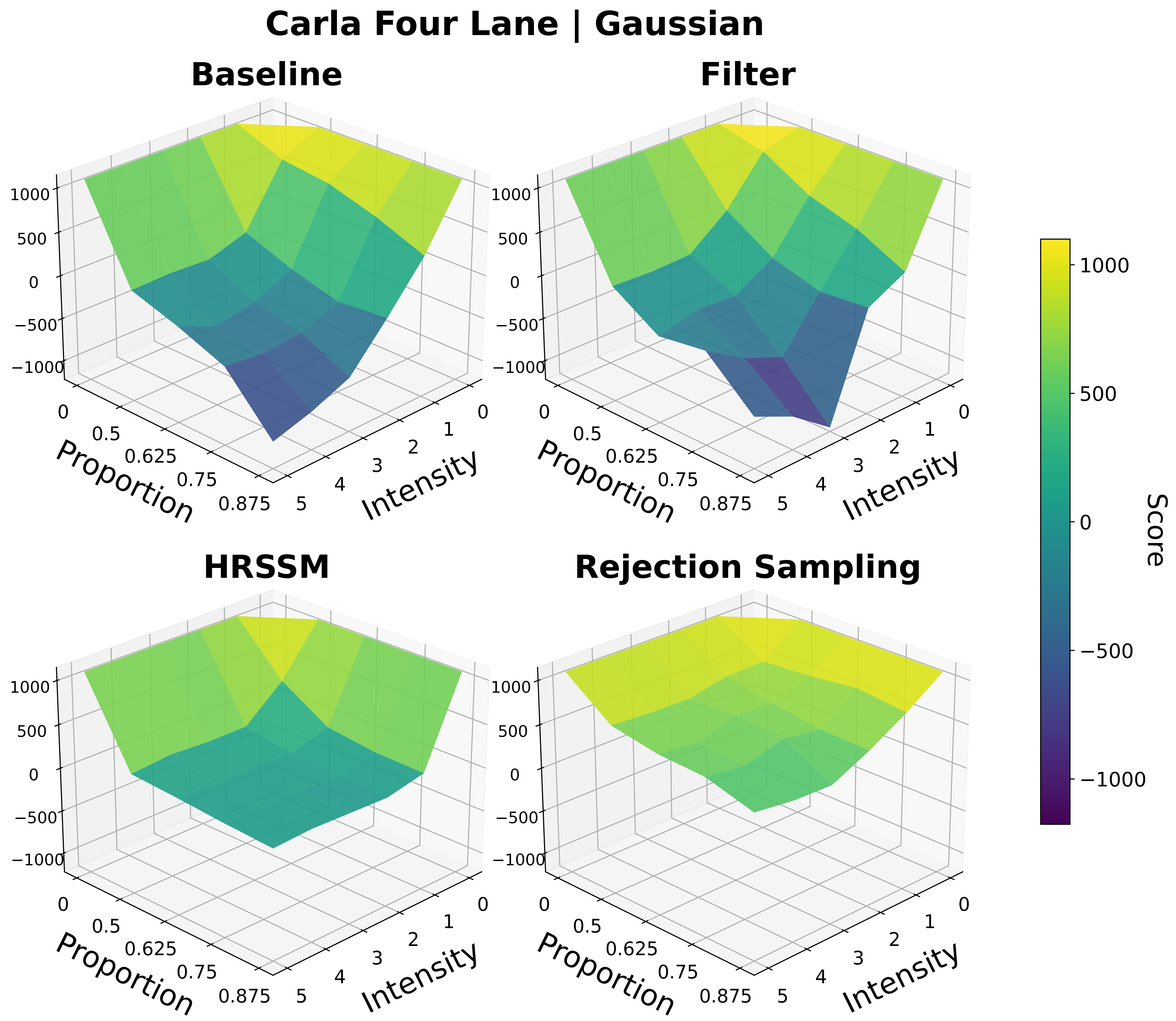}
        \caption{Gaussian noise}
    \end{subfigure}

    \caption{
        Visualization of policy behavior in the CARLA four-lane environment under different visual corruptions, accounting for 192 settings. Each subfigure shows the methods’ potential to safeguard the agent from a specific perturbation type. For results on other tasks see~\ref{fig:carla_task_corruptions}.
    }
    \label{fig:carla_four_lane_single_rep_scores}
\end{figure*}

In these experiments, the world model-based agent uses a single sensor and only receives a single representation at every time step. 
In the single representation setting, we consider the effect that different combinations of noise, intensity, and proportion have on an agent during the course of an episode. To empirically test the behavior of our proposed method, 
We empirically validate our results against the original world model (\emph{Base}), Median Filtering (\emph{Filter})~\cite{wang2024comprehensivesurveydataaugmentation}, and Hybrid RSSM (\emph{HRSSM})~\cite{DBLP:conf/icml/SunZ0I24}, which augments the RSSM framework to capture task-relevant dynamics while suppressing noise. We set $\tau$ to be 5 standard deviations from the average reconstruction loss during transitions in the normal environment. Figure~\ref{fig:carla_four_lane_single_rep_scores} shows results from the Carla Four Lane task. Rejection Sampling consistently achieves the highest scores across all tested noise types, maintaining robustness as both the proportion and intensity of noise increase. Unlike other methods, it does not exhibit a decline toward lower negative scores under higher noise levels. We find that across all methods, the performance surfaces are generally skewed to the right, suggesting that increasing the proportion of corrupted observations has a stronger detrimental effect than increasing noise intensity.
\subsection{Cosmos World Model}
\begin{table}[t]
\centering
\caption{Average PAI Quality score~\citep{PAIBench2025}, between the base model and our proposed Rejection Sampling technique. Separated by noise type.}
\label{tab:cosmos_overall_scores}
\resizebox{\columnwidth}{!}{%
\begin{tabular}{lcccc}
\toprule
\textbf{Aug Type} & \textbf{Base Model} & \textbf{Rejection Sampling} & \textbf{Avg Diff} & \textbf{Rel \%} \\
\midrule
Chrome     & 0.774 & 0.808 & 0.034 & 3.13  \\
Gaussian   & 0.767 & 0.810 & 0.043 & 5.00  \\
Glare      & 0.726 & 0.809 & 0.083 & 11.98  \\
Jitter     & 0.719 & 0.812 & 0.093 & 12.25  \\
Occlusion  & 0.787 & 0.810 & 0.023 & 3.18  \\
\midrule
\textbf{Overall} & \textbf{0.755} & \textbf{0.810} & \textbf{0.055} & \textbf{7.11} \\
\bottomrule
\end{tabular}%
}
\end{table}
To further verify that the proposed rejection-based filtering mechanism generalizes beyond latent-state world models, we extend our study to the Cosmos Predict 2.5 world model---a Diffusion-based video world model that performs pixel-space prediction through large-scale diffusion and spatio-temporal self-attention. 

We experiment with utilizing the \emph{Rejection Sampling} process displayed in Figure~\ref{fig:rejection_sampling} in an alternative world model setting. We utilize each of the noises discussed in Appendix~\ref{app:noises} at a proportion of 75\% to distort input videos, and monitor Cosmos-Predict 2.5's generation quality. In this setting, we apply our proposed Rejection Sampling process on the Robot Pouring Task (See Appendix~\ref{app:cosmos} for prompt, visualization, and implementation details), and compare with the base Cosmos generation procedure. 

For comparison with the standard Cosmos generation procedure, we monitor the PAI-Bench quality score~\citep{PAIBench2025}: An overall summary of eight scores pertaining to quality, consistency, and smoothness (Explicit descriptions can be found in Appendix ~\ref{app:cosmos_eval}). 
We define $M(x_*)$ as the metric score measuring reconstruction quality:
\begin{equation*}
M(x_*) = \frac{1}{CHW}\sum_{c,h,w}(x_*^{c,h,w} - \hat{x}_*^{c,h,w})^2
\end{equation*}
where $\hat{x}_* = f_\theta(x_*)$ is the immediate next frame generated by the model from conditioning only on the previous frame $x_*$, and $C$, $H$, $W$ denotes the channel, height, and width dimensions. We evaluate each of the last \( N \) input frames \( x_* \) by computing their corresponding rejection score \( M(x_*) \) (See Figure~\ref{fig:cosmos_robot_pouring_chrome2} for a visual aid). Similar to denoising methods found in unsupervised settings~\cite{nie2022diffusionmodelsadversarialpurification,Zollicoffer_Vu_Nebgen_Castorena_Alexandrov_Bhattarai_2025,10378437}, we employ $D_\sigma(x_*)$ as a noise-and-denoise operator that: (1) encodes reference frame $x_*$ to latent space, (2) corrupts the latent with Gaussian noise at strength $\sigma$, and (3) applies reverse diffusion denoising for reconstruction. The reverse diffusion process leverages the model's learned Gaussian noise characteristics to remove the corruption and recover the true image.
We report our findings in Table~\ref{tab:cosmos_overall_scores}. We find that Rejection Sampling is capable of improving the relative PAI quality score by roughly 7\% overall when the input video is corrupted. Additionally, we find that rejection sampling brings all augmentation types to a similar improved performance level ($\sim0.81$), regardless of their initial degradation severity.

\section{Conclusion}
We introduced a surprise-driven filtering framework that leverages the world model’s internal uncertainty to selectively suppress unreliable sensory inputs. Our approach introduces two variants of rejection sampling—multi-sensor and single-representation—that enable world models to remain stable and effective under diverse noisy, out-of-distribution sensor failures. Empirical evaluations across self-driving tasks in CARLA and Safety Gymnasium, as well as across both diffusion-based and VAE-based world models, demonstrate that our method achieves robust performance without compromising adaptability. These results highlight the potential of surprise-based mechanisms for building world-model-based systems that maintain safety and stability in the face of unknown failures.
\clearpage
{
    \small
    \bibliographystyle{ieeenat_fullname}
    \bibliography{main}
}
\clearpage
\newpage
\appendix

\section{Sensors}
\label{app:representations}

\paragraph{CARLA~\cite{Dosovitskiy17}}
\label{app:carla_representations}
\begin{itemize}
    \item \textit{Lidar}: Captures a 3D point cloud of the surrounding environment.
    \item \textit{Camera}: Provides a forward-facing RGB view from the vehicle.
    \item \textit{Bird's-Eye View (BEV) Raw}: Displays the roadmap and surrounding vehicles from a top-down perspective.
    \item \textit{Collision}: Detects collisions with objects in the environment.
    \item \textit{Bird's-Eye View with Traffic Signals}: Renders traffic lights and signs directly on the BEV.
    \item \textit{Bird's-Eye View with Waypoints}: Renders only waypoints on the BEV (GPS setting).
    \item \textit{Bird's-Eye View Full}: Renders all available semantic information on the BEV.
\end{itemize}
\paragraph{Safety Gymnasium~\cite{ji2023safety}}
\label{app:safe_representations}
\begin{itemize}
    \item \textit{Bird Eye View}: A view of the agent and the entire map from above.
    \item \textit{Front Vehicle View}: The immediate front view of the agent.
    \item \textit{Dash Camera View}: The immediate front view of the agent from the dash cam perspective.
\end{itemize}

\section{CARLA Noise Corruption and Sensor Degradation}

\subsection{Environment and Representations}

In CARLA \citep{Dosovitskiy17}, an open-source urban driving simulator, the agent is equipped with a diverse sensor suite comprising seven modalities (see~\ref{app:carla_representations}). We evaluate performance on three representative scenarios—Stop Sign, Right Turn, and Four Lane Driving—which test the agent’s ability to perceive, plan, and act under different traffic configurations.

\subsection{Robustness under Visual Corruptions}
\label{app:noises}
We evaluate the robustness of the driving agent in the CARLA simulator by subjecting input images to five types of visual corruptions:

\begin{itemize}
    \item Chromatic Aberration: introduces small spatial shifts between RGB channels, producing color fringing and edge distortions similar to lens dispersion.
    \item Gaussian Noise: adds pixel-level random variations, modeling low-light sensor interference.
    \item Glare: causes extreme overexposure that whitens the whole frame, erasing most visual information.
    \item Jitter: applies random shifts in contrast or brightness, simulating unstable or high-noise sensor conditions.
    \item Occlusion: masks random regions of the frame, simulating partial blockage of the camera view.
\end{itemize}

\begin{figure*}[!htbp]
    \centering

    \begin{subfigure}{0.19\linewidth}
        \centering
        \includegraphics[width=\linewidth]{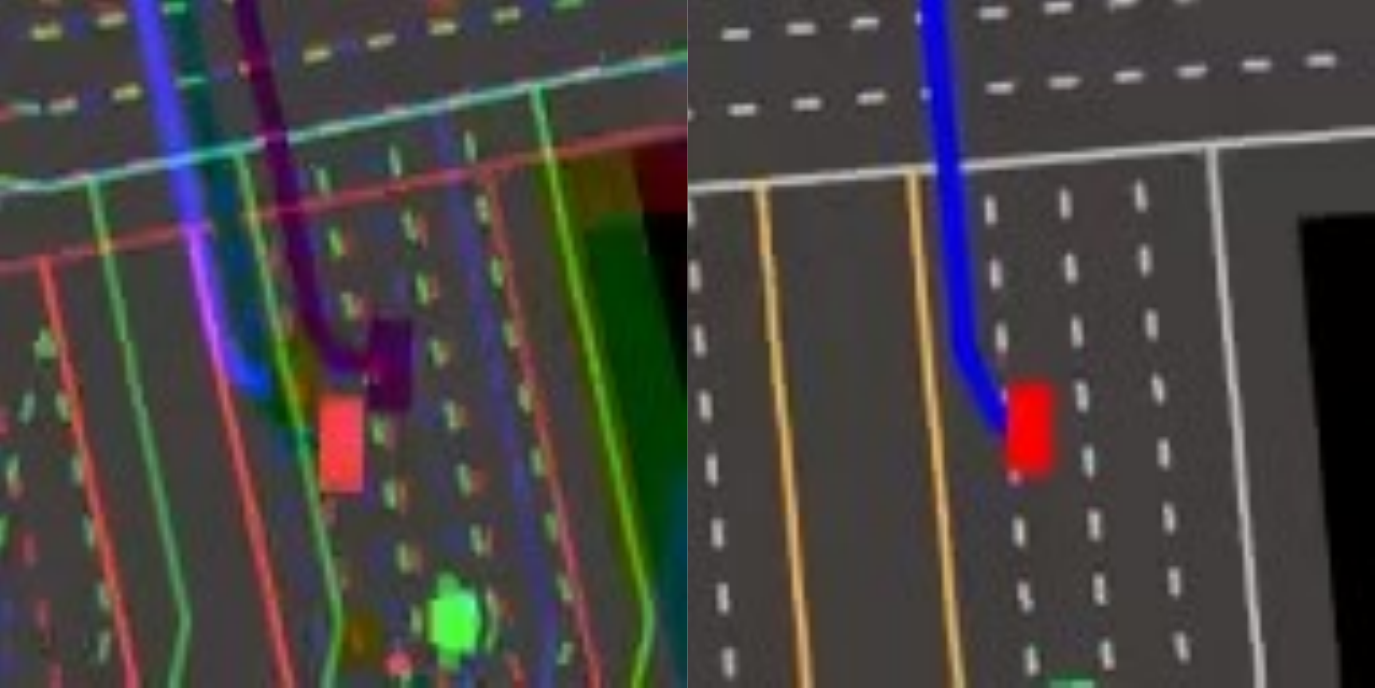}
        \caption{Chrome}
    \end{subfigure}
    \hfill
    \begin{subfigure}{0.19\linewidth}
        \centering
        \includegraphics[width=\linewidth]{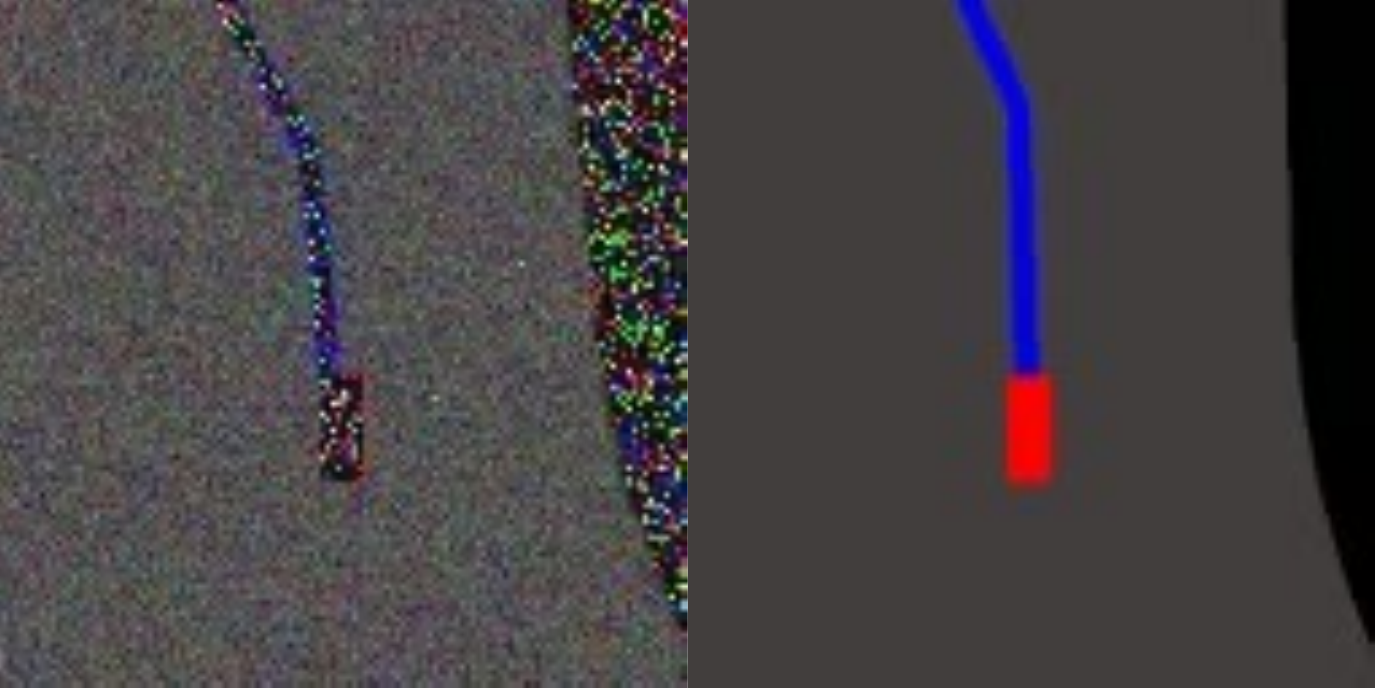}
        \caption{Gaussian}
    \end{subfigure}
    \hfill
    \begin{subfigure}{0.19\linewidth}
        \centering
        \includegraphics[width=\linewidth]{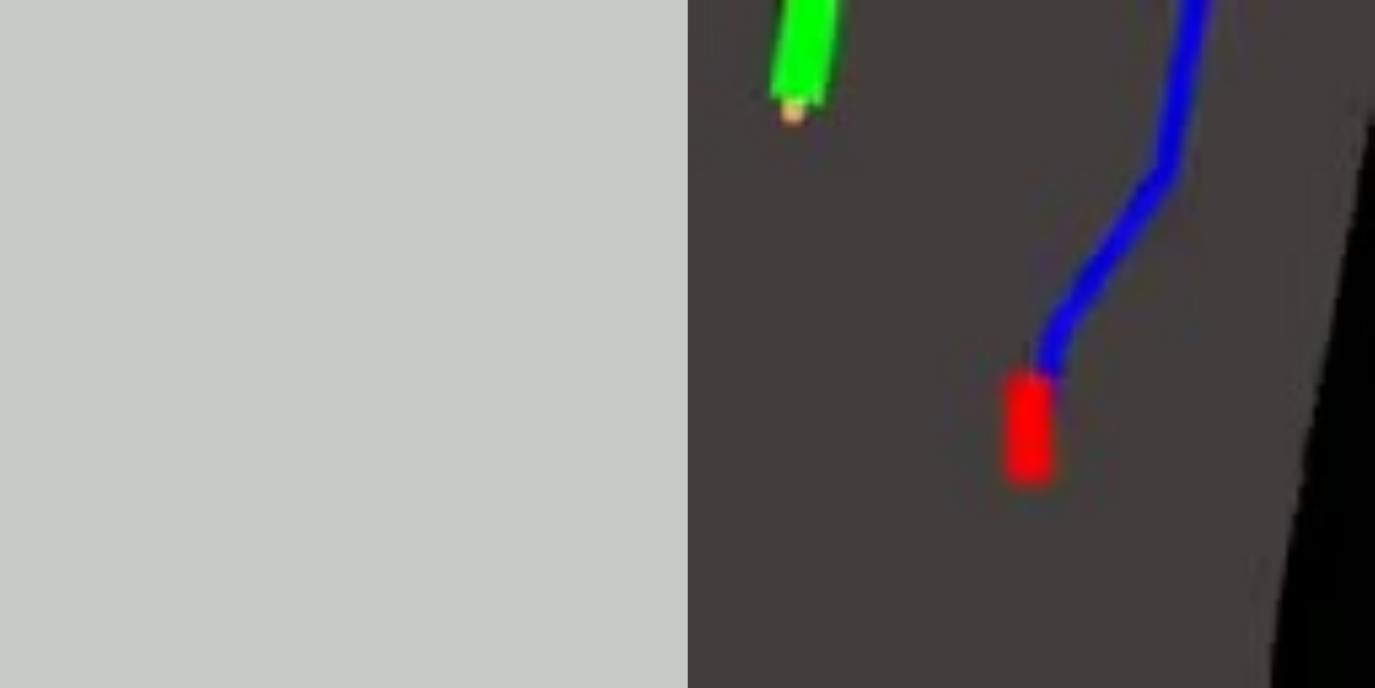}
        \caption{Glare}
    \end{subfigure}
    \hfill
    \begin{subfigure}{0.19\linewidth}
        \centering
        \includegraphics[width=\linewidth]{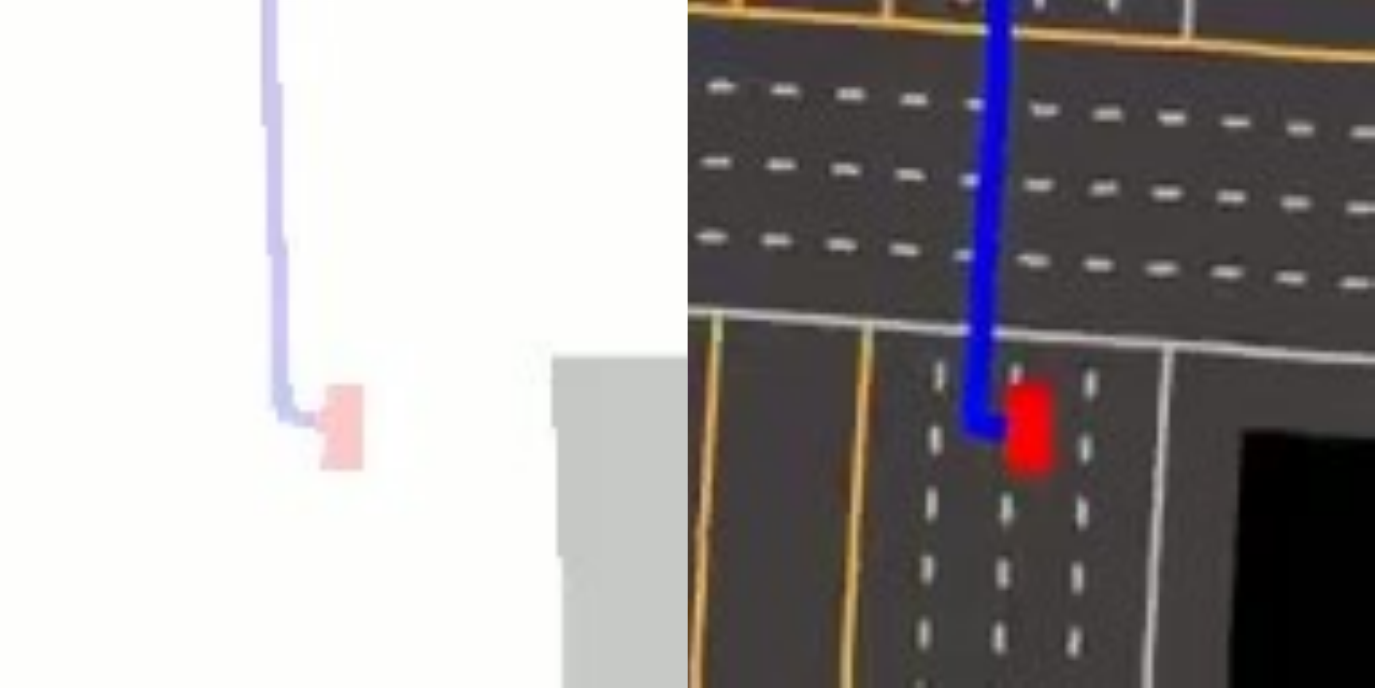}
        \caption{Jitter}
    \end{subfigure}
    \hfill
    \begin{subfigure}{0.19\linewidth}
        \centering
        \includegraphics[width=\linewidth]{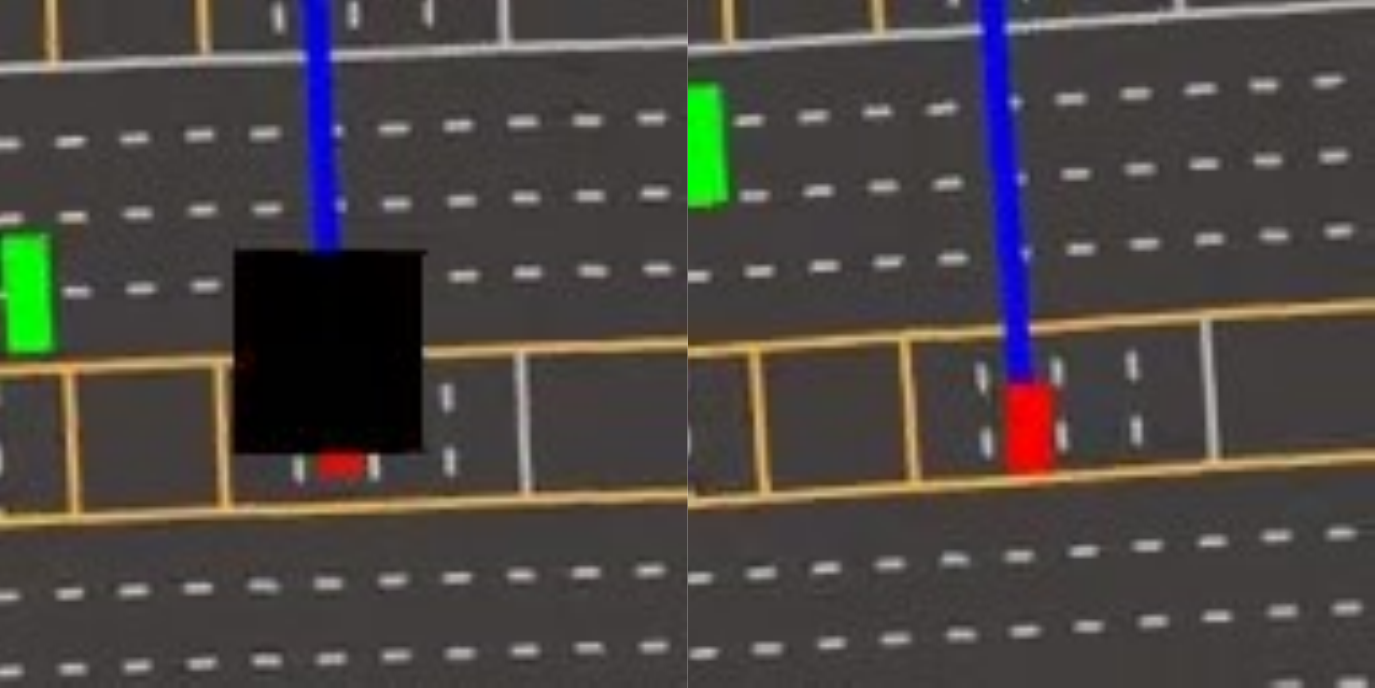}
        \caption{Occlusion}
    \end{subfigure}

    \caption{Visual comparison of the clean and corrupted from the CARLA BEV perspective. The left frame is the corrupted observation, and the right frame is the ground truth observation.}
    \label{fig:carla_noise_vertical}
\end{figure*}

\vspace{-1mm}

\begin{figure*}[!htbp]
    \centering

    \begin{subfigure}{\linewidth}
        \centering
        \includegraphics[width=0.19\linewidth]{Figures/carla_four_lane_chrome_combined.png}
        \includegraphics[width=0.19\linewidth]{Figures/carla_four_lane_gaussian_combined.png}
        \includegraphics[width=0.19\linewidth]{Figures/carla_four_lane_glare_combined.png}
        \includegraphics[width=0.19\linewidth]{Figures/carla_four_lane_jitter_combined.png}
        \includegraphics[width=0.19\linewidth]{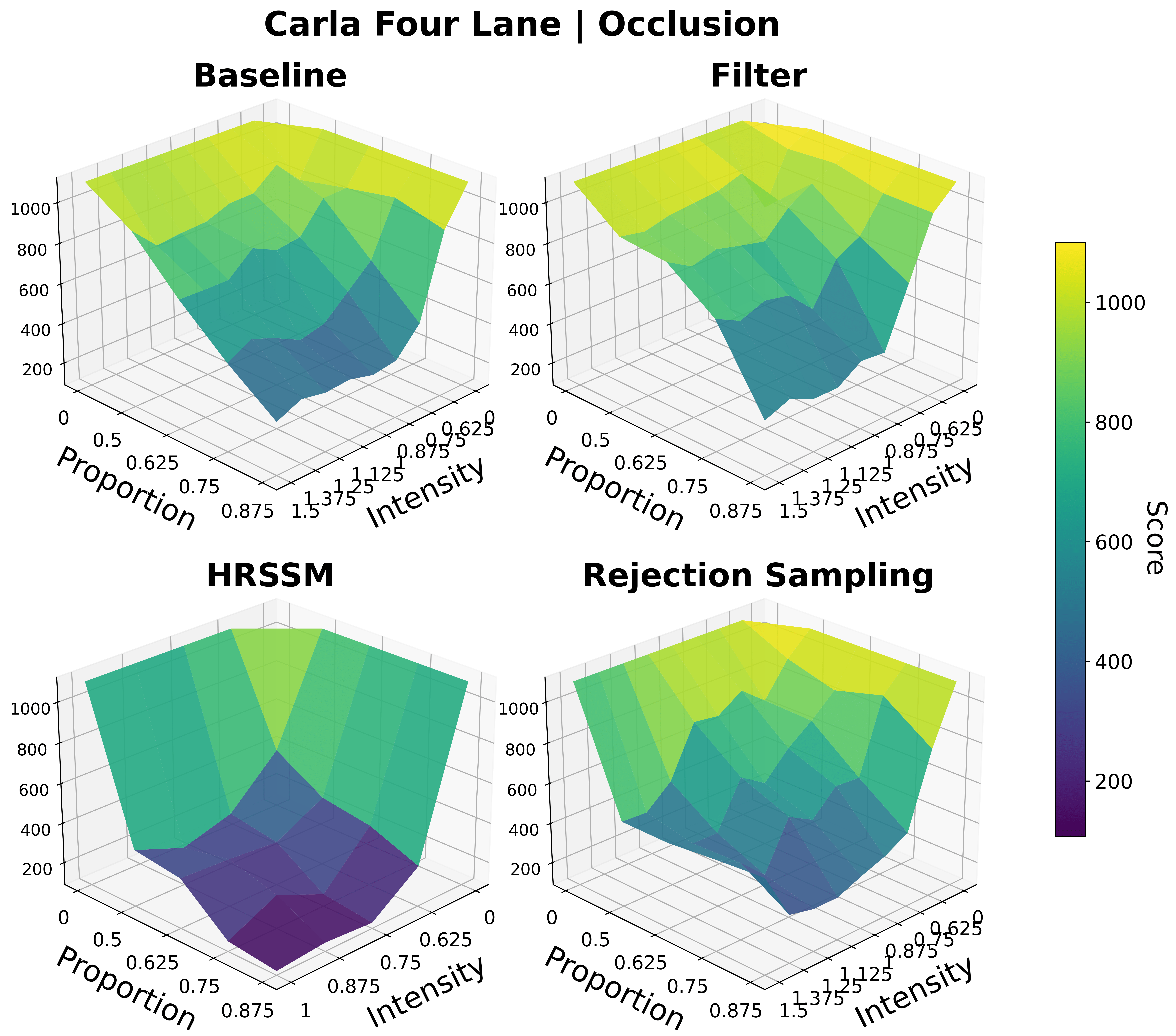}
        \caption{Four-Lane Driving}
    \end{subfigure}

    \begin{subfigure}{\linewidth}
        \centering
        \includegraphics[width=0.19\linewidth]{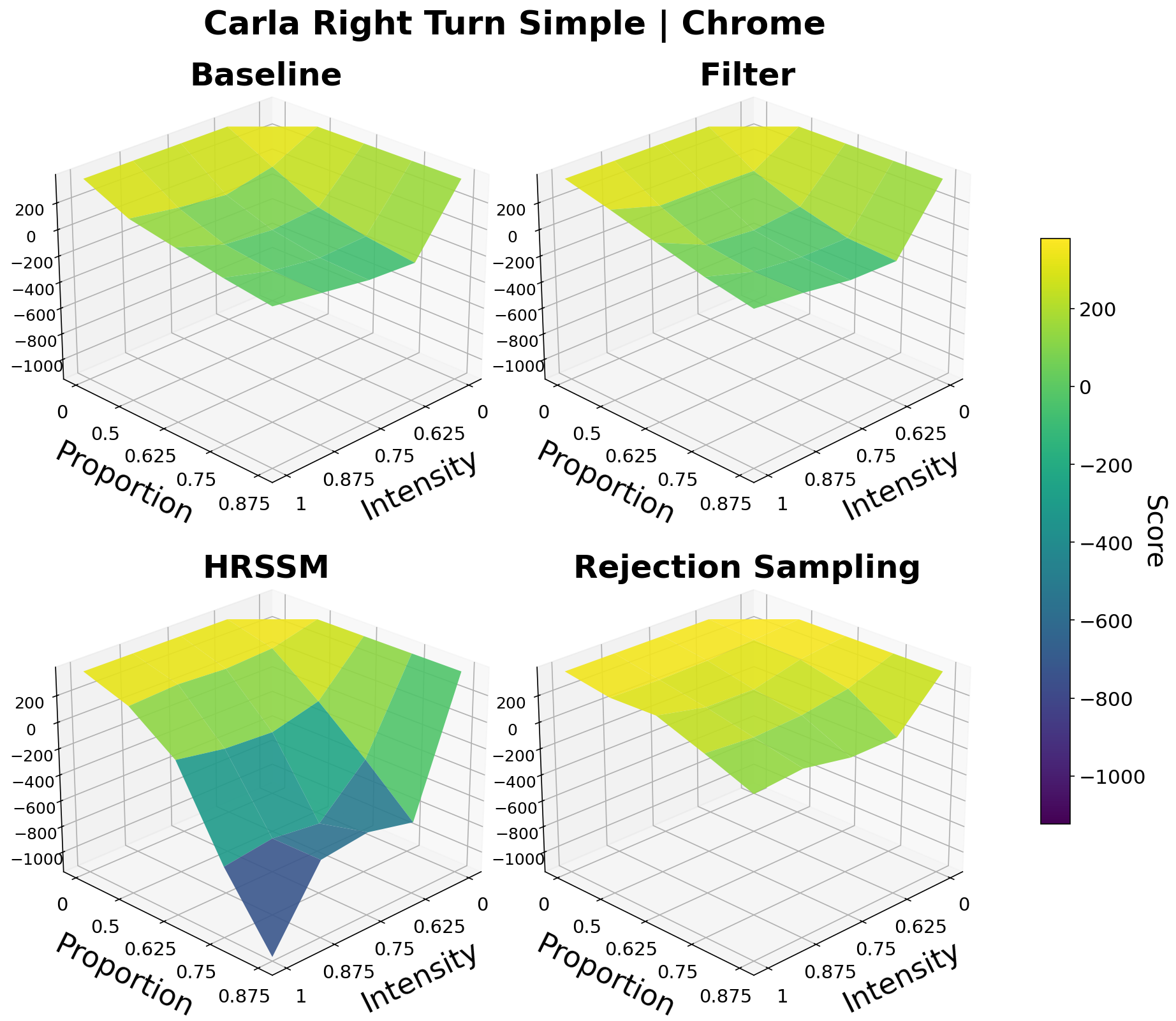}
        \includegraphics[width=0.19\linewidth]{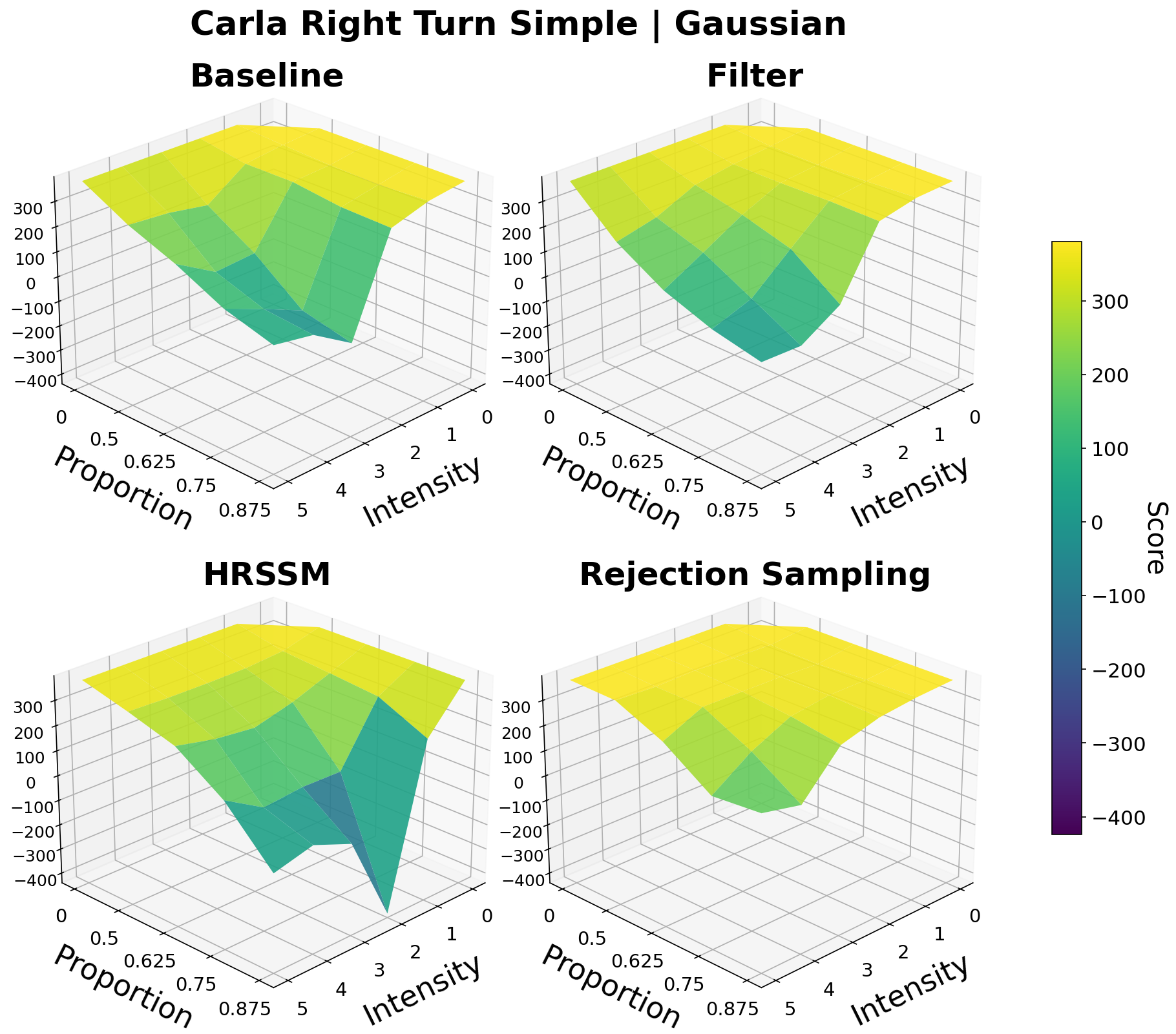}
        \includegraphics[width=0.19\linewidth]{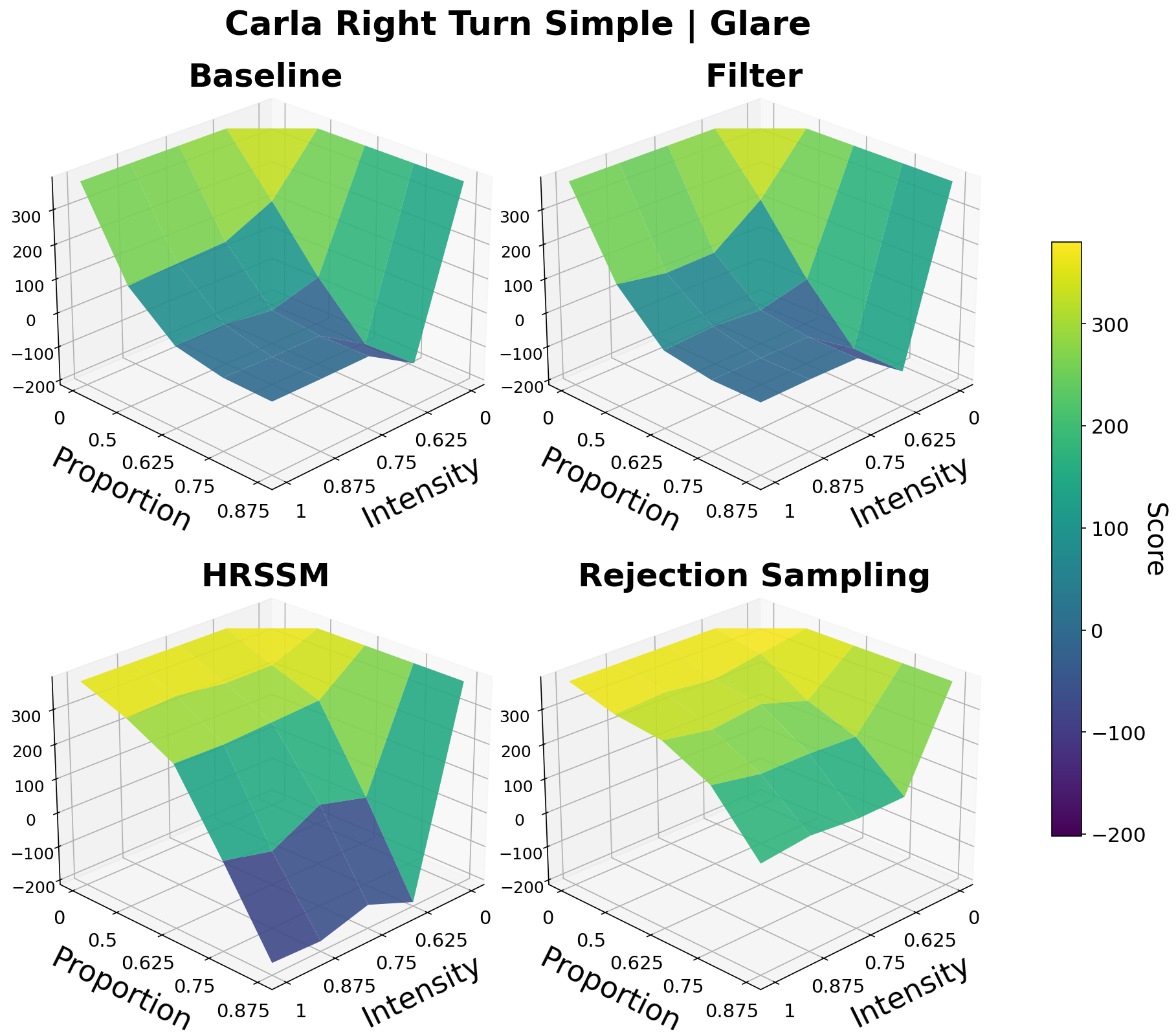}
        \includegraphics[width=0.19\linewidth]{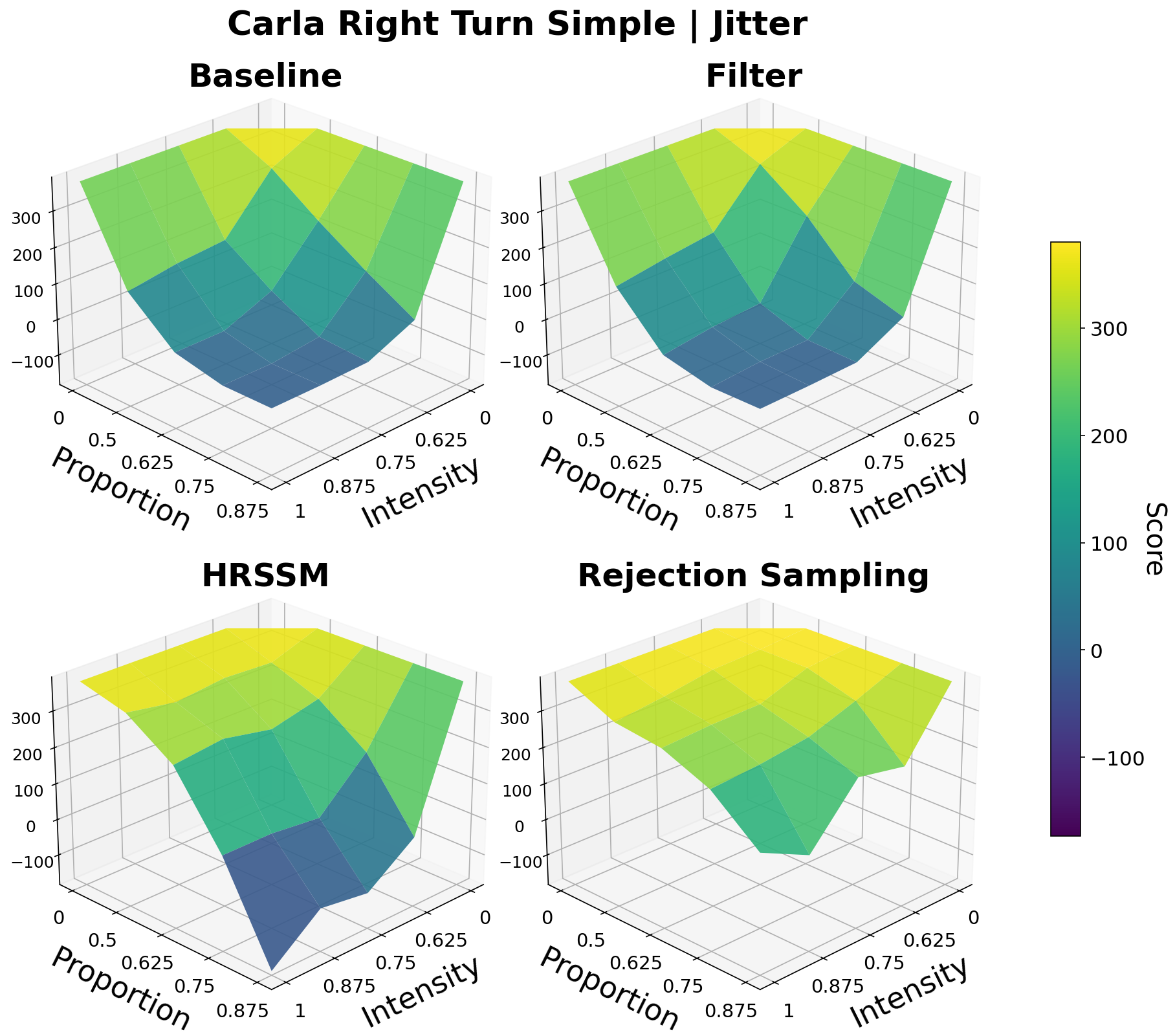}
        \includegraphics[width=0.19\linewidth]{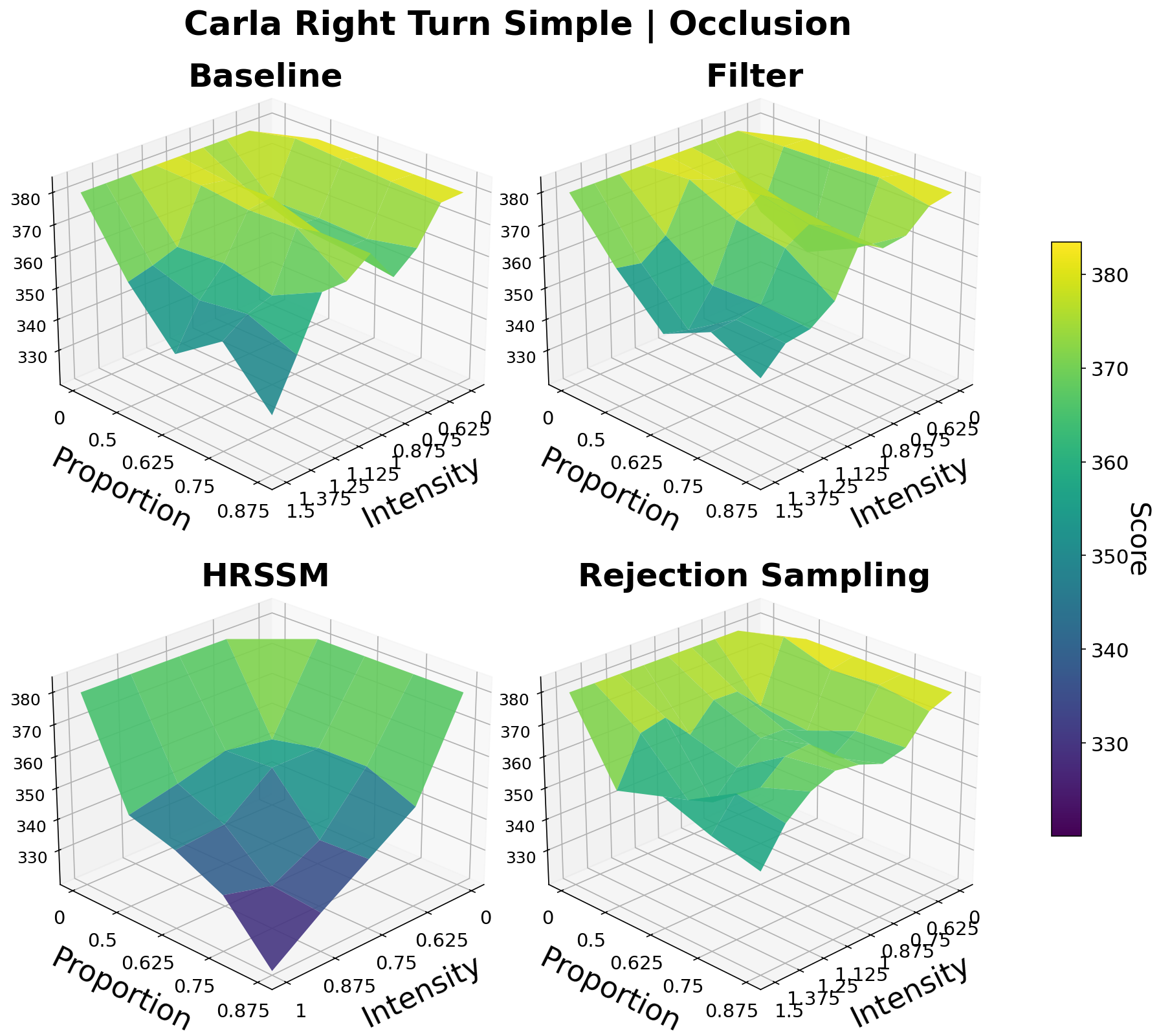}
        \caption{Right Turn}
    \end{subfigure}
    
    \begin{subfigure}{\linewidth}
        \centering
        \includegraphics[width=0.19\linewidth]{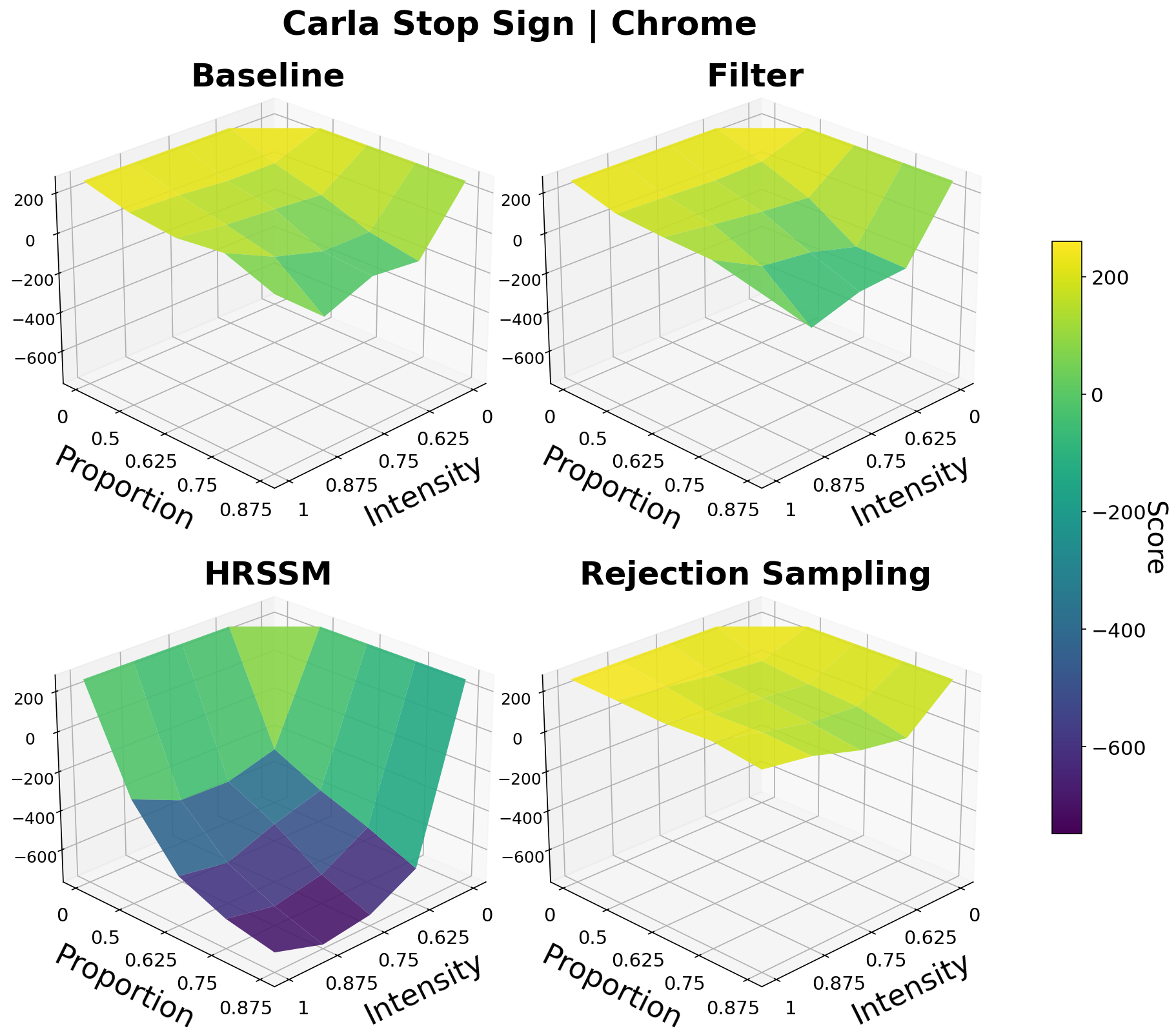}
        \includegraphics[width=0.19\linewidth]{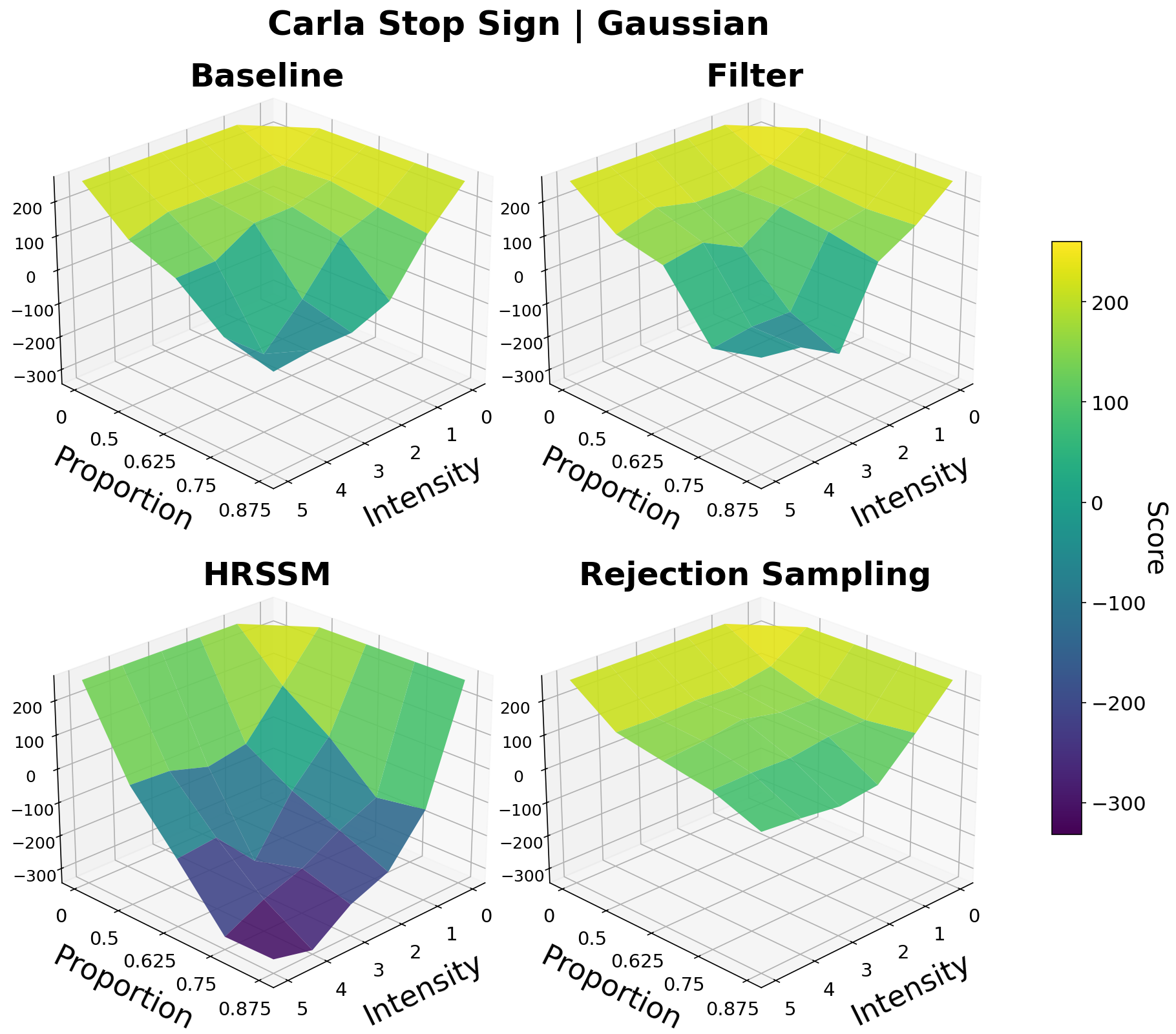}
        \includegraphics[width=0.19\linewidth]{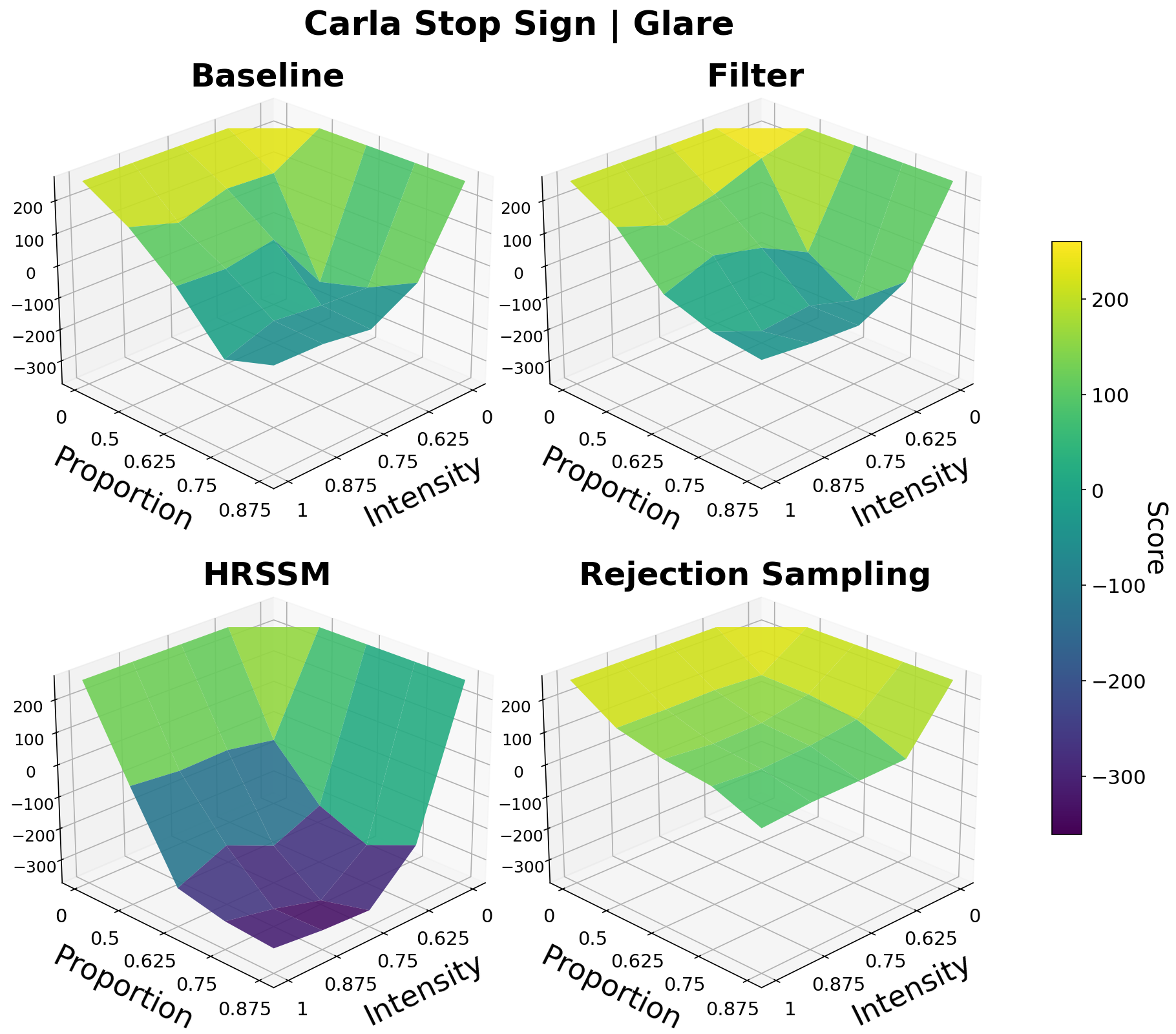}
        \includegraphics[width=0.19\linewidth]{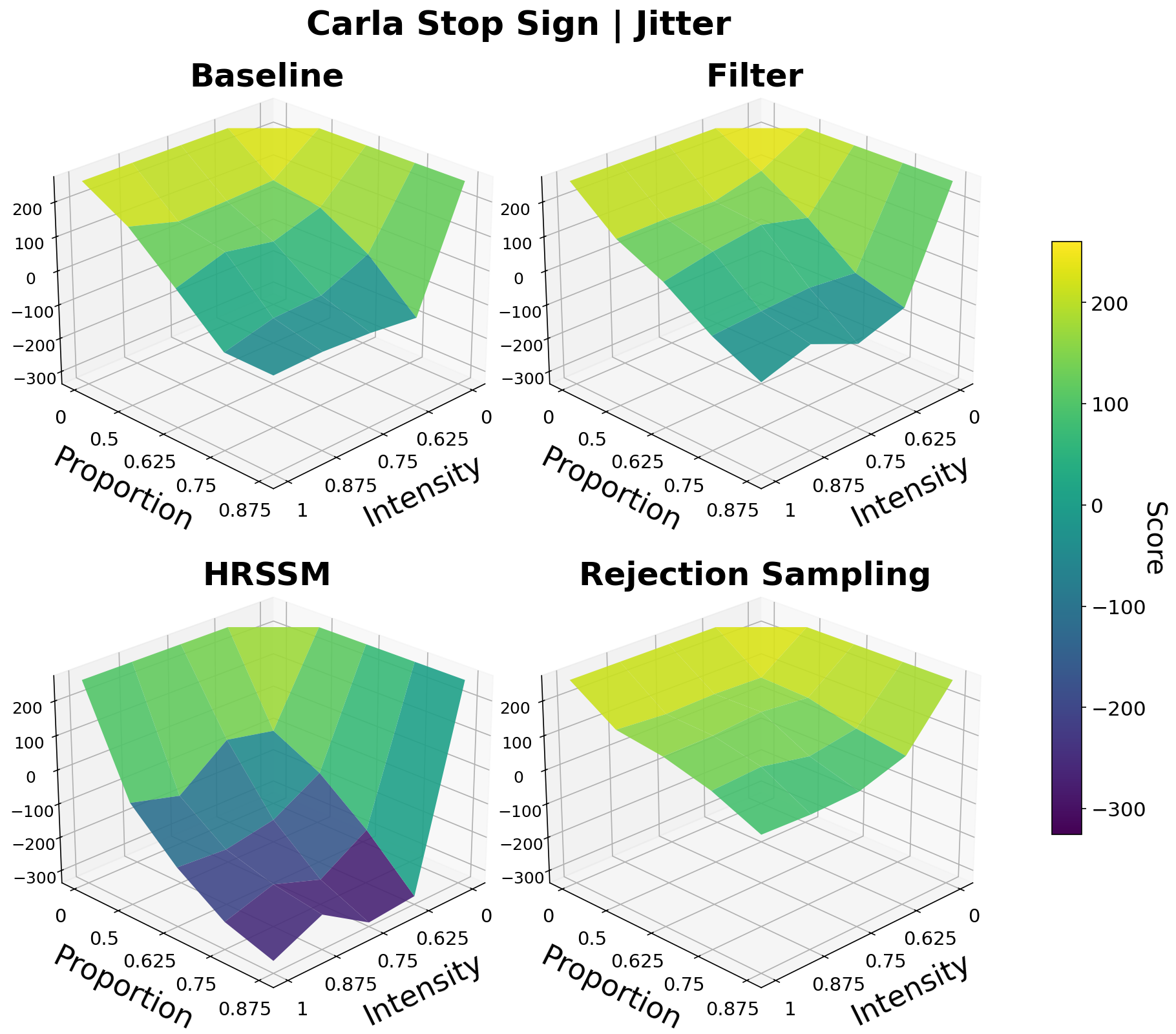}
        \includegraphics[width=0.19\linewidth]{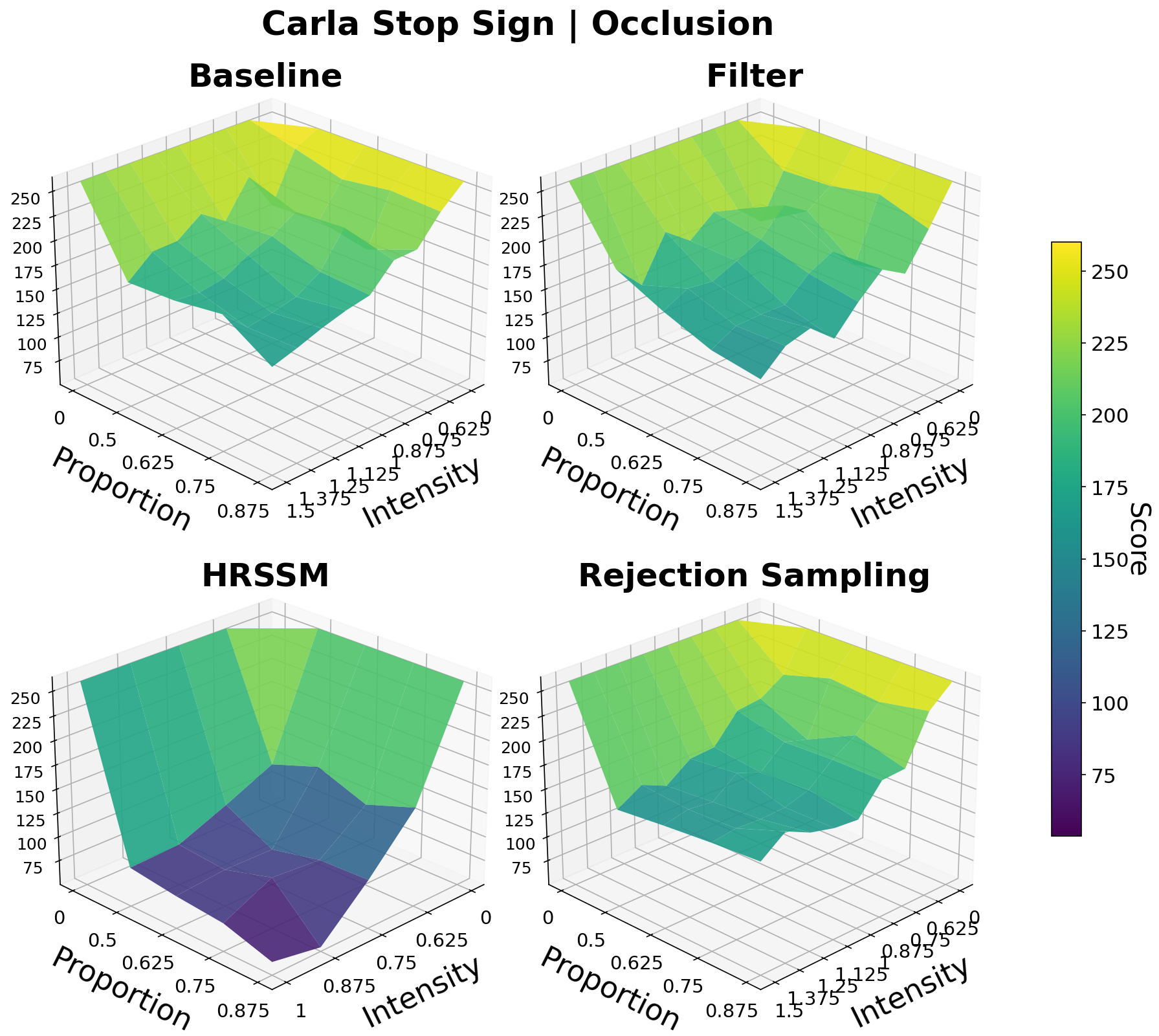}
        \caption{Stop Sign}
    \end{subfigure}

    \caption{
        Qualitative results across three CARLA tasks under five corruption types:
        Chrome, Gaussian, Glare, Jitter, and Occlusion.
    }
    \label{fig:carla_task_corruptions}
\end{figure*}

The goal is to test how well the agent can still perform its task under degraded visual inputs. Quantitatively, we plot 3D surface graphs of the task performance metric as a function of corruption intensity and proportion. Each surface corresponds to one CARLA task—Stop Sign, Right Turn, and Four-Lane Driving—and compares four methods: the baseline world-model agent, Median Filtering, our proposed Rejection Sampling, and HRSSM. For all tasks shown in Figure~\ref{fig:carla_task_corruptions}, our Rejection Sampling maintains a much smoother and higher performance surface, indicating stable behavior even under severe sensor degradation. This trend demonstrates that the proposed Filter not only delays the onset of failure but also preserves task continuity across extreme perturbations, achieving consistently superior robustness and reliability in all CARLA environments.

\begin{figure*}[!htbp]
    \centering

    \begin{subfigure}{0.95\linewidth}
        \centering
        \includegraphics[width=0.24\linewidth]{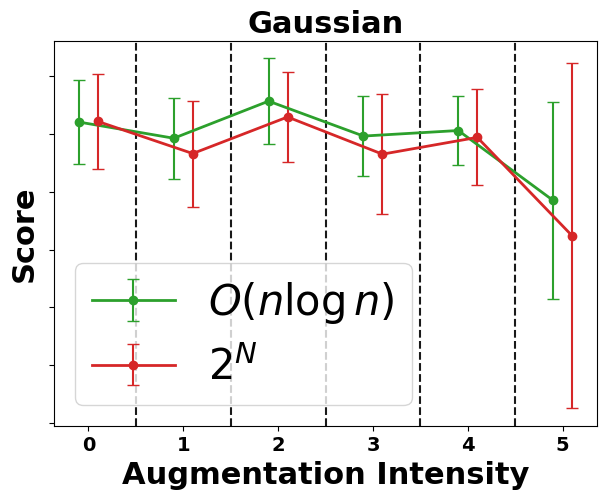}
        \includegraphics[width=0.24\linewidth]{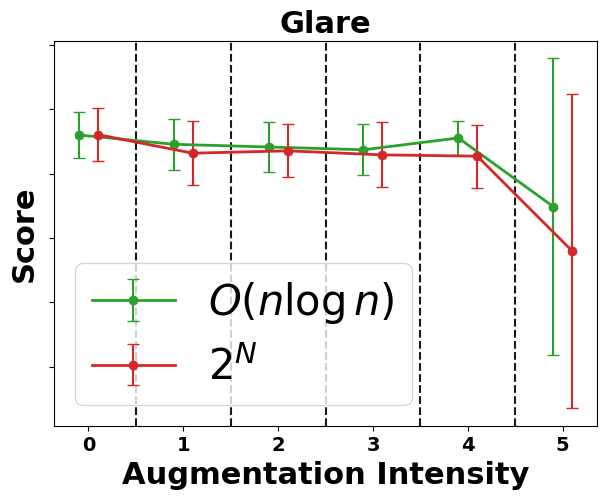}
        \includegraphics[width=0.24\linewidth]{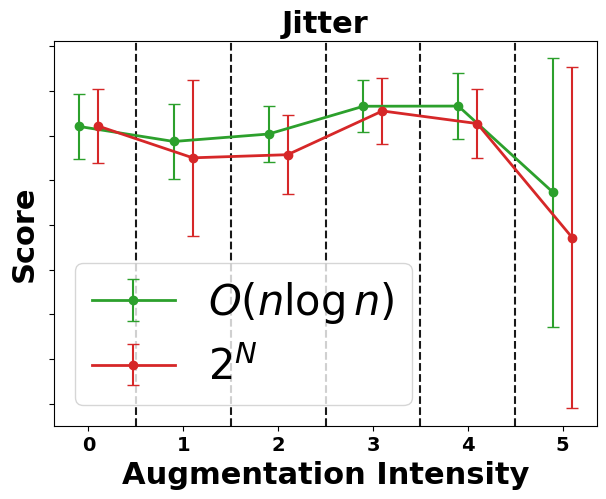}
        \includegraphics[width=0.24\linewidth]{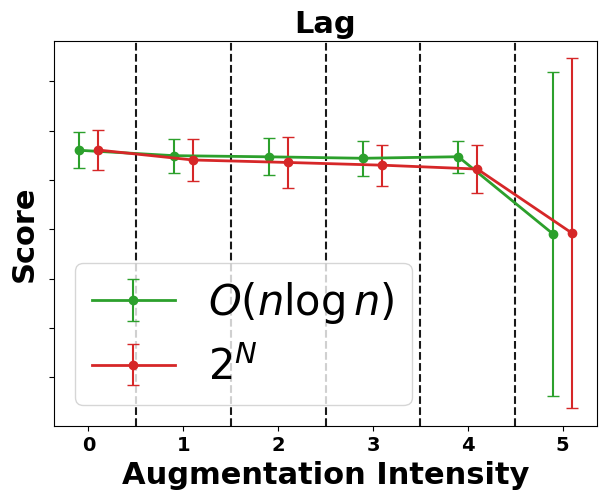}
        \caption{Right Turn}
    \end{subfigure}

    \vspace{0.8em}

    \begin{subfigure}{0.95\linewidth}
        \centering
        \includegraphics[width=0.24\linewidth]{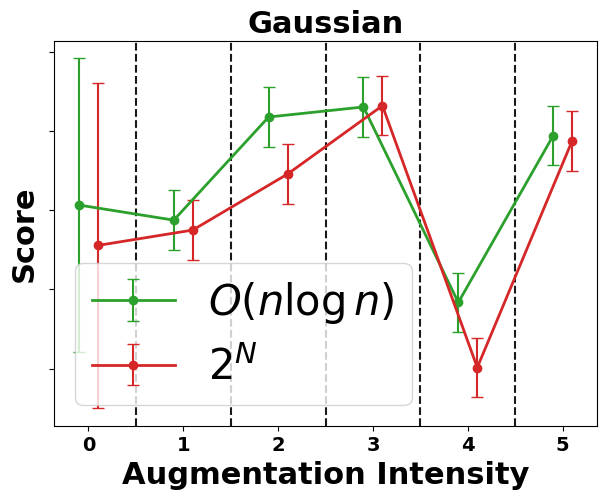}
        \includegraphics[width=0.24\linewidth]{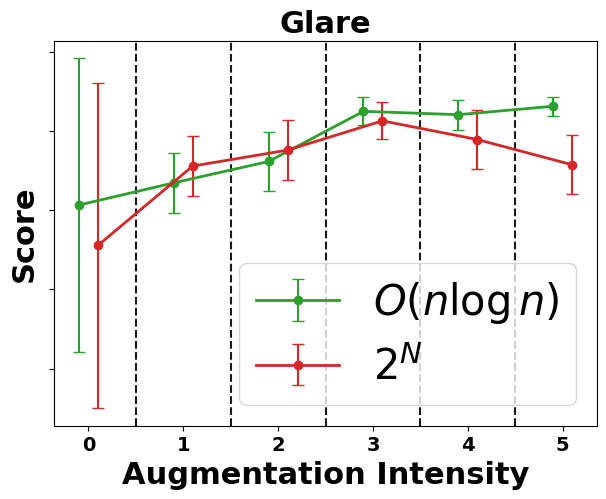}
        \includegraphics[width=0.24\linewidth]{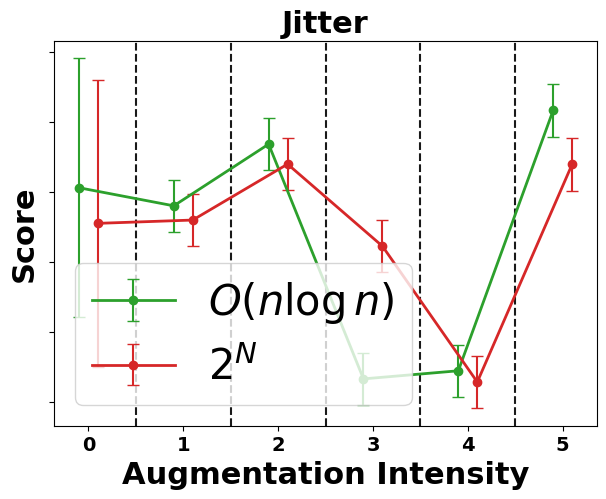}
        \includegraphics[width=0.24\linewidth]{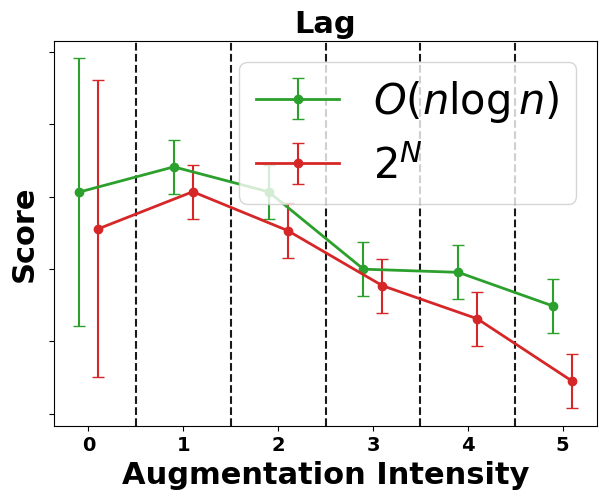}
        \caption{Four Lane Driving}
    \end{subfigure}

    \vspace{0.8em}

    \begin{subfigure}{0.95\linewidth}
        \centering
        \includegraphics[width=0.24\linewidth]{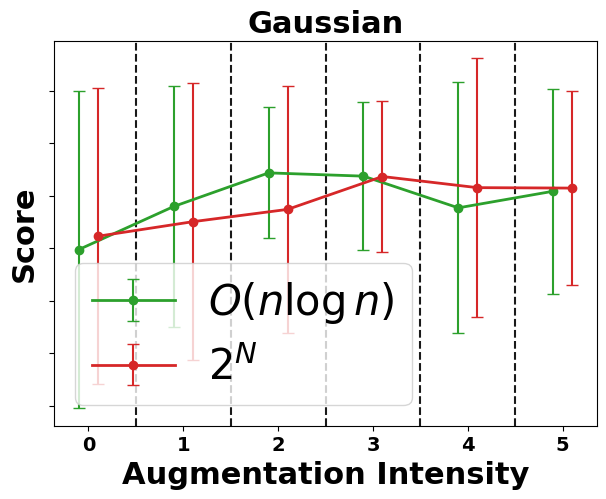}
        \includegraphics[width=0.24\linewidth]{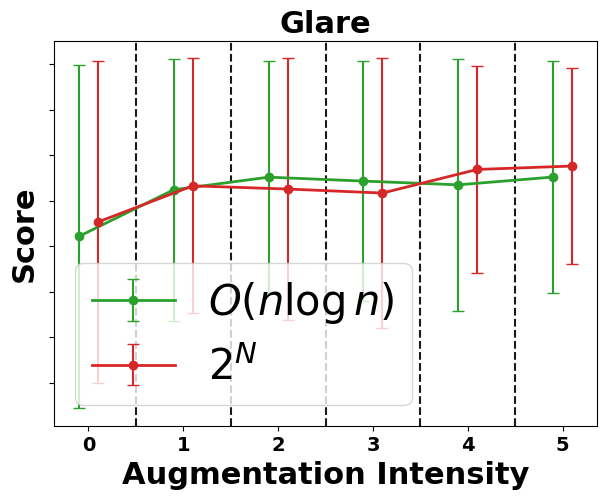}
        \includegraphics[width=0.24\linewidth]{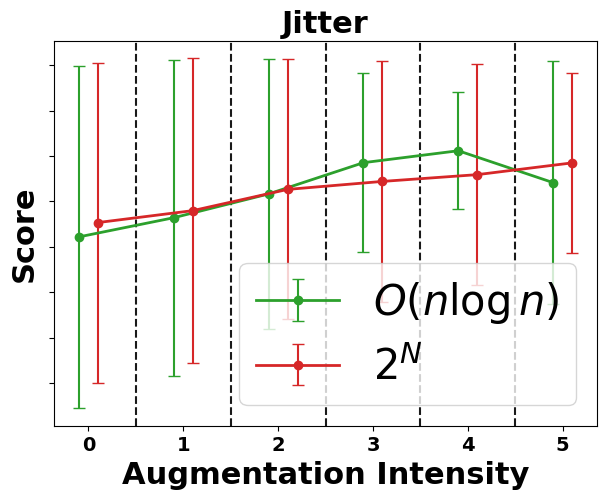}
        \includegraphics[width=0.24\linewidth]{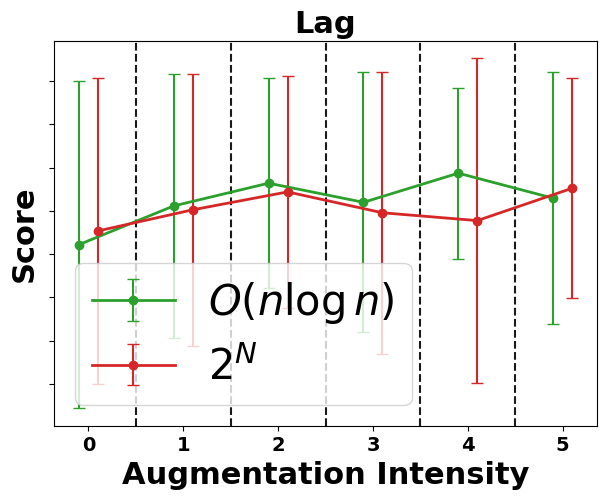}
        \caption{Stop Sign}
    \end{subfigure}

    \caption{Comparison across three tasks under different noise conditions. We compare the $2^N$ brute force variant against our proposed $O(n\log n)$ Algorithm variant~\ref{alg:nonadaptive_masking}.}
    \label{fig:2nComparisonFull}
\end{figure*}

\subsection{Efficiency of Surprise-Guided Filter}
\label{subsec:surprise}

Figure~\ref{fig:2nComparisonFull} illustrates the efficiency of our surprise-guided filtering approach by comparing the $O(n \log n)$ sensor-selection strategy against the exhaustive $2^N$ subset search across the three CARLA tasks—Right Turn, Four-Lane, and Stop Sign—under four corruption types: Gaussian, Glare, Jitter, and Lag. The specific perturbations applied to the camera view are defined as follows:

\begin{itemize}
    \item Gaussian Noise: adds pixel-level random variations, modeling low-light sensor interference.
    \item Glare: causes extreme overexposure that whitens the whole frame, erasing most visual information.
    \item Jitter: applies random shifts in contrast or brightness, simulating unstable or high-noise sensor conditions.
    \item Latency (Lag): causes the observation to lag behind the true state by reusing stale frames for several steps, emulating delayed or frozen sensor updates.
\end{itemize}

The evaluation spans a wide range of augmentation intensities to test how each method scales as visual degradation increases. For Gaussian Noise and Jitter, both methods display non-monotonic trends: mild noise can be mitigated by masking high-surprise views, while stronger perturbations introduce sharp drops, particularly in the Four-Lane task at intensity level~4. Glare yields more stable performance because overexposed channels are consistently identified and suppressed. Lag produces a near-monotonic decline across all tasks due to its strong disruption of temporal coherence. Crucially, the green $O(n \log n)$ curves typically track the red $2^N$ curves closely, despite the exponential cost associated with the brute-force method. This demonstrates that the surprise-guided filter attains robustness comparable to exhaustive subset evaluation at only a fraction of the computational burden. The results confirm that the learned surprise signal provides a reliable ranking over sensor utility, making combinatorial search unnecessary and enabling practical, scalable deployment in multi-sensor RL settings.

\subsection{Sensor Failure Analysis}
\label{subsec:sensor_failure}

\begin{figure*}[htbp]
    \centering

    \begin{subfigure}{\linewidth}
        \centering
        \includegraphics[width=0.9\linewidth]{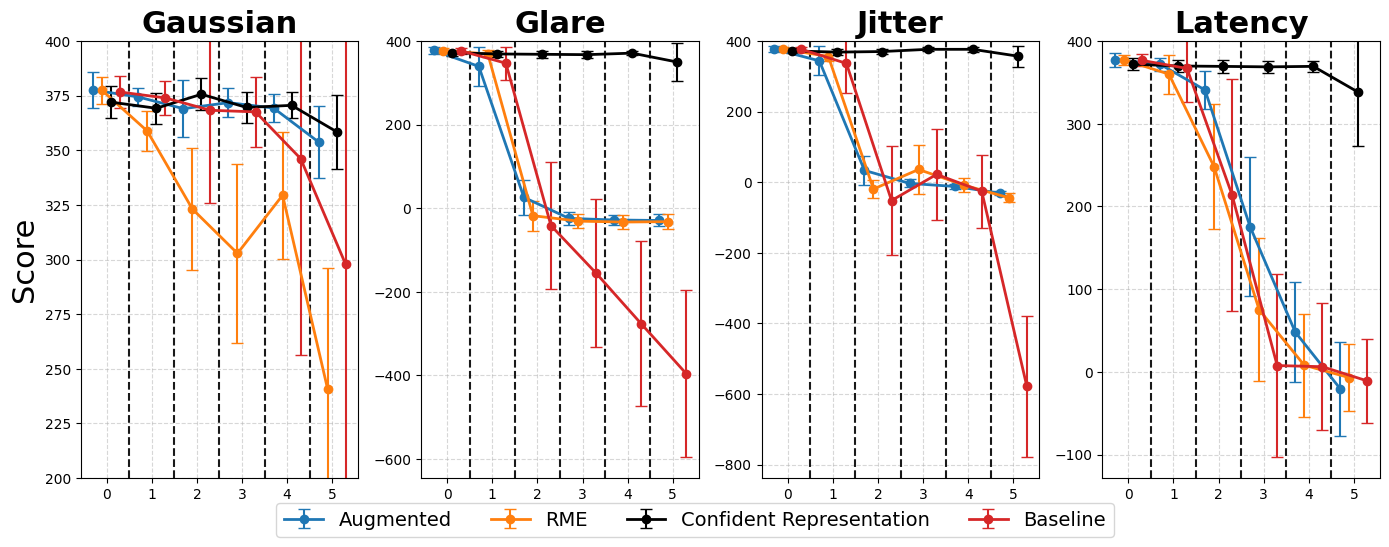}
        \caption{Right Turn}
    \end{subfigure}

    \vspace{0.8em}

    \begin{subfigure}{\linewidth}
        \centering
        \includegraphics[width=0.9\linewidth]{Figures/carla_stop_sign.png}
        \caption{Stop Sign}
    \end{subfigure}

    \vspace{0.8em}

    \begin{subfigure}{\linewidth}
        \centering
        \includegraphics[width=0.9\linewidth]{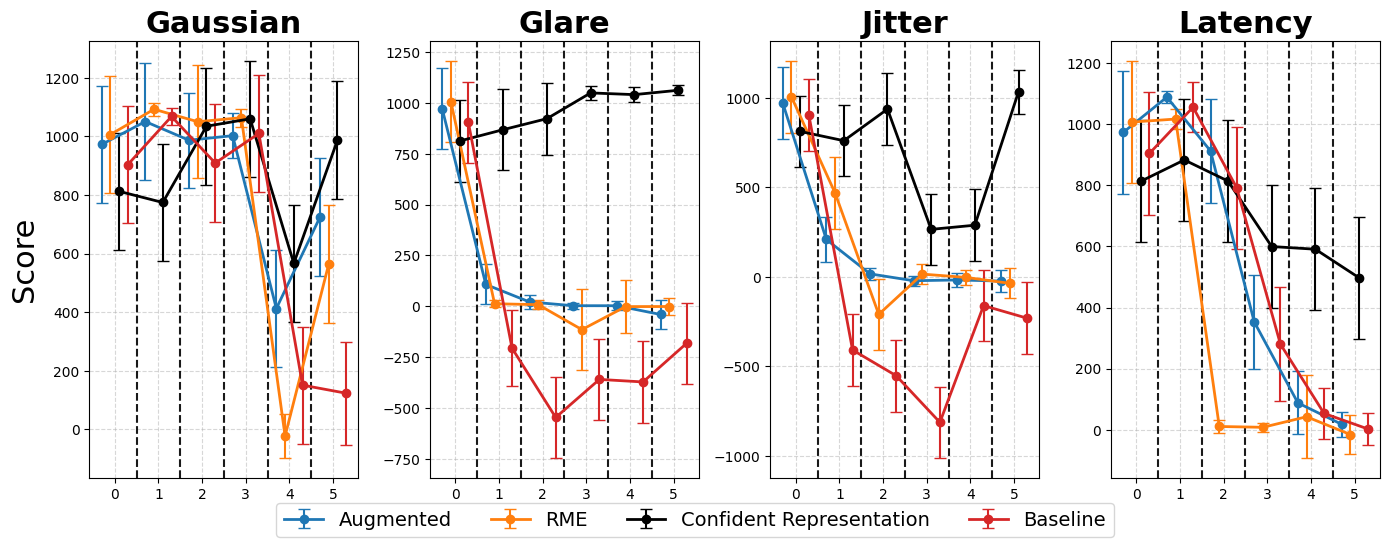}
        \caption{Four-Lane Driving}
    \end{subfigure}

    \caption{Agent performance as sensor failures increase.}
    \label{fig:All_Coarse_Carla}
\end{figure*}

Figure~\ref{fig:All_Coarse_Carla} provides a comprehensive breakdown of how increasing sensor failures affect agent performance across all three CARLA tasks—Right Turn, Stop Sign, and Four-Lane Driving—under four primary corruption types detailed in Section~\ref{subsec:surprise}: Gaussian Noise, Glare, Jitter, and Latency.

Each subfigure in Figure~\ref{fig:All_Coarse_Carla} plots episodic score as a function of the number of failed sensors, comparing four methods: the Baseline world-model agent, the Augmented agent trained with Gaussian noise, the RME masked-encoder agent, and our Confident Representation approach. Across all tasks and corruption types, Baseline exhibits the steepest decline, often collapsing to near-zero or negative scores as failures accumulate. RME reduces variance but remains sensitive to structured failures such as Glare and Latency. Augmented improves stability under moderate Gaussian noise, but still degrades noticeably when failures intensify.

In contrast, Confident Representation consistently maintains the highest performance curve, showing only mild degradation even with multiple failed sensors. Its advantage is particularly notable under Glare and Jitter, where corrupted visual channels severely mislead other methods. Under Latency, all agents experience some decline, yet Confident Representation preserves stable, positive performance while avoiding the abrupt failures observed in Baseline and RME.

Overall, these results show that surprise-guided representation selection remains effective across diverse CARLA configurations, enabling the agent to identify corrupted views, suppress unreliable inputs, and sustain robust control even as sensor failures accumulate.

\section{Safety Gymnasium Experiments}

In the Safety Gymnasium domain \citep{ji2023safety}, the agent’s goal is to navigate safely to objectives in an environment with obstacles. We provide three overlapping camera views as described in Section~\ref{app:carla_representations}, which capture similar scene content from complementary perspectives. We evaluate performance on three tasks: PointGoal1, where a point-mass agent must reach a goal location while avoiding hazards; PointButton1, where a point-mass must press buttons scattered in the map while avoiding obstacles, and CarGoal1, where a car-like agent must reach a goal on a road with obstacles. Each task also penalizes unsafe collisions with a cost.

Figure~\ref{fig:safetygymnasium} in the main text summarizes performance under four visual corruption
types—Gaussian, Jitter, Glare, and Occlusion—applied independently to each of
the three camera views. Since these views are partially redundant, corruption in a
single view does not immediately prevent task completion but can mislead the
policy when the corrupted view dominates the latent inference. The surprise-guided
filter addresses this by downweighting or suppressing views whose inferred
representations deviate strongly from the world model’s predictive prior. As shown
in the figure, this leads to consistent improvements in both performance and the
Score–Cost ratio across all Safety Gymnasium tasks. In particular, the method
effectively handles Glare and Occlusion, where one view becomes unreliable but
the remaining views still convey meaningful structure. The resulting gains in safety
and stability demonstrate that even in environments with overlapping visual inputs,
surprise-guided representation selection provides measurable robustness against
view-specific corruptions.

\section{Hardware Details and Hyperparameters}
\begin{table}[h!]
\centering
\resizebox{0.9\linewidth}{!}{%
\begin{tabular}{|l|c|c|}
\toprule
\textbf{Name} & \textbf{Symbol} & \textbf{Value} \\
\midrule
\multicolumn{3}{l}{\textbf{General}} \\
\midrule
Replay capacity & --- & $5 \times 10^6$ \\
Batch size & $B$ & 16 \\
Batch length & $T$ & 64 \\
Activation & --- & $\operatorname{RMSNorm}+\operatorname{SiLU}$ \\
Learning rate & --- & $4 \times 10^{-5}$ \\
Gradient clipping & --- & $\operatorname{AGC}(0.3)$ \\
Optimizer & --- & $\operatorname{LaProp}(\epsilon=10^{-20})$ \\
\midrule
\multicolumn{3}{l}{\textbf{World Model}} \\
\midrule
Reconstruction loss scale & $\beta_{\mathrm{pred}}$ & 1 \\
Dynamics loss scale & $\beta_{\mathrm{dyn}}$ & 1 \\
Representation loss scale & $\beta_{\mathrm{rep}}$ & 0.1 \\
Latent unimix & --- & $1\%$ \\
Free nats & --- & 1 \\
\midrule
\multicolumn{3}{l}{\textbf{Actor Critic}} \\
\midrule
Imagination horizon & $H$ & 15 \\
Discount horizon & $1/(1-\gamma)$ & 333 \\
Return lambda & $\lambda$ & 0.95 \\
Critic loss scale & $\beta_{\mathrm{val}}$ & 1 \\
Critic replay loss scale & $\beta_{\mathrm{repval}}$ & 0.3 \\
Critic EMA regularizer & --- & 1 \\
Critic EMA decay & --- & 0.98 \\
Actor loss scale & $\beta_{\mathrm{pol}}$ & 1 \\
Actor entropy regularizer & $\eta$ & $3 \times 10^{-4}$ \\
Actor unimix & --- & $1\%$ \\
Actor RetNorm scale & $S$ & $\operatorname{Per}(R, 95) - \operatorname{Per}(R, 5)$ \\
Actor RetNorm limit & $L$ & 1 \\
Actor RetNorm decay & --- & 0.99 \\
\bottomrule
\end{tabular}%
}
\caption{Hyperparameters of the Dreamerv3 model. We use the exact same parameters as discussed in~\cite{hafner2025mastering}.}
\label{tab:hyperparams}
\end{table}

\paragraph{Crafter}
All experiments can be reproduced on a system equipped with two NVIDIA GeForce GTX 1080 GPUs with 8 GB VRAM each to handle environment complexity, an AMD Ryzen 5 5600X 6-core processor, and at least 500 MB of storage for files (excluding training data, which depends on the environment and model hyperparameters).
\paragraph{CARLA and Safety Gymnasium} All reported experiments related to Safety-Gymnasium and CARLA domains can be conducted on a workstation equipped with an NVIDIA GeForce RTX 4090 GPU (24 GB VRAM), an AMD Ryzen 9 7950X (16-core, 32-thread) CPU, 128 GB DDR5 RAM, and running Ubuntu 22.04 LTS (CUDA 12.4).
\paragraph{Cosmos}
All reported experiments relating to the Cosmos world model can be conducted on a workstation equipped with dual Intel Xeon 6737P CPUs (64 cores total), 2.0~TiB of DDR5 system memory, a NVIDIA RTX PRO 6000 utilizing driver version 580.82.07 (CUDA 13.0). 
\label{app:hardware}


\section{Further Results}








\subsection{Cosmos World Model}
\label{app:cosmos}

To further verify that the proposed rejection-based filtering mechanism generalizes beyond latent-state world models, we abstract the process of Figure~\ref{fig:rejection_sampling} and extend our study to the Cosmos world model---a Diffusion-based video world model that performs pixel-space prediction through large-scale diffusion and spatio-temporal self-attention. 
Unlike the DreamerV3 architecture used in our main experiments, which relies on a Variational Autoencoder with explicit latent variables and KL-based surprise, Cosmos models dynamics directly in the image domain via autoregressive diffusion-based video prediction. 
This enables high-fidelity frame synthesis but also makes it highly sensitive to corruption in its first-frame conditioning.

\begin{figure}[!htbp]
    \centering
    \begin{minipage}{0.49\linewidth}
        \centering
        \includegraphics[width=\linewidth]{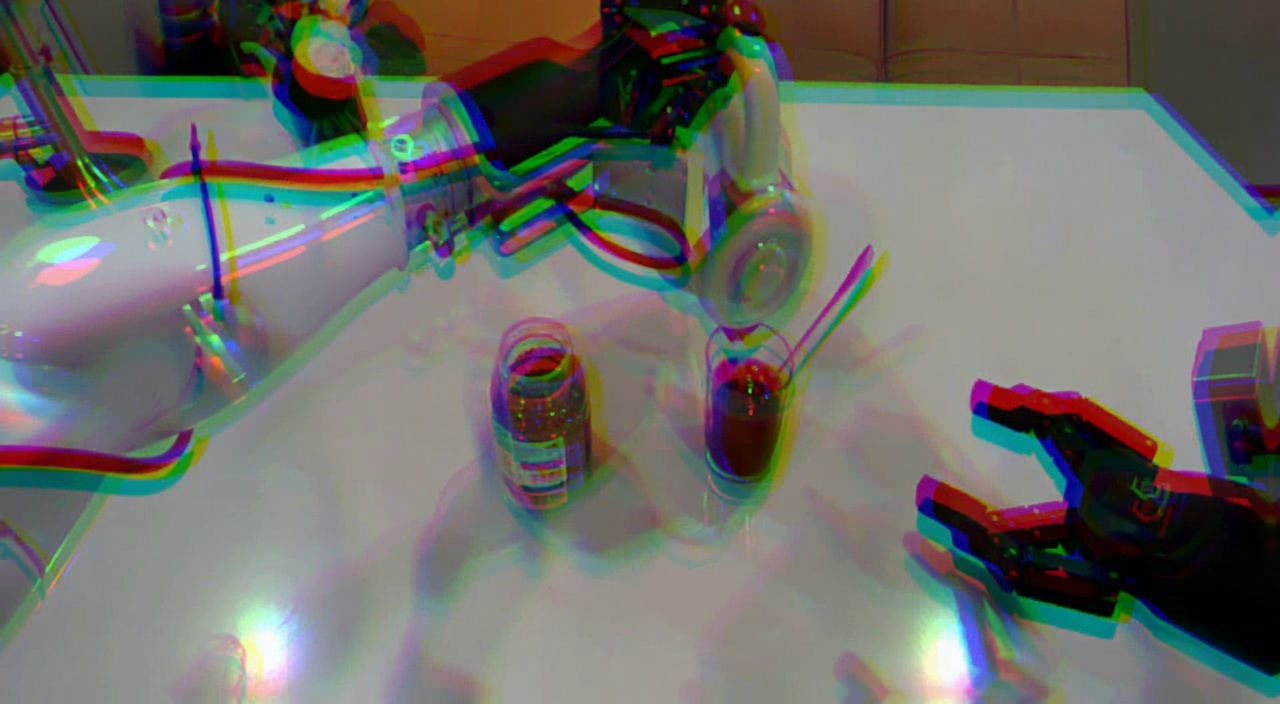}
        \subcaption{Chromatic Aberration Frame 1}
    \end{minipage}
    \hfill
    \begin{minipage}{0.49\linewidth}
        \centering
        \includegraphics[width=\linewidth]{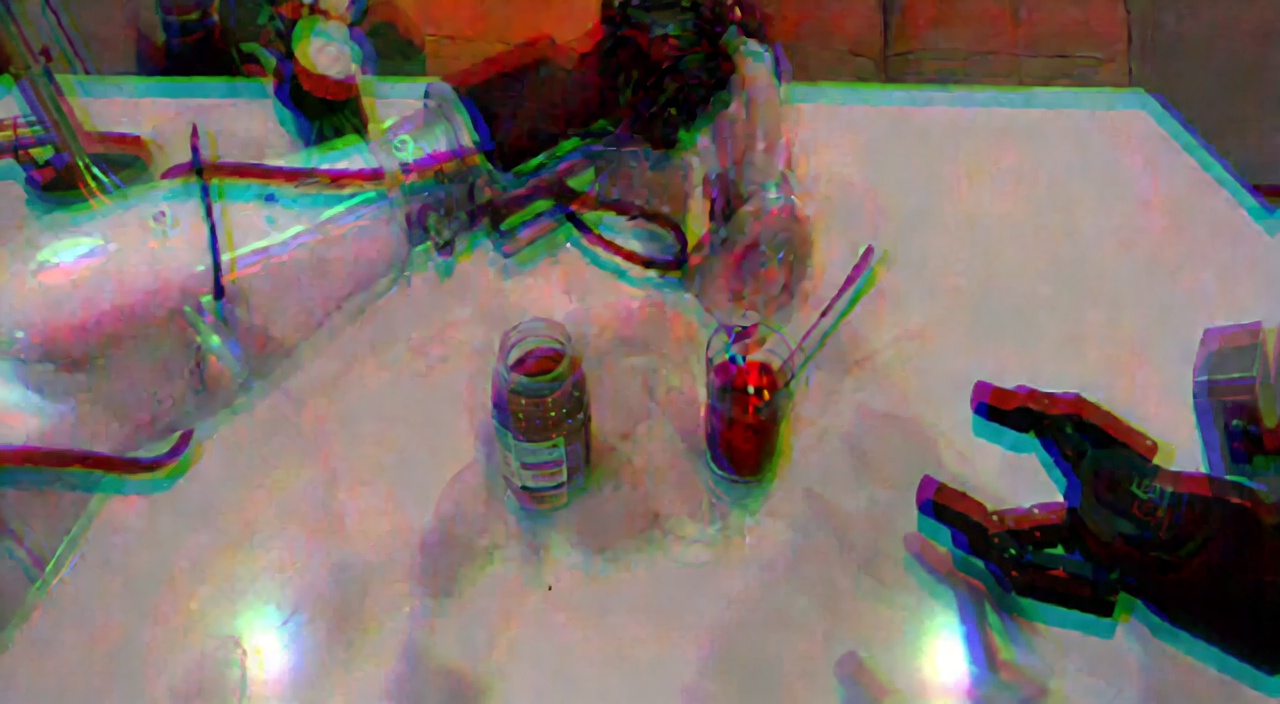}
        \subcaption{Chromatic Aberration Frame 2}
    \end{minipage}

    \vspace{0.8em}

    \begin{minipage}{0.49\linewidth}
        \centering
        \includegraphics[width=\linewidth]{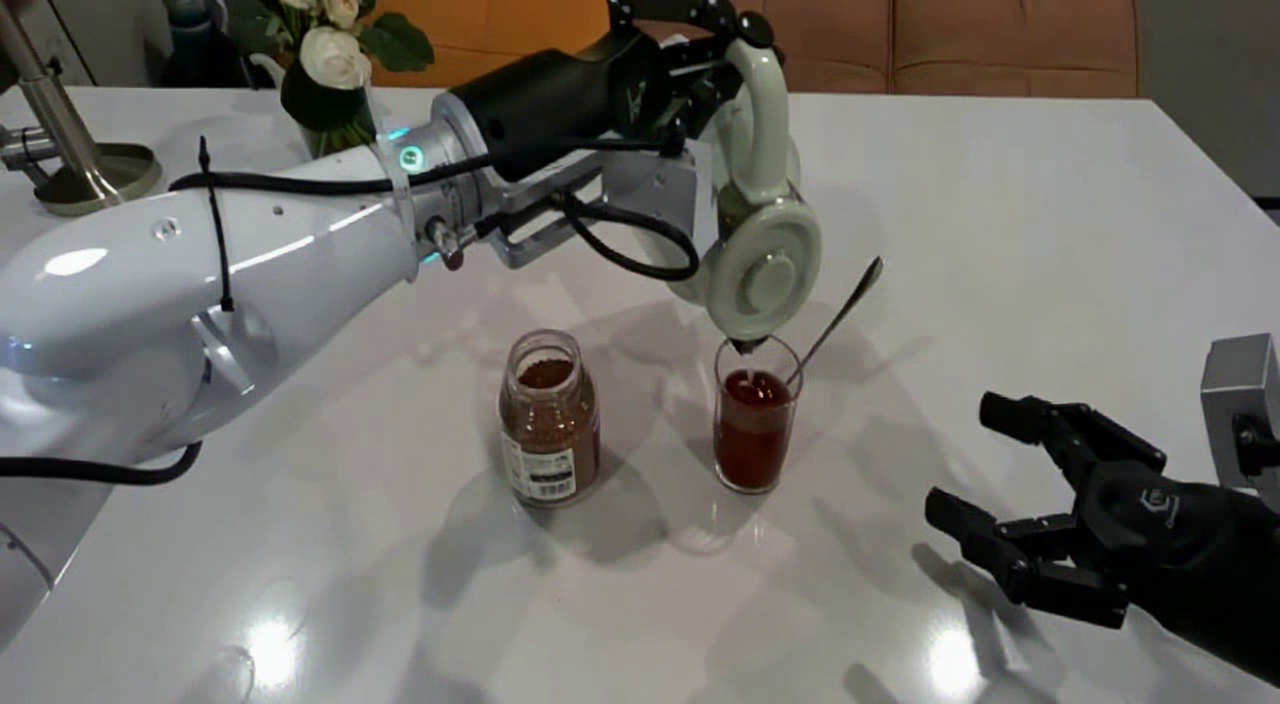}
        \subcaption{Clean Observation Frame 1}
    \end{minipage}
    \hfill
    \begin{minipage}{0.49\linewidth}
        \centering
        \includegraphics[width=\linewidth]{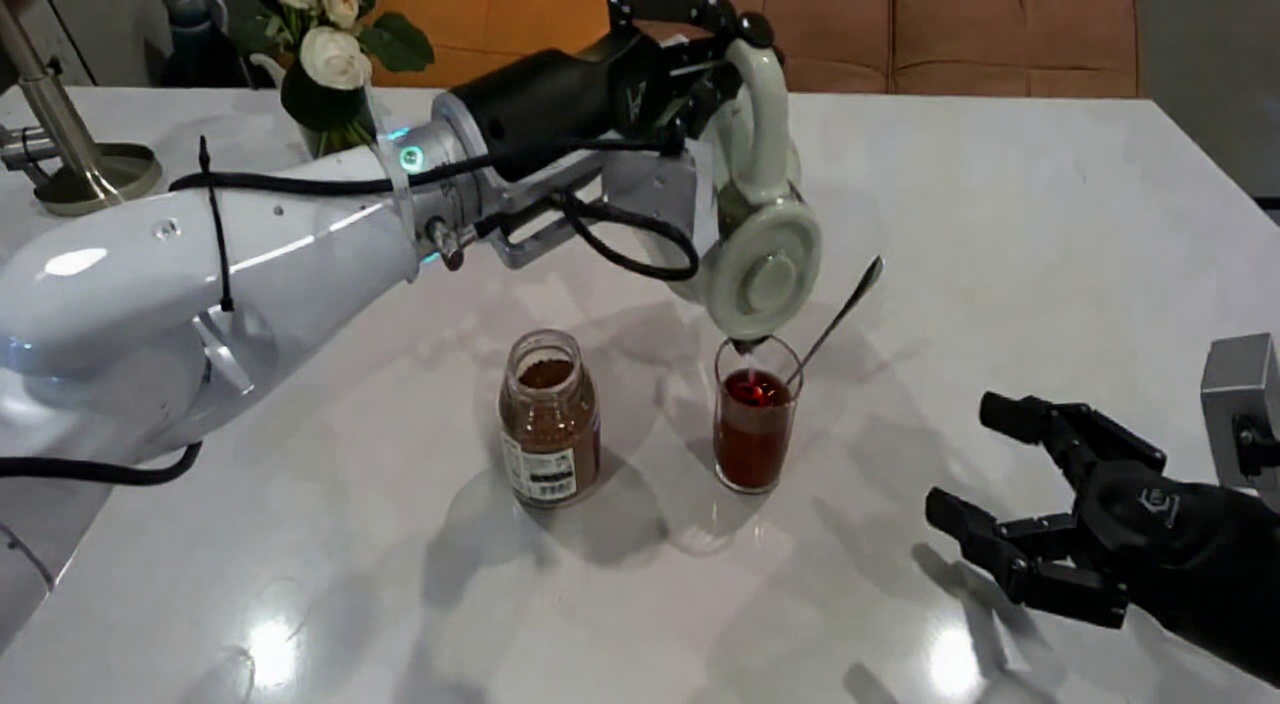}
        \subcaption{Clean Observation Frame 2}
    \end{minipage}

    \caption{
        First-frame conditioning comparison under the Cosmos model: the first column shows the input frame, and the second column shows the corresponding generated frame that was conditioned by the first frame. We utilize the Mean Squared error between two frames as our rejection score $M(x_*)$.
    }
    \label{fig:cosmos_robot_pouring_chrome2}
\end{figure}

We evaluate Cosmos on the robot pouring video, where a robotic manipulator pours liquid between jars on a tabletop. 
In our experiments, 75\% of the input video is randomly perturbed with one of the noises discussed in~\ref{app:noises}. 
Because Cosmos conditions its rollout on the last N frames of the input video, this corruption propagates through immediate subsequent predictions, as shown in the top row of Figure~\ref{fig:cosmos_robot_pouring_chrome2}. 
After enabling our rejection mechanism, the system detects the corrupted immediate next frame as high-surprise and instead relies on internally predicted latent trajectories to reconstruct the scene. 
The bottom row shows markedly improved geometry and color fidelity around the robot arm and jars, indicating a safer frame to accept, demonstrating that selective rejection of corrupted conditioning frames can effectively suppress artifact propagation even in Transformer-based video world models.
\subsubsection{Video Prompt: Robotic Arm Pouring Sequence}
\begin{tcolorbox}[colback=gray!5!white, colframe=black!60, title=Scene Description, sharp corners, boxrule=0.5pt]
A robotic arm, primarily white with black joints and cables, is shown in a clean, modern indoor setting with a white tabletop. The arm, equipped with a gripper holding a small, light green pitcher, is positioned above a clear glass containing a reddish-brown liquid and a spoon. The robotic arm is in the process of pouring a transparent liquid into the glass. To the left of the pitcher, there is an open jar with a similar reddish-brown substance visible through its transparent body. In the background, a vase with white flowers and a brown couch are partially visible, adding to the contemporary ambiance. The lighting is bright, casting soft shadows on the table. The robotic arm's movements are smooth and controlled, demonstrating precision in its task. As the video progresses, the robotic arm completes the pour, leaving the glass half-filled with the reddish-brown liquid. The jar remains untouched throughout the sequence, and the spoon inside the glass remains stationary. The other robotic arm on the right side also stays stationary throughout the video. The final frame captures the robotic arm with the pitcher finishing the pour, with the glass now filled to a higher level, while the pitcher is slightly tilted but still held securely by the gripper.
\end{tcolorbox}

\subsubsection{Cosmos Evaluation Metrics}
\label{app:cosmos_eval}

We evaluate the denoised and base Cosmos generations using metrics from the PAI-Bench~\cite{PAIBench2025}, a comprehensive benchmark for assessing physical-world video generation and prediction quality. 
Following the official protocol, we use the last frame of the ground truth clean input video as the reference image to compute the image-to-video (i2v) fidelity scores. The metrics are described as follows:

\begin{itemize}
    \item \textbf{Aesthetic Quality.} Measures the visual appeal and perceptual realism of the generated video, including composition, color balance, and absence of artefacts.
    \item \textbf{Background Consistency.} Evaluates the temporal stability of background regions across frames—high scores indicate minimal flicker or spatial drift.
    \item \textbf{Imaging Quality.} Quantifies camera-like clarity, sharpness, and signal-to-noise ratio of the generated video frames.
    \item \textbf{Motion Smoothness.} Assesses the temporal coherence of motion; low values indicate jitter or unnatural frame transitions.
    \item \textbf{Overall Consistency.} Aggregates spatial and temporal coherence into a single holistic quality measure of the entire video.
    \item \textbf{Subject Consistency.} Examines whether the primary object or agent remains stable in shape, color, and identity over time.
    \item \textbf{i2v Background.} Image-to-video fidelity for background regions, comparing each frame to the last frame of the input video.
    \item \textbf{i2v Subject.} Image-to-video fidelity for the subject region, capturing how faithfully the generated subject matches the reference frame.
\end{itemize}

Higher scores across these dimensions indicate smoother, more coherent, and visually faithful video synthesis. 
In our experiments shown in Table~\ref{tab:cosmos_overall_scores}, the proposed rejection sampling consistently improves both the i2v metrics and overall perceptual quality compared to the baseline generations.

\subsection{Semantic Noise Experiments}
Although outside the main scope of our work, we briefly consider experimentation with semantic noise.
\begin{figure}[]
    \centering
    \includegraphics[width=0.95\linewidth]{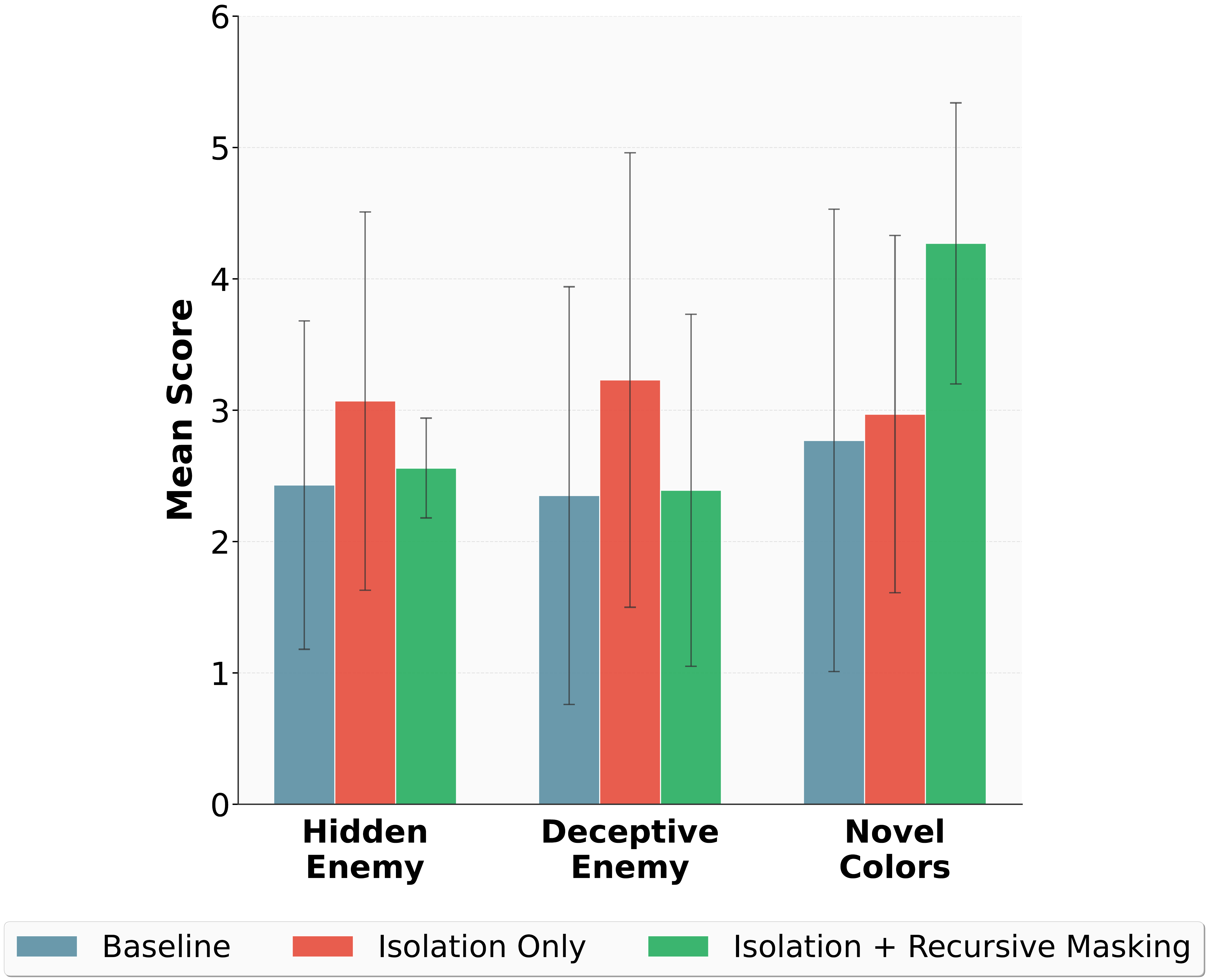}
    
    \caption{Crafter: Baseline vs Confident Representation Performance. Scores
    improve as the agent utilizes masking in Algorithm \ref{alg:nonadaptive_masking} to adapt to different types of novelties.}
    \label{fig:crafter_baseline_surprise}
\end{figure}

We choose the Crafter domain~\citep{hafner2022crafter} due to its utility in embedding abstract skins and unknown colors during inference time, primarily meant to distract the agent.

\begin{figure}[hbtp]
    \centering
    \begin{subfigure}[b]{0.32\linewidth}
        \centering
        \includegraphics[width=\linewidth]{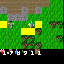}
        \caption{Hidden Enemy}
        \label{fig:crafter_hidden_enemy1}
    \end{subfigure}
    \hfill
    \begin{subfigure}[b]{0.32\linewidth}
        \centering
        \includegraphics[width=\linewidth]{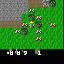}
        \caption{Deceptive Enemy}
        \label{fig:crafter_deceptive_enemy1}
    \end{subfigure}
    \hfill
    \begin{subfigure}[b]{0.32\linewidth}
        \centering
        \includegraphics[width=\linewidth]{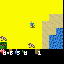}
        \caption{Novel Colors}
        \label{fig:crafter_novel_colors1}
    \end{subfigure}
    \caption{Qualitative visualization of agents’ behavior in different semantic novelty scenarios. Figure~\ref{fig:crafter_hidden_enemy1} shows the \emph{Hidden Enemy} scenario, where a yellow blob attacks the agent like zombies despite appearing harmless on the danger map. Figure~\ref{fig:crafter_deceptive_enemy1} shows the \emph{Deceptive Enemy} scenario, where certain cows are reprogrammed to behave like zombies.  Figure~\ref{fig:crafter_novel_colors1} shows the \emph{Novel Colors}, in which all grass blocks turn bright yellow, creating a purely visual domain shift. }
    \label{fig:crafter_scenarios}
\end{figure}
\begin{figure}[hbtp]
    \centering
    \begin{subfigure}[b]{0.32\linewidth}
        \centering
        \includegraphics[width=\linewidth]{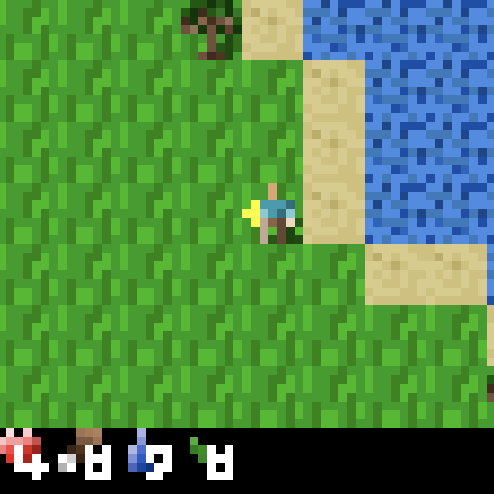}
        \caption{Bird's Eye View}
        \label{fig:crafter_image}
    \end{subfigure}
    \hfill
    \begin{subfigure}[b]{0.32\linewidth}
        \centering
        \includegraphics[width=\linewidth]{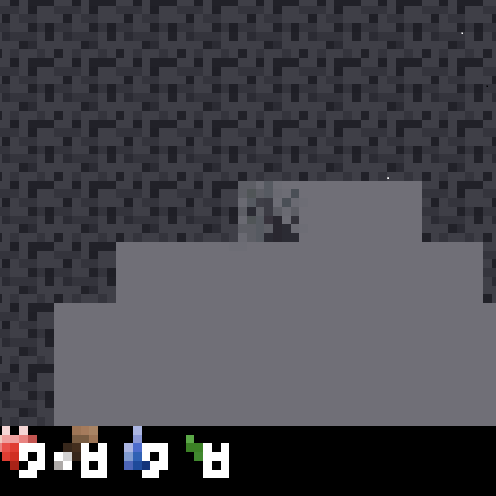}
        \caption{Grayscale}
        \label{fig:crafter_grayscale}
    \end{subfigure}
    \hfill
    \begin{subfigure}[b]{0.32\linewidth}
        \centering
        \includegraphics[width=\linewidth]{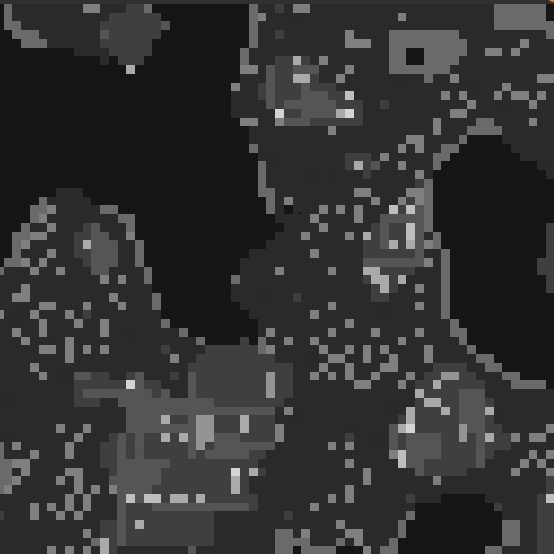}
        \caption{Semantic View}
        \label{fig:crafter_semantic_map}
    \end{subfigure}
    
    \vspace{1em} 
    
    \begin{subfigure}[b]{0.32\linewidth}
        \centering
        \includegraphics[width=\linewidth]{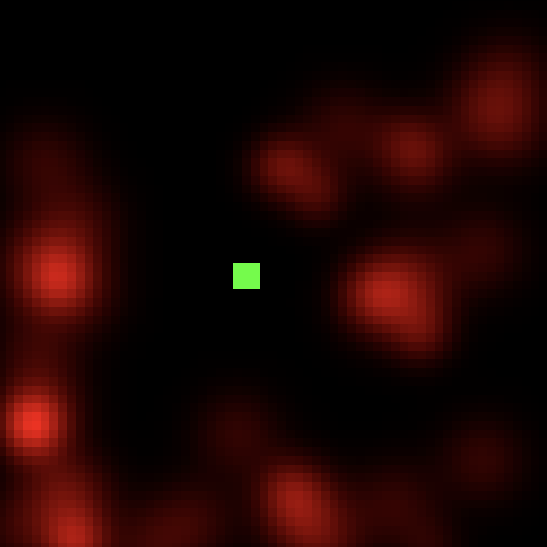}
        \caption{Danger Heatmap}
        \label{fig:crafter_danger_heatmap}
    \end{subfigure}
    \hfill
    \begin{subfigure}[b]{0.32\linewidth}
        \centering
        \includegraphics[width=\linewidth]{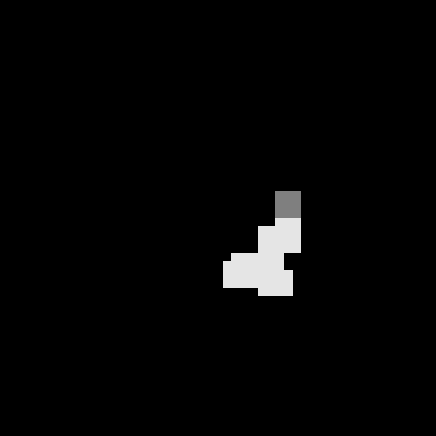}
        \caption{Health Trail}
        \label{fig:crafter_health_trail}
    \end{subfigure}
    \hfill
    \begin{subfigure}[b]{0.32\linewidth}
        \centering
        \includegraphics[width=\linewidth]{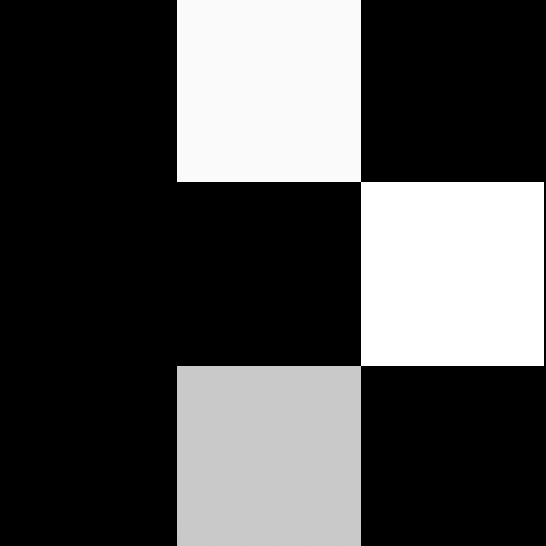}
        \caption{Proximity Grid}
        \label{fig:crafter_proximity_map}
    \end{subfigure}
    
    \caption{Six different representations of the Crafter environment, capturing a diverse range of data, were used to train the agent.}
    \label{fig:crafter_representations1}
\end{figure}

To furnish the agent with a rich state representation, we provide and train with six complementary observation channels (visualizations provided in Figure~\ref{fig:crafter_representations1}):
\begin{itemize}
    \item \textit{Bird's Eye View}: A top-down, default view of the Crafter gameplay centered on the agent and a few blocks around it.
    \item \textit{Grayscale}: A grayscale version of Bird's Eye View.
    \item \textit{Semantic View}: A top-down view of the entire game map showing object locations; built-in Crafter.
    \item \textit{Danger Heatmap}: Shows sources of danger (zombies, skeletons, lava) as red with Gaussian decay; player represented with a green dot.
    \item \textit{Health Trail}: Records agent health at its location through pixel intensity, widened to a 3x3 region to improve training.
    \item \textit{Proximity Grid}: A 3x3 grid representing the eight cardinal directions around the agent; intensities reflect the presence of objects cumulatively, scaled up to 64x64 for training.
\end{itemize}
Unlike Safety Gymnasium and CARLA, the Crafter experiments use a single sensor (top down local view) but with multiple representations (grayscale, danger heatmap, proximity heatmap, etc.).

\begin{figure}[t]
    \centering
    \includegraphics[width=\linewidth]{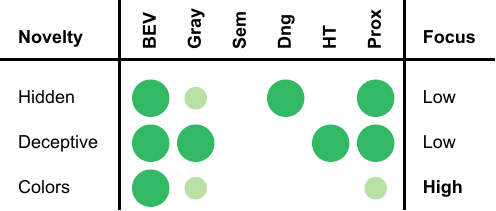}
    \caption{Representation modalities affected by each novelty type. Large circles indicate representations directly modified by the novelty, while small circles indicate representations indirectly affected through downstream effects. The Focus column indicates whether the novelty's impact is concentrated (High) or distributed across multiple modalities (Low).}
    \label{fig:affected_representations}
\end{figure}

We first train the world model in the Crafter domain with Algorithm~\ref{alg:random_masking_training} to prepare for missing representations. To simulate unknown semantic noise, we inject unfamiliar contexts (invisible enemies, different enemy skins, visual color shifts) during inference time. In this setting, we create alterations that have the potential to conditionally affect multiple representations at once rather than independently 
(see Table~\ref{fig:affected_representations} for a brief summary of expected representations affected given the semantic noise introduced).

We monitor the performance of a (1) Base agent, (2) an agent that only implements Step 1 \emph{Isolation} (an agent only selects a single representation) and (3) an agent that implements Step 1 and Step 3 \emph{Recursive Masking} (an agent that recursively masks the observations in order of surprise). We include hardware details in Appendix~\ref{app:hardware}.

We report our findings in Figure~\ref{fig:crafter_scenarios}. We hypothesize that the level of depth generally affects the performance of the agent, depending on how focused the semantic noise is on a subset of representations. Recursive masking appears to improve the performance on changes that are expected to focus on a single representation, such as color based alterations that primarily affect the image representation of the state (Novel Colors). Whereas representation isolation tends to be more useful towards returning the agent towards a predictable policy when many of the representations are affected (Hidden Enemy, Deceptive Enemy, Invert Health). Although we primarily test with out-of-distribution sensor failures, we find agents equipped with Algorithm~\ref{alg:nonadaptive_masking} appear to show some potential towards exhibiting improved stability through automated selection of representations in this setting, potentially opening avenues for future research.

\subsection{Ablations}
\label{sec:ablations}
\subsubsection{Representation Dropout Training}
\begin{figure}[htbp]
    \centering
    \includegraphics[width=.75\linewidth]{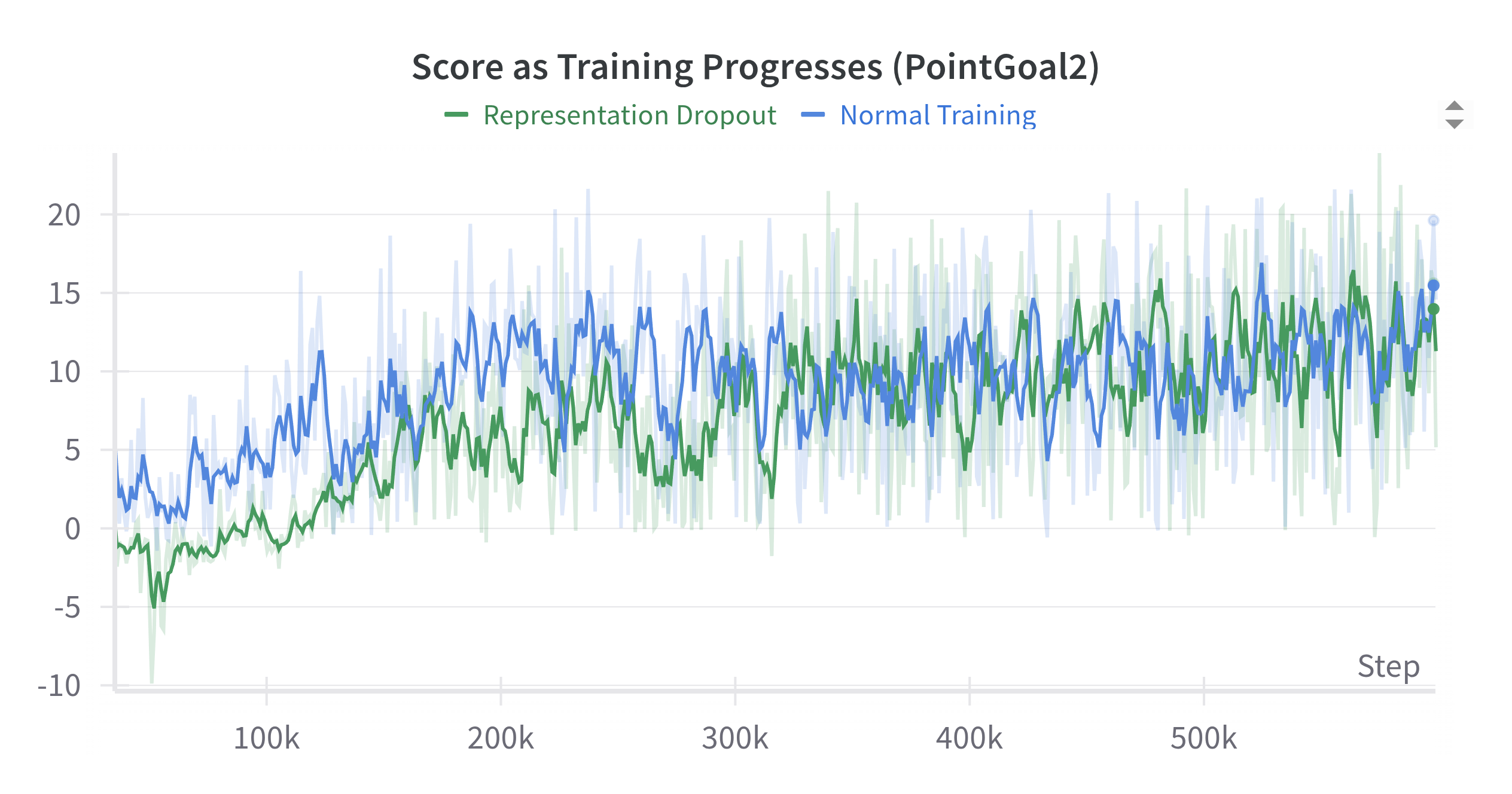}
    \includegraphics[width=.75\linewidth]{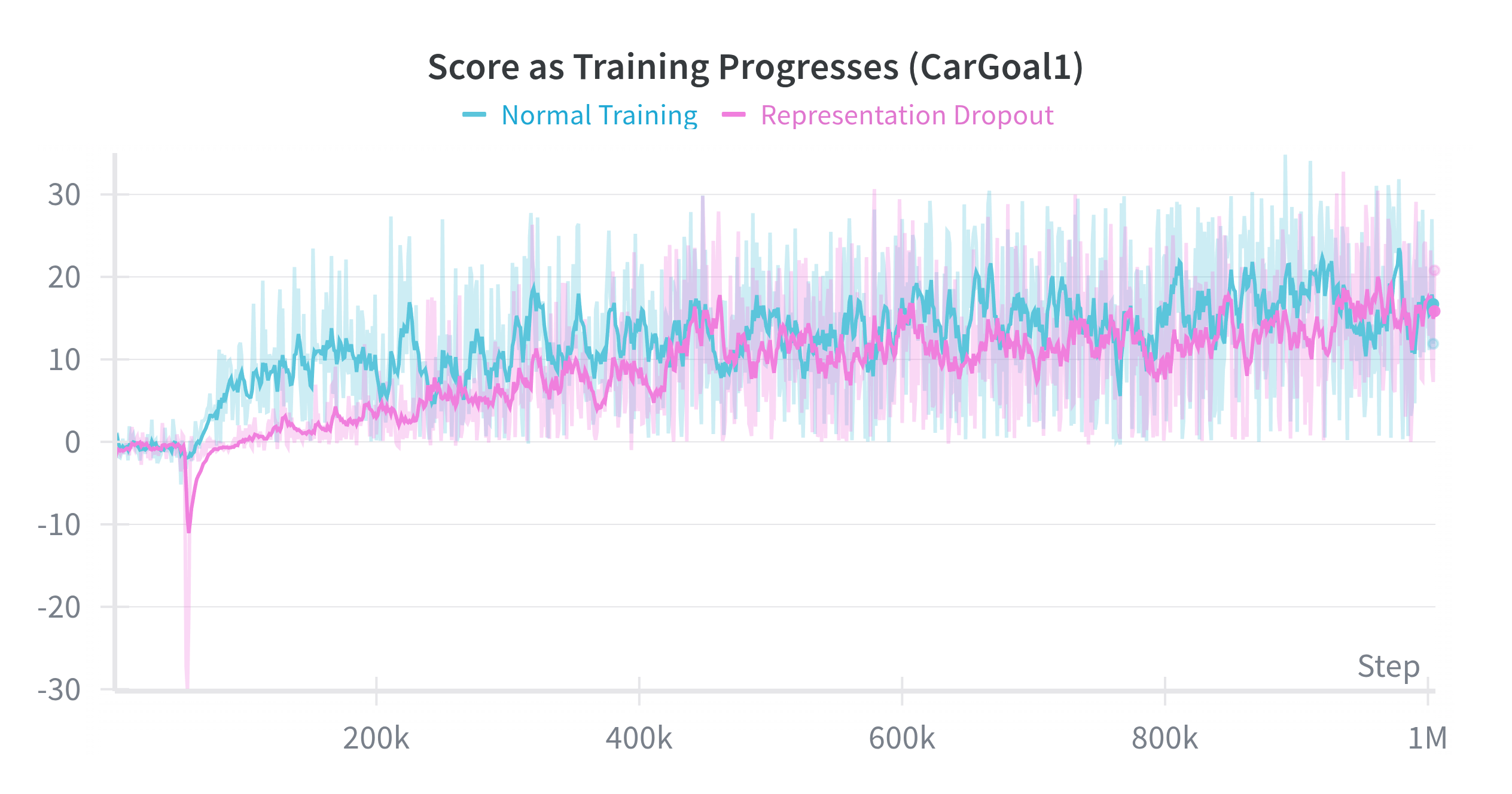}
    \includegraphics[width=.75\linewidth]{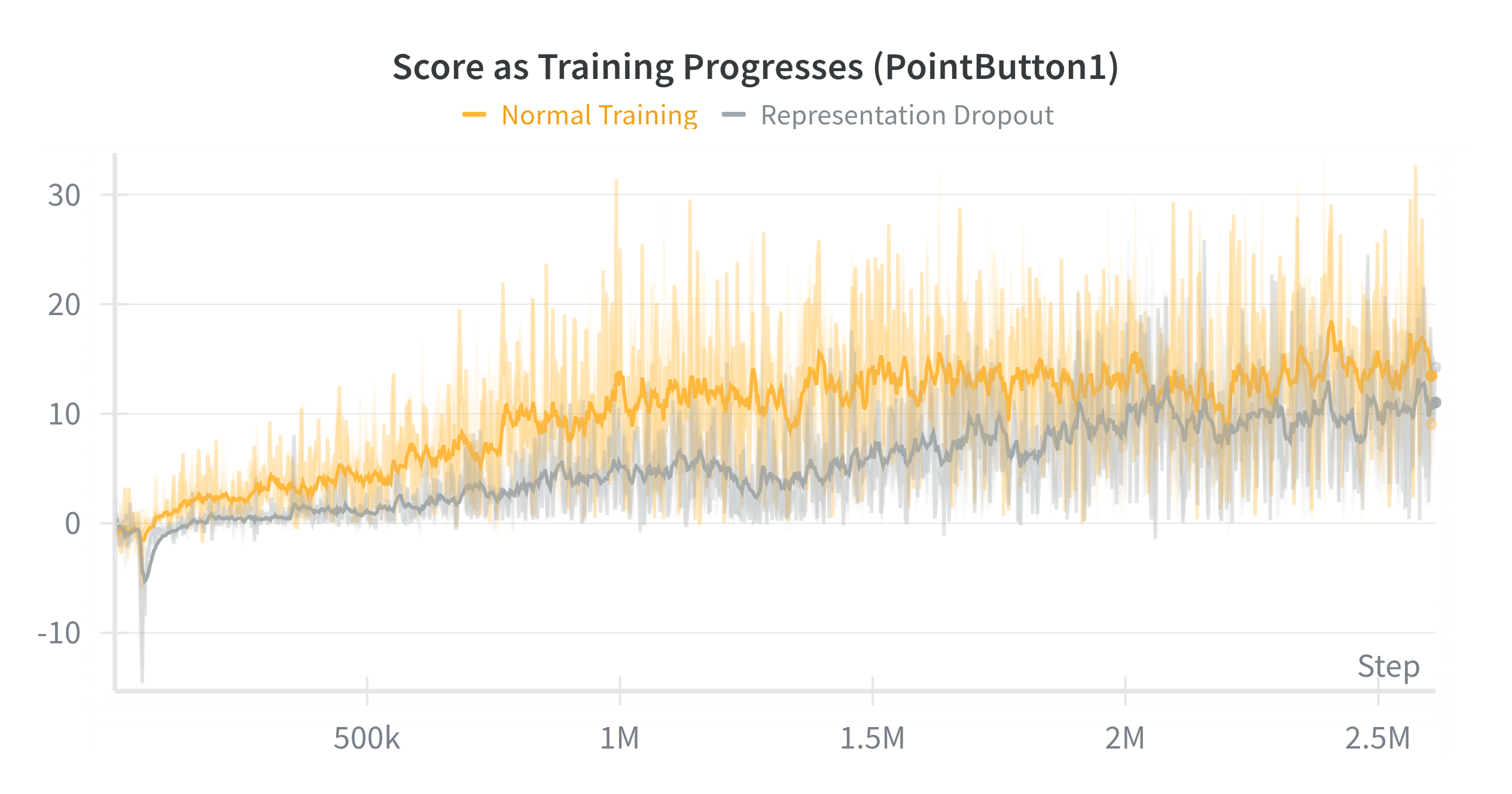}
    \includegraphics[width=.75\linewidth]{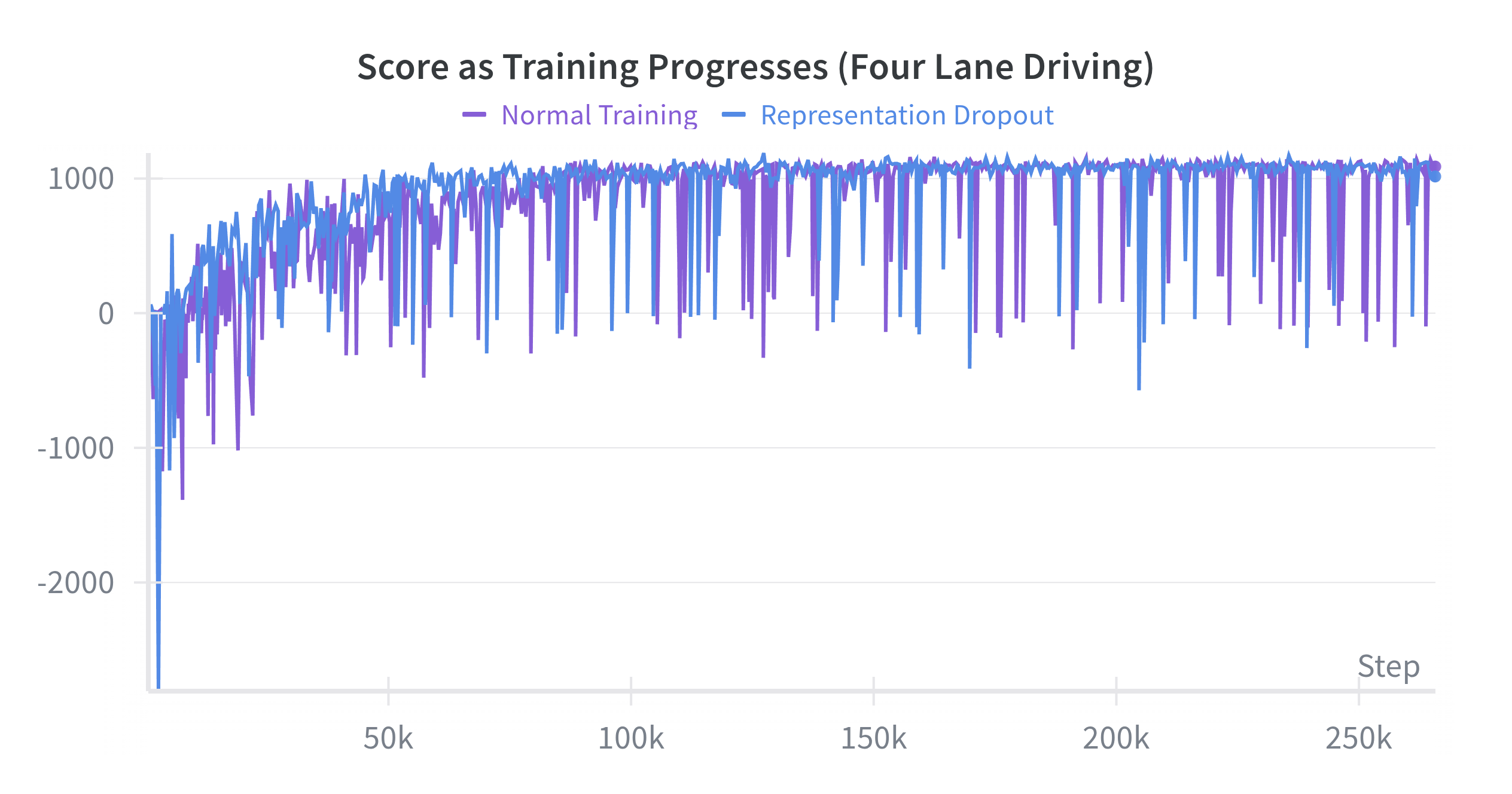}
    \includegraphics[width=.75\linewidth]{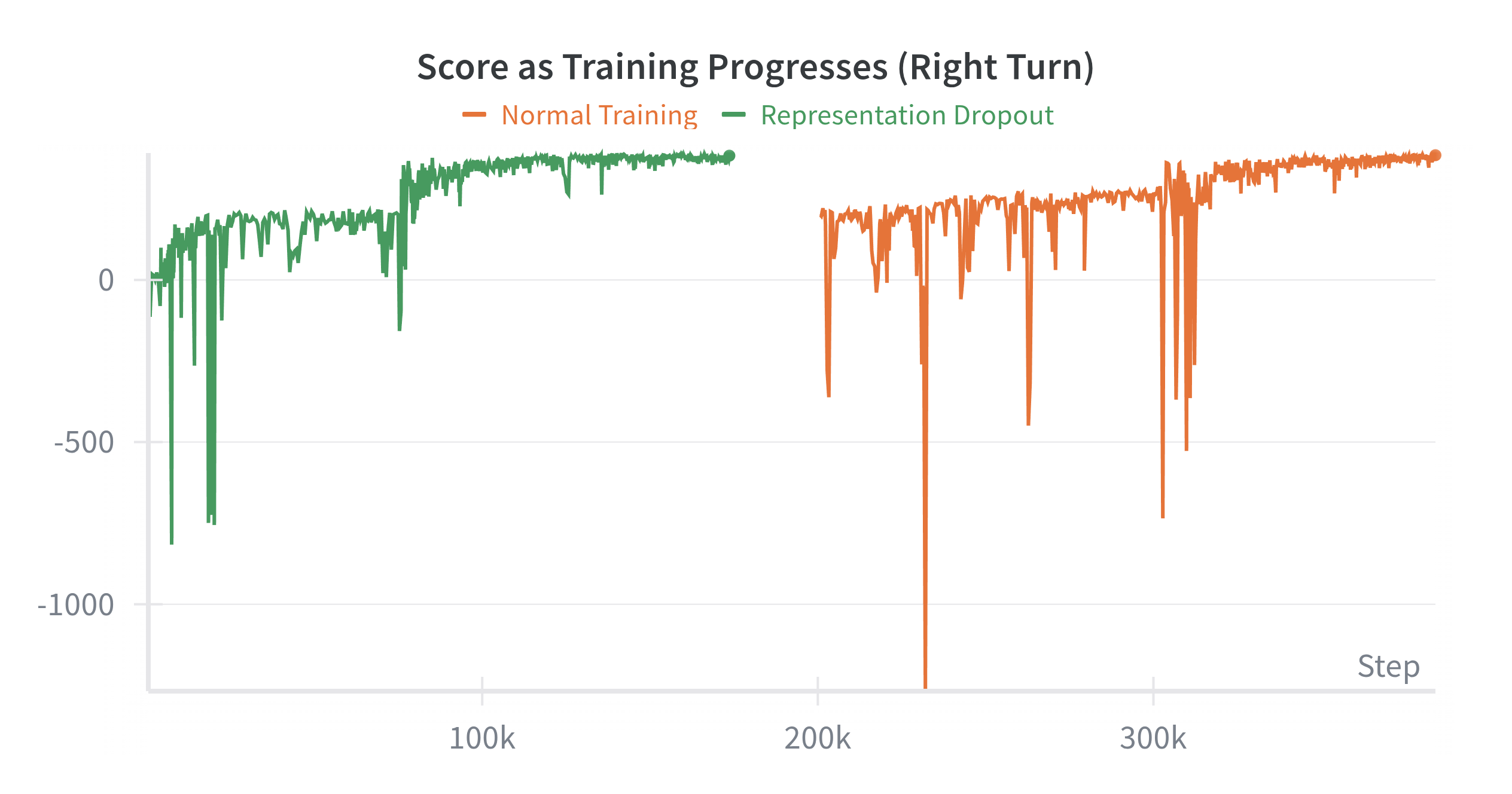}
    \includegraphics[width=.75\linewidth]{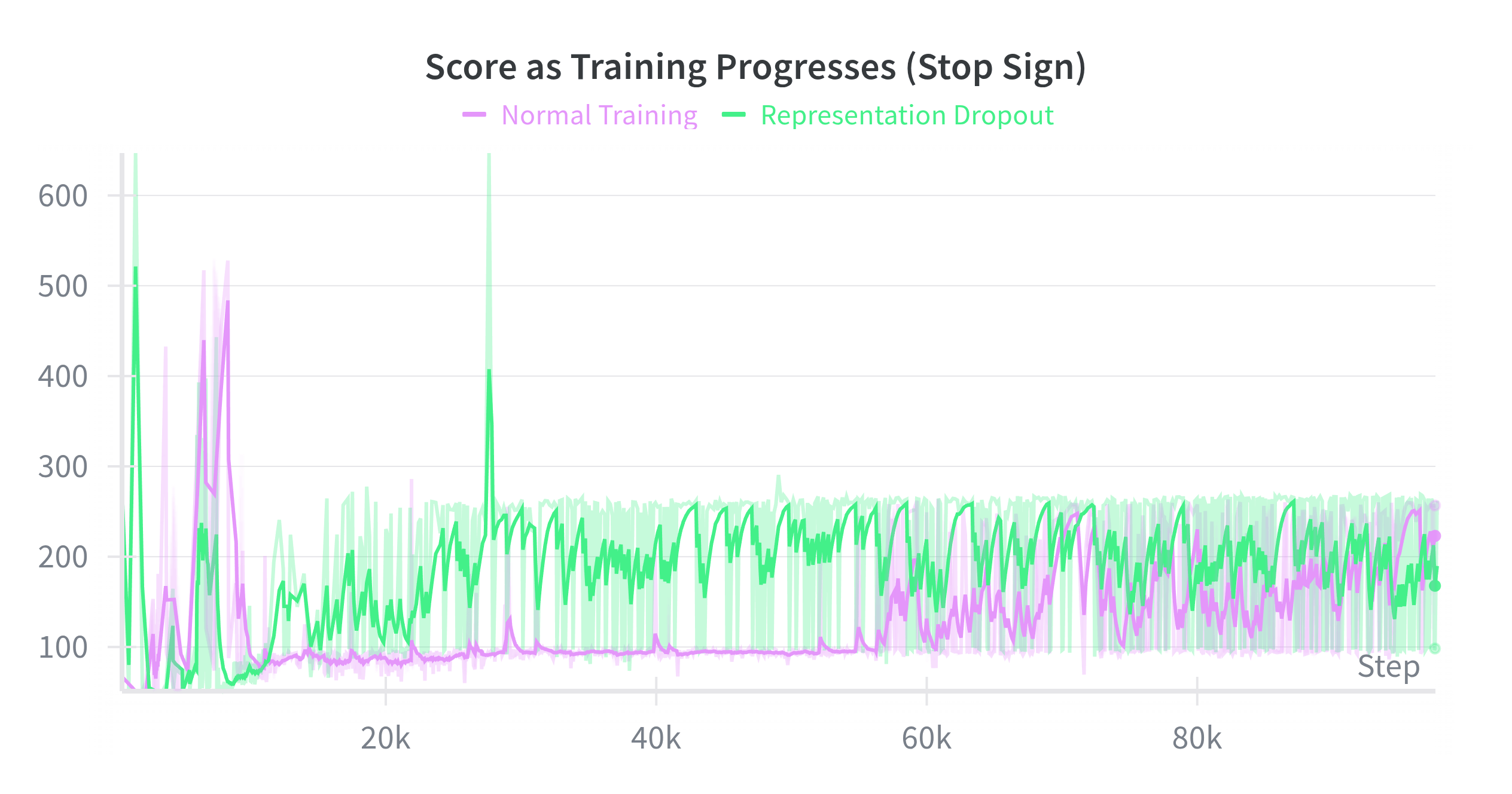}
    \caption{We compare the original world model training in Safety Gymnasium and Carla Domains to the representation dropout training.} 
    \label{fig:dropout_vs_normal_training}
\end{figure}

We further conduct an ablation study shown as Figure~\ref{fig:dropout_vs_normal_training} to evaluate the effect of representation-dropout training on the stability and robustness of the world model. In this setting, a random subset of input modalities or encoded features is masked during each training step, encouraging the model to infer consistent latent dynamics even under partial observations. 

Across all six tasks— PointGoal2, CarGoal1, PointButton1, Four-Lane Driving, Right Turn, and Stop Sign —we observe that the dropout-trained agents achieve comparable or slightly higher final scores than normal training while exhibiting smoother and more stable learning dynamics. Although the early stages of training progress more slowly due to random masking, the representation-dropout models converge to similar or better performance with noticeably reduced variance. This demonstrates that exposing the world model to incomplete or corrupted representations during training improves its ability to maintain coherent latent dynamics under missing-sensor or noisy conditions. The results validate that representation dropout enhances generalization and forms the foundation for the robust filtering and rejection mechanisms used in later experiments.

\subsubsection{Alternative Rejection Scores $M(x_*)$}
We test additional choices of $M(x_*)$ that a user may provide on the Carla Four Lane Task (Denoising is still determined by $D(x_*)$ used in the main results):
\begin{itemize}
    \item Surprise: To decide whether to enter predictive mode, we experiment with utilizing the Bayesian surprise measure discussed in Eq~\ref{eq:bayesian_sup_norm}. 
    \item Resnet-18~\cite{he2015deepresiduallearningimage}: To decide whether to enter predictive mode, we train a Binary Resnet-18 classifier on a synthetic dataset of observations (200,000) with and without Gaussian noise, the goal of the classifier is to reject or accept the observation given to the agent.
\end{itemize}
We report our results across two levels of noise \emph{proportion} in Table~\ref{tab:mx_ablate}.

\begin{table}[h]
\centering
\footnotesize
\resizebox{\columnwidth}{!}{
\setlength{\tabcolsep}{3pt}
\begin{tabular}{lccccc}
\toprule
\textbf{Method} & Gaussian & Chromatic & Jitter & Glare & Occlusion \\
\midrule
Res-18. (.75)     & 571.9$\pm$223.6   & $-188.4\pm359.9$  & $-466.1\pm474.7$  & $-578.1\pm633.9$  & 483.5$\pm$285.9 \\
Surp. (.75)      & 241.7$\pm$287.7   & 105.5$\pm$191.3   & 71.7$\pm$194.9    & 100.4$\pm$130.7   & 275.6$\pm$332.8 \\
Res-18. (.875)    & 391.2$\pm$191.4   & $-439.1\pm421.7$  & $-1237.0\pm638.3$ & $-717.2\pm622.4$  & 228.6$\pm$199.5 \\
Surp. (.875)     & 164.1$\pm$259.1   & $-18.14\pm250.1$  & 39.18$\pm$104.38  & 55.1$\pm$96.93     & 257.5$\pm$281.4 \\
\bottomrule
\end{tabular}}
\caption{Comparison of surprise and trained ResNet-18 rejectors across five noise types with different proportions (.75 and .875).}
\label{tab:mx_ablate}
\end{table}

\subsubsection{Alternative Denoising Methods $D(x_*)$}
We test additional choices of $D(x_*)$ that a user may provide on the Carla Four Lane Task (Acceptance is still determined by $M(x_*)$ used in the main results):
\begin{itemize}
    \item Prior Interpolation: We experiment with a linear interpolation between the world model's expected observation and the current observation.
    \item Restormer~\cite{Zamir2021Restormer}: We leverage an off-the-shelf denoiser with a transformer architecture. 
\end{itemize}
We report our results across two levels of noise \emph{intensity} in Table~\ref{tab:dx_ablate}.
\begin{table}[h]
\centering
\footnotesize
\resizebox{\columnwidth}{!}{
\setlength{\tabcolsep}{3pt}
\begin{tabular}{lccccc}
\toprule
\textbf{Method} & Gaussian & Chromatic & Jitter & Glare & Occclusion \\
\midrule
Interp (.625)    & 316.7$\pm$256.1 & 228.2$\pm$194.6 & 265.3$\pm$295.0 & 466.4$\pm$242.9 & 334.8$\pm$163.0 \\
Restormer (.625) & 262.7$\pm$221.1 & 180.1$\pm$177.3 & 196.1$\pm$170.1 & 308.8$\pm$211.7 & 305.7$\pm$185.4 \\
Interp (.75)     & 258.3$\pm$288.6 & 205.6$\pm$273.4 & 183.8$\pm$401.4 & 448.2$\pm$274.3 & 255.9$\pm$189.9 \\
Restormer (.75)  & 237.0$\pm$299.1 & 144.7$\pm$218.4 & 186.9$\pm$193.3 & 256.6$\pm$217.4 & 215.4$\pm$185.4 \\
\bottomrule
\end{tabular}}
\caption{Comparison of Interpolation and Restormer denoising methods across five noises types with different intensities (.625 and .75).}
\label{tab:dx_ablate}
\end{table}

\subsection{Additional Algorithms}


\label{sec:algorithms}
\begin{algorithm}[h!]
\scriptsize
\caption{Random Multi-Representation Dropout Training}
\label{alg:random_masking_training}
\KwIn{Data dictionary $\mathcal{D}$ with image keys $K$, mask value $m$}
\KwOut{Masked data dictionary $\mathcal{D}'$}
\BlankLine

\textbf{Step 1: Select available images}\;

$K' \gets \{k \in K : k \in \mathcal{D}\}$\;
$n \gets |K'|$\;

\If{$n = 0$}{\Return $\mathcal{D}$}

\BlankLine
\textbf{Step 2: Sample images to mask each step}\;

For each batch $b$ and timestep $t$:\;
\Indp
Draw $u_{b,t} \sim \text{Uniform}\{0,\dots,n-1\}$\;
\Indm

\BlankLine
\textbf{Step 3: Assign random rankings to images}\;

For each $(b,t)$:\;
\Indp
Draw random values $r_{b,t,1},\dots,r_{b,t,n}$\;
Compute rankings $\pi_{b,t}$ by sorting these values\;
\Indm

\BlankLine
\textbf{Step 4: Construct masking matrix}\;

For each $(b,t,i)$:\;
\Indp
Set $\mathrm{mask}[b,t,i] \gets 1$ if $\pi_{b,t}(i) < u_{b,t}$, else $0$\;
\Indm

\BlankLine
\textbf{Step 5: Apply masking}\;

For each $k_i \in K'$:\;
\Indp
$\mathcal{D}'[k_i][b,t] \gets 
\begin{cases}
m, & \text{if mask}[b,t,i] = 1 \\
\mathcal{D}[k_i][b,t], & \text{otherwise}
\end{cases}$\;
\Indm

\BlankLine
\Return $\mathcal{D}'$\;
\end{algorithm}

\begin{algorithm}[htbp]
\scriptsize
\caption{\(O(n \log n)\) Surprise-Guided Representation Selection}
\label{alg:nonadaptive_masking}

\KwIn{Observation dictionary $\mathbf{obs_t}$, sensor keys $K$, world model $\mathrm{WM}$, prev state $z_{t-1}$, prev action $a_{t-1}$, Optional Depth $D$}

\KwOut{Surprise values $\mathcal{S}$, latents $\mathcal{Z}$}

\BlankLine
\textbf{Step 1: Compute surprise for each sensor individually}\;

\ForEach{$k \in K$}{
    Initialize empty observation:\;
    \Indp
        $\mathbf{obs}_{t'} \gets \mathbf{0}$\;
        
        Isolate observation sensor:\;
        $\mathbf{obs}^{k}_{t'} \gets \mathbf{obs}^{k}_t$\;
        
        Encode: $e_{t} \gets \mathrm{Encoder}(\mathbf{obs}_{t'})$\;
        
        Predict posterior distribution:\;
        
        $P_\phi(z^k_t|e_t,h_t) \gets \mathrm{WM}(z_{t-1}, a_{t-1}, e_{t})$\;
        
        Surprise:\;
        
        $S_k \gets \mathrm{KL}\left[P_\phi(z^k_t|e_t,h_t)\,\Vert\, P_\phi(z^k_t|h_t)\right]$\;
        
        Append $S_k$ to $\mathcal{S}$ and $z^k_t \sim P_\phi(z^k_t|e_t,h_t)$ to $\mathcal{Z}$\;
        
    \Indm
}
\BlankLine

\textbf{Step 2: Sort sensors by decreasing surprise}\;

$\pi \gets \mathrm{argsort}\big([S_k]_{k \in K}\big)$\;

\BlankLine
\textbf{Step 3: Iteratively mask sensors in sorted order}\;
$\mathbf{obs}_{t^{'}} \gets \mathbf{obs}_t$\;

\For{$i \gets 1$ \textbf{to} $min(|K|,D)$}{
    \Indp
    Mask sensor $\pi_i$:\;
    $\mathbf{obs}^{\pi_i}_{t^{'}} \gets \mathbf{0}$\;

    Encode: $e_{t} \gets \mathrm{Encoder}(\mathbf{obs}_{t'})$\;
    
    Predict posterior distribution:\;
    
        $P_\phi(z^k_t|e_t,h_t) \gets \mathrm{WM}(z_{t-1}, a_{t-1}, e_{t})$\;
    
    Surprise:\;
    
        $S_k \gets \mathrm{KL}\left[P_\phi(z^k_t|e_t,h_t)\,\Vert\, P_\phi(z^k_t|h_t)\right]$\;
    
    Append $S_k$ to $\mathcal{S}$ and $z^k_t \sim P_\phi(z^k_t|e_t,h_t)$ to $\mathcal{Z}$\;

    \Indm
}
$\mathcal{Z}^* \gets \arg\min_{\mathcal{Z}}
\mathcal{S}(\mathcal{Z})$  

\Return{$\mathcal{Z}^*$}
\end{algorithm}

\end{document}